\documentclass{article}

\usepackage{nac_preprint}
\usepackage[utf8]{inputenc} % allow utf-8 input
\usepackage[T1]{fontenc}    % use 8-bit T1 fonts
\usepackage{hyperref}       % hyperlinks
\usepackage{url}            % simple URL typesetting
\usepackage{booktabs}       % professional-quality tables
\usepackage{amsfonts}       % blackboard math symbols
\usepackage{dsfont}
\usepackage{nicefrac}       % compact symbols for 1/2, etc.
\usepackage{microtype}      % microtypography
\usepackage{comment}
\usepackage{tabularx}

\usepackage{booktabs} % For formal tables
\usepackage{subcaption}
\usepackage{amsmath}
\usepackage{amssymb}
\usepackage{cleveref}
\usepackage{mathtools}
\usepackage[toc,page]{appendix}
\usepackage{enumitem}
\usepackage{graphics}
\usepackage{graphicx}
\usepackage{xcolor}
\usepackage[export]{adjustbox}
\usepackage{algorithm}
\usepackage{algorithmicx}
\usepackage[noend]{algpseudocode}
\algnewcommand{\LineComment}[1]{\State \(//\) #1}
\algnewcommand{\RLineComment}[1]{\State \(\triangleright\) #1}
\usepackage{bm}
\usepackage{multirow}

\usepackage{wrapfig}

\usepackage{etoolbox}
\usepackage{tikz}
\usetikzlibrary{tikzmark}
\usetikzlibrary{calc}

\errorcontextlines\maxdimen

\newcommand{\ALGtikzmarkcolor}{black}% customise this, if you want
\newcommand{\ALGtikzmarkextraindent}{4pt}% customise this, if you want
\newcommand{\ALGtikzmarkverticaloffsetstart}{-.5ex}% customise this, if you want
\newcommand{\ALGtikzmarkverticaloffsetend}{-.5ex}% customise this, if you want
\makeatletter
\newcounter{ALG@tikzmark@tempcnta}

\newcommand\ALG@tikzmark@start{%
    \global\let\ALG@tikzmark@last\ALG@tikzmark@starttext%
    \expandafter\edef\csname ALG@tikzmark@\theALG@nested\endcsname{\theALG@tikzmark@tempcnta}%
    \tikzmark{ALG@tikzmark@start@\csname ALG@tikzmark@\theALG@nested\endcsname}%
    \addtocounter{ALG@tikzmark@tempcnta}{1}%
}

\def\ALG@tikzmark@starttext{start}
\newcommand\ALG@tikzmark@end{%
    \ifx\ALG@tikzmark@last\ALG@tikzmark@starttext
    \else
        \tikzmark{ALG@tikzmark@end@\csname ALG@tikzmark@\theALG@nested\endcsname}%
        \tikz[overlay,remember picture] \draw[\ALGtikzmarkcolor] let \p{S}=($(pic cs:ALG@tikzmark@start@\csname ALG@tikzmark@\theALG@nested\endcsname)+(\ALGtikzmarkextraindent,\ALGtikzmarkverticaloffsetstart)$), \p{E}=($(pic cs:ALG@tikzmark@end@\csname ALG@tikzmark@\theALG@nested\endcsname)+(\ALGtikzmarkextraindent,\ALGtikzmarkverticaloffsetend)$) in (\x{S},\y{S})--(\x{S},\y{E});%
    \fi
    \gdef\ALG@tikzmark@last{end}%
}

\apptocmd{\ALG@beginblock}{\ALG@tikzmark@start}{}{\errmessage{failed to patch}}
\pretocmd{\ALG@endblock}{\ALG@tikzmark@end}{}{\errmessage{failed to patch}}
\makeatother
\title{
Meta-Representational Predictive Coding: Neuroscience-Informed Self-Supervised Learning
}  

\author{%
Alexander G. Ororbia\\
Rochester Institute of Technology \\
Rochester, New York, USA \\
\texttt{ago@cs.rit.edu}
\And
Karl Friston\\
VERSES AI Research Lab \\
Los Angeles, California, USA\\
\texttt{karl.friston@verses.ai}
\And
Rajesh P. N. Rao\\
University of Washington \\
Seattle, Washington, USA  \\
\texttt{rao@cs.washington.edu}
}

\begin{document}

\setlength{\abovedisplayskip}{0.065cm}
\setlength{\belowdisplayskip}{0pt}

\maketitle

\begin{abstract} 
Self-supervised learning has become an important paradigm in the domains of machine intelligence and computational neuroscience. Nevertheless, current work on self-supervised learning (SSL) relies on biologically implausible credit assignment, i.e., backpropagation of errors, and feedforward inference, i.e., a sequential, non-parallel flow of information. Predictive coding (PC) offers a biologically plausible means to avoid backprop-specific limitations. However, unsupervised PC requires learning a generative model of raw input, which entails predicting high dimensional input; on the other hand, supervised PC learns a mapping between inputs to target labels and thus requires human annotation and incurs the drawbacks of supervised learning. In this work, we present a neuroscience-informed SSL model based on PC and active perception that we call meta-representational predictive coding (MPC). MPC sidesteps the need for a generative model of sensory input by learning to predict representations of data across parallel streams, resulting in an encoder-only learning-and-inference scheme. 
%\vspace{-0.5cm}

\end{abstract}

\section*{Introduction}
%\label{sec:intro}

Self-supervised learning has become an increasingly important paradigm in the domain of machine intelligence \cite{ericsson2022self,geiping2023cookbook}. Furthermore, some forms of self-supervised adaptation, such as contrastive setups -- which learn how to invert the process generating data samples \cite{zimmermann2021contrastive} -- have emerged in computational neuroscience and brain-inspired computing \cite{ororbia2024contrastive}. % \cite{ororbia2024contrastive,prince2024contrastive}
Nevertheless, current work on self-supervised learning (SSL) relies on biologically-implausible credit assignment, i.e., in the form of backpropagation of errors \cite{ororbia2023brain} %\cite{rumelhart1986learning} 
(backprop), and inference, i.e., typically where information flows step-by-step in one direction (in a `forward-locked' feedforward pass \cite{jaderberg2017decoupled,ororbia2023brain}). A scheme that could conduct this kind of learning yet in a biologically-plausible (backprop-free) fashion manner would be valuable. However, current computational and mechanistic frameworks, such as many predictive coding schemes \cite{rao1999predictive,ororbia2022ngc,salvatori2023brain} %\cite{rao1999predictive,rao1999visionresearch,ororbia2022ngc,salvatori2023brain} 
that provide viable accounts of neurobiological inference schemes (via message passing) and credit assignment (via local plasticity rules), are primarily formulated for learning complex generative models of raw sensory input or mapping functions between input and supervisory signals. These unsupervised and supervised forms of predictive coding, however, do not speak to a reverse perspective of neuronal inference and learning: \emph{what might predictive coding look like if it only learned a generator of latent states (an encoder), as opposed to a generator of sensory states (a generative decoder)?} This could open the door to a form of \emph{self-supervised predictive coding} focused on learning distributed representations of sensory stimuli without explicitly modeling high dimensional inputs. 

In this work, we invert the premise of predictive processing 
\cite{salvatori2023brain} %\cite{elias1955predictive,srinivasan1982predictive,friston2008hierarchical,clark2015surfing,seth2021predictive} 
from top-down generative learning to bottom-up representation acquisition, casting the goal of free energy \cite{friston2010free} %\cite{friston2009predictive,friston2010free} 
minimization as prediction in latent distributed representation spaces. To do so, we draw inspiration from how the visual system \cite{felleman1991distributed,livingstone1988segregation,nealey1994magnocellular} processes stimuli through central and peripheral \textit{streams} and eye movements such as saccades (i.e., active vision). Concretely, we frame inference and learning in the context of a predictive coding (PC) scheme that comprises an architecture of neuronal streams, some of which process central (high-resolution) views of the input while others process peripheral (low-resolution) ones. These central and peripheral streams interact by predicting the dynamics/behavior of one another. %
As a result, we propose a generalization of predictive coding that conducts a form of encoder-only, joint-embedding \cite{assran2023self} SSL that we call \textbf{meta-representational predictive coding} (MPC); our work presents evidence that this NeuroAI \cite{zador2026neuroai} form of SSL (what we call ``Neuro-SSL'') is capable of robustly, efficiently acquiring latent representations of data useful for downstream machine intelligence tasks. 

\section*{Results}

\paragraph{Notation.} In this work, we use $\odot$ to indicate the Hadamard product (or element-wise multiplication), $\cdot$ to denote matrix/vector multiplication, and $*$ to denote convolution. 
$(\mathbf{v})^T$ is the transpose of $\mathbf{v}$. Matrices/vectors are depicted in bold font, e.g., matrix $\mathbf{M}$ or vector $\mathbf{v}$ (scalars shown in italics). $\mathbf{z}_j$ will refer to the $j$th element of vector $\mathbf{z}$. $<\mathbf{a},\mathbf{b}>$ denotes vector concatenation along the column dimension. Finally, $||\mathbf{v}||_2$ denotes the Euclidean norm of vector $\mathbf{v}$. Sensory input has shape $\mathbf{x} \in \mathcal{R}^{J_0 \times 1}$ ($J_0$ is the number of input features), and a neural layer has shape $\mathbf{z}^\ell \in \mathcal{R}^{J_\ell \times 1}$ ($J_\ell$ is the number of neurons for layer $\ell$). Matrix flattening (to a column vector of length equal to the product of the matrix's number of column and rows) is denoted as $\text{Flat}()$.

\paragraph{Problem setting and motivation.} 
As a starting point for how we structure the requisite neuronal model, we formulate its ability to selectively sample its sensorium. In terms of the model structure, we draw inspiration from the functional anatomy \cite{zeki1988functional} (of animals and humans) in terms of central and peripheral vision \cite{levy2001center}. %\cite{strasburger2011peripheral,wang2017central}.
\footnote{We regard this morphological separation underlying central and peripheral cortical anatomy as a useful biological mean-field approximation (MFA). There are many MFAs that emerge in biological self-organization and the general idea behind biological manifestations of inference-and-learning that leverage them is that the biological system will work to ``repair'' the falsehood(s) induced by such MFAs. Thus, a neural system, like the one we propose in this work, must work to compensate for the statistical independence assumptions between central and peripheral structures thus requiring mechanisms for passing signals between these specialized regions. This motivates our (later-described) message-passing scheme that links these structures.}   
In visual processing, % \cite{rensink2000dynamic}, 
central and peripheral vision both play a key role. Central vision, further decomposed into foveal (extending to one degree of eccentricity from the visual field's center containing the highest density of cone receptors with highest resolution \cite{polyak1941retina,curcio1990human,loschky2005limits}) and parafoveal (extending $4$-$5$ degrees of eccentricity yet containing a high density of slightly lower-resolution rod receptors \cite{rayner1981masking,van1998functional}) vision reports higher-resolution, detailed sensory information whereas peripheral vision focuses on encoding coarser-grained, lower-resolution data features. Due to its higher density (and smaller receptive field size) of rods and cones, central vision is thought to be important for high spatial frequency recognition tasks \cite{musel2013retinotopic} % \cite{sasaki2001local,musel2013retinotopic} 
(such as recognizing an object or face). On the other hand, peripheral vision, with the highest proportion of rods at the lowest spatial resolution, is generally viewed as important for low spatial frequency tasks that require obtaining a global gist (``bigger picture'') of a scene \cite{mccotter2005use,loschky2007importance}. %\cite{sanocki2003representation,mccotter2005use,loschky2007importance}. 
While it is the case that vision tends to call on central and peripheral vision differently, depending on the task \cite{wang2017central}, it is clear that these two visual streams are complementary in constructing a complete representation of the stimuli being observed at any instant. 

% %%%%%%%%%%%%%%%%%%%%%%%%%%%%%%%%%%%%%%%%%%%%%%%%%%%%%%%%%%%%%%%%%%%%%%%%%%%%%%%%%%%%
% %% glimpse sequence and foveal/peripheral views
% \begin{wrapfigure}{r}{0.525\textwidth}
% \vspace{-0.4cm}
%   \begin{center}
%     \includegraphics[width=0.27\textwidth]{figs/mpc_saccades.png}
%   \end{center}
%   \vspace{-0.25cm}
%   \caption{
%   \small{
%   \textbf{Illustration of two consecutive MPC observations of an image.} Shown is one of the digits processed by the MPC scheme over the course of two consecutive saccade-produced glimpses. On the left is the full source image with a red dashed box showing the approximate subspace sampled by the glimpse. The right panels, within the expanded dot-dashed rectangle, show how the sampled data within the dashed box on the left is converted into six input representations, i.e., four overlapping ``fovea patches'', a ``parafovea patch'', and a ``peripheral patch''.
%   }
%   }
%   \label{fig:saccade_scheme_example}
%   \vspace{-0.35cm}
% \end{wrapfigure}
% %%%%%%%%%%%%%%%%%%%%%%%%%%%%%%%%%%%%%%%%%%%%%%%%%%%%%%%%%%%%%%%%%%%%%%%%%%%%%%%%%%%%  

Neuroscientific evidence, in the form of brain-imaging studies, further demonstrates that orderly peripheral and central representations form in both low-level retinotopic visual areas, i.e., V1 to V4, as well in higher regions that characterize the ventral temporal cortex \cite{levy2001center,grill2004human,arcaro2009retinotopic} %\cite{levy2001center,malach2002topography,hasson2003large,grill2004human,arcaro2009retinotopic} 
as a result of the inference and learning, evinced as visual perception or recognition (of faces, objects, or scenes). We draw inspiration from the structural organization underwriting visual perception -- the visual processing afforded by foveal, parafoveal, and peripheral viewing -- in constructing a neuronal circuit that (loosely) emulates this computational architecture. Furthermore, we present a PC perspective \cite{friston2010free,salvatori2023brain} on the accompanying neuronal circuit's message passing and plasticity -- our neuronal model's inference and synaptic adjustments are driven by the predictions induced across foveal, parafoveal, and peripheral streams/systems, in service of optimizing free energy \cite{friston2010free}.  

In addition to structure, we draw inspiration from the oculomotor system that underwrites active vision \cite{perrinet2014active}. %\cite{zeki2005ferrier,wurtz2011thalamic,perrinet2014active,ognibene2014ecological}. 
The ensuing visual palpation (and implicit epistemic foraging) \cite{hayhoe2005eye} can be roughly broken down into four key movements: 
saccades, smooth pursuit movements, and vergence and vestibulo-ocular movements. Saccades \cite{liversedge2000saccadic} %\cite{liversedge2000saccadic,krauzlis2005control,findlay2009saccadic} 
are rapid, ballistic (jumpy/jerky) movements which change the eye's fixation point, the movement/shift of which ranges from smaller (in tasks such as reading) to larger saccades (examining a scene). Saccades are generally more involuntary (unconscious) than voluntary and can, from a modeling perspective, appear to be more itinerant in nature. Smooth pursuit movement \cite{robinson1965mechanics} entails slower movements that focus on keeping a moving stimulus on the fovea and are more voluntary in nature. Vergence movement \cite{pierrot2004eye} works toward aligning the fovea of each eye with targets at different distances (from the observer) whereas vestibulo-ocular movements \cite{westheimer1975visual} serve to stabilize the eyes relative to the observer's niche (working to compensate for head movement, preventing the slippage of visual stimuli on the retina's surface as the head moves). 

\begin{figure}[t]
\centering
    \begin{subfigure}[b]{0.485\textwidth}
    \centering
    \includegraphics[width=0.7\textwidth]{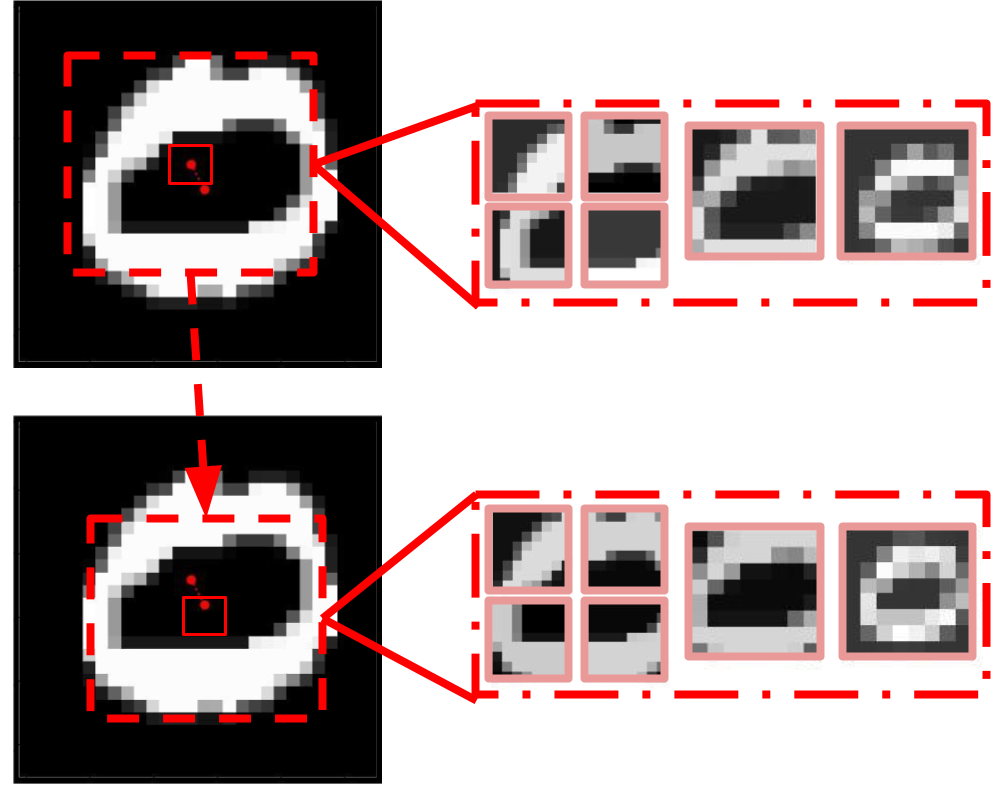}
    \caption{MPC saccade patch-extraction.}
    \label{fig:saccade_scheme_example}
    \end{subfigure}
    \begin{subfigure}[b]{0.485\textwidth}
    \centering
    \includegraphics[width=1.0\textwidth]{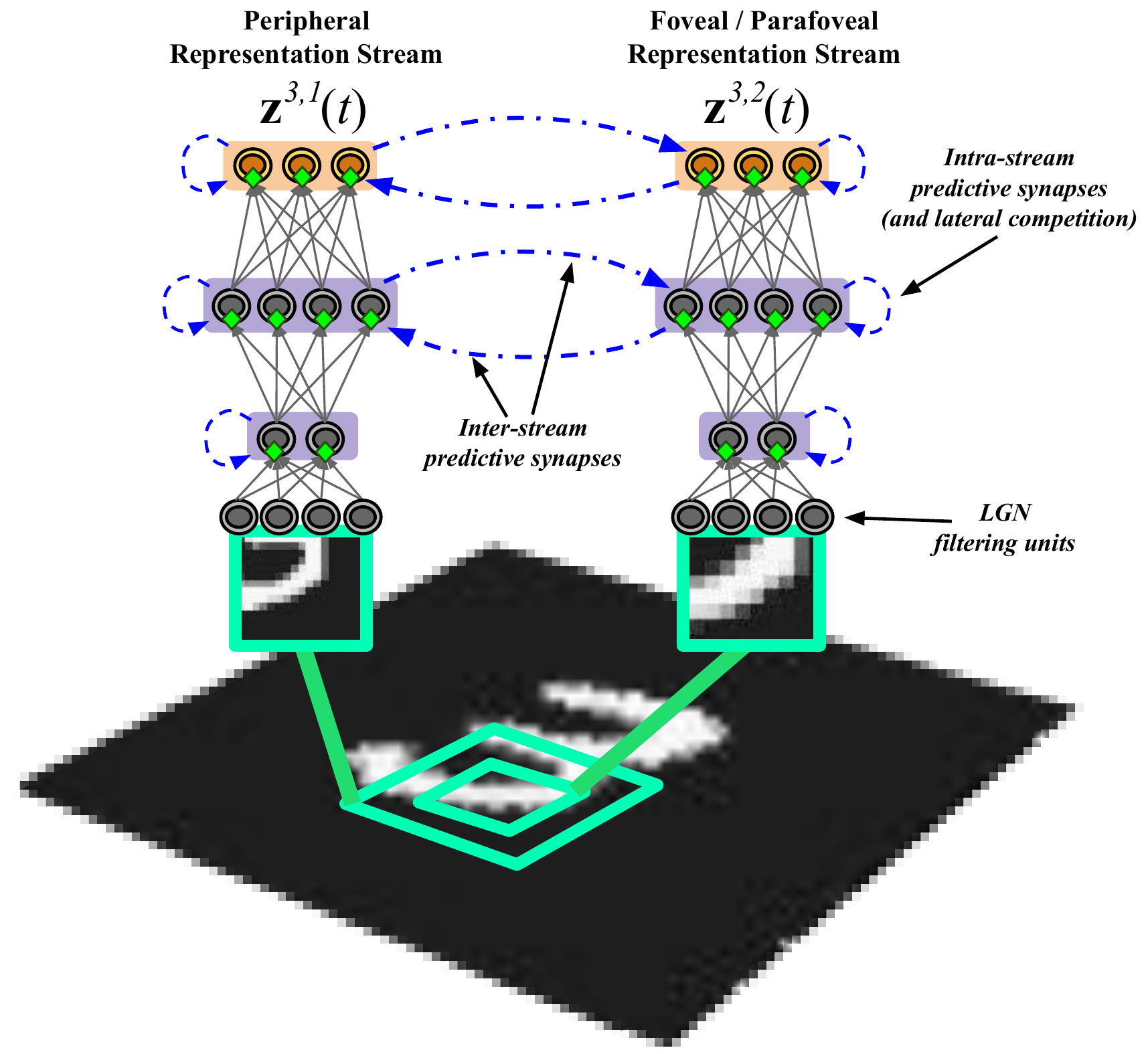}
    \caption{MPC architectural illustration.}
    \label{fig:mpc_architecture_simple}
    \end{subfigure}
\caption{
\small{
\textbf{Graphical depiction of a simple dual stream, MPC architecture and its glimpsing of sensory input.} 
(Panel \textbf{a}) %Illustration of two consecutive MPC observations of an image.
Shown is one of the digits processed by the MPC scheme over the course of two consecutive saccade-produced glimpses. On the left is the full source image with a red dashed box showing the approximate subspace sampled by the glimpse. The sub-images to its right, within the red dot-dashed rectangular inset, show how the sampled data within the dashed box on the left is converted into six input representations, i.e., four overlapping ``fovea'' patches, a ``parafovea'' patch, and a ``peripheral'' patch. 
(Panel \textbf{b}) A meta-representational predictive coding (MPC) architecture works by processing (LGN-filtered) inputs via two or more resolutions of the sensory input, mimicking the coarseness (acuity) of the spatial features extracted by the human eye. 
%generally at different resolutions which mimic the coarseness (acuity) of the spatial features extracted by the foveal/parafoveal and the peripheral streams of the human eye. 
In this image, we depict a two-stream architecture, where, at time $t$, one  stream produces a peripheral representation $\mathbf{z}^{3,1}(t)$ (in its third hidden layer) of its input while another circuit produces a foveal (`central') representation $\mathbf{z}^{3,2}(t)$ of the input. 
The peripheral MPC stream attempts to predict the activities of the foveal MPC circuit (and itself) and \textit{vice versa} (for the $tau$-th glimpse of an image). Notice that all streams are conditioned by actions, i.e., normalized x-y coordinates of the fixation point of all foveal/parafoveal/peripheral views, taken by a saccade over the input. %as well as possibly their prior expectation (at time $t-1$). 
In this work, a fixed $K$-length saccade sequence is produced by actively choosing where to jump across the sensory space, resulting in a reflexive perceptual input sampling policy. 
Green diamonds indicate error neuron populations, light-gray or orange circles with slightly darker colors within denote state cell populations, light gray arrows represent synaptic connections, dashed black circular arcs depict recurrent synapses, and blue dash-dotted arcs denote lateral cross-circuit prediction synapses (not shown, to improve visual clarity, are feedback synapses).
}
}
\label{fig:fov_rpc}
\vspace{-0.5cm}
\end{figure}

For the purposes of this study, we focus on saccades as the driver of our model's information foraging, since the saccade is among the most common eye movements in humans (during waking hours) \cite{liversedge2000saccadic}. %\cite{efron2019lidwiper,liversedge2000saccadic}. 
In future research, we will extend MPC to incorporate further contextual control \cite{rothkopf2007task} %\cite{rothkopf2007task,ballard2009modelling}
to drive motion such as smooth pursuit. Using saccades means that our models use an approximation of the quick, jumpy saccadic movements of biological eyes to extract partial information (``glimpses'') from the visual scene, e.g., a pixel image. 
See Methods and Figure \ref{fig:fov_rpc} (Panel \textbf{a}) for the  specification of how saccades are made and how glimpse information is extracted from image data. 

\subsection*{Meta-Representational Predictive Coding} 

\paragraph{LGN input-level processing.} At the very bottom layer (the sensory input layer) of any one of the MPC streams, raw pixel intensities of a particular glimpse are first encoded through a set of filtering units; these units will aid the MPC circuit in processing particularly natural images effectively. Specifically, these units are designed to work as embedded sensory layer neurons that filter input information, inspired by the functionality of the lateral geniculate nucleus (LGN) \cite{weyand2016multifunctional} %\cite{croner1995receptive,weyand2016multifunctional}
and classical early vision models. Specifically, in our input LGN process layer, we model the antagonistic center-surround receptive fields \cite{rodieck1965quantitative} of retinal ganglion cells, coupled with LGN relay neurons, through the combination of a difference/ratio-of-Gaussians (DoG) %\cite{marr1980theory} 
spatial filtering and a form of %divisive, 
contrast normalization \cite{carandini2012normalization} based on dynamic gain control, using the Naka-Rushton shunting mechanism \cite{naka1966s}. 
In Methods, we describe the mechanics and formal implementation of these LGN sensory-input units. 

% \paragraph{Free energy optimization and dynamics.} \textcolor{red}{TODO:}
% \subsection{Dynamic Prediction of Latent Space Characteristics} %%
% \label{sec:pc_latent}
\paragraph{Dynamic prediction of latent space characteristics. } 
To describe the non-input layers of the MPC circuit (which, in a sense, functionally play the role of V1), we start with the objective that it seeks to optimize. In effect, we require predictions to be made with respect to only ``parts'' of its internal representations, i.e., predictions are made in portions of latent representation space. To decide what makes predictions (and what the targets of these will be), we treat the circuit as an architecture of foveal, parafoveal, and peripheral streams where each stream is specialized to receive and encode only a single foveal, parafoveal, or peripheral input. Crucially, this architecture of streams, further inspired by the multi/dual-streams hypothesis \cite{goodale1992separate} 
\footnote{Note that the two-streams hypothesis refers to specialized circuitry related to ``what'' (ventral) and ``where'' (dorsal) pathways. Although our architecture does not faithfully implement the neuroanatomical dorsal and ventral pathways, it does embody the spirit of the what-where distinction and implicit factorisation (mean-field approximation). In effect, our model's foveal streams acquire fine-grained information (high-resolution stroke/arc components) whereas the parafoveal and peripheral streams acquire coarse-grained information (low-resolution, object/part chunks) within which the finer-grained information is situated, i.e., the foveal visual primitives indicate what is being detected whereas the parafoveal/peripheral streams indicate, in coordination with the encoded glimpse coordinates (see Supplementary Note 6), where the primitives can be found in the context of the ``bigger picture''.}, %\cite{mishkin1982contribution,goodale1992separate}
must learn to predict the activity dynamics of one another. This means that each stream is continually guessing what the other streams are encoding and each stream must adjust its own activities based on how wrong its guesses are. This guessing game results in a lateral, cross-stream prediction scheme, of which there are many variations that could be studied (one simple design is considered in this work). 
This type of prediction scheme enables neuronal circuits to predict the statistical properties of a latent space without making ``downward'' (decoder-oriented) predictions of raw sensory input. From this perspective, a successful architecture of streams would be one that seeks consistency or coherence among distributed representations.

%%%%%%%%%%%%%%%%%%%%%%%%%%%%%%%%%%%%%%%%%%%%%%%%%%%%%%%%%%%%%%%%%%%%%%%
\begin{figure}[!t]
     \centering
     \includegraphics[width=0.775\textwidth]{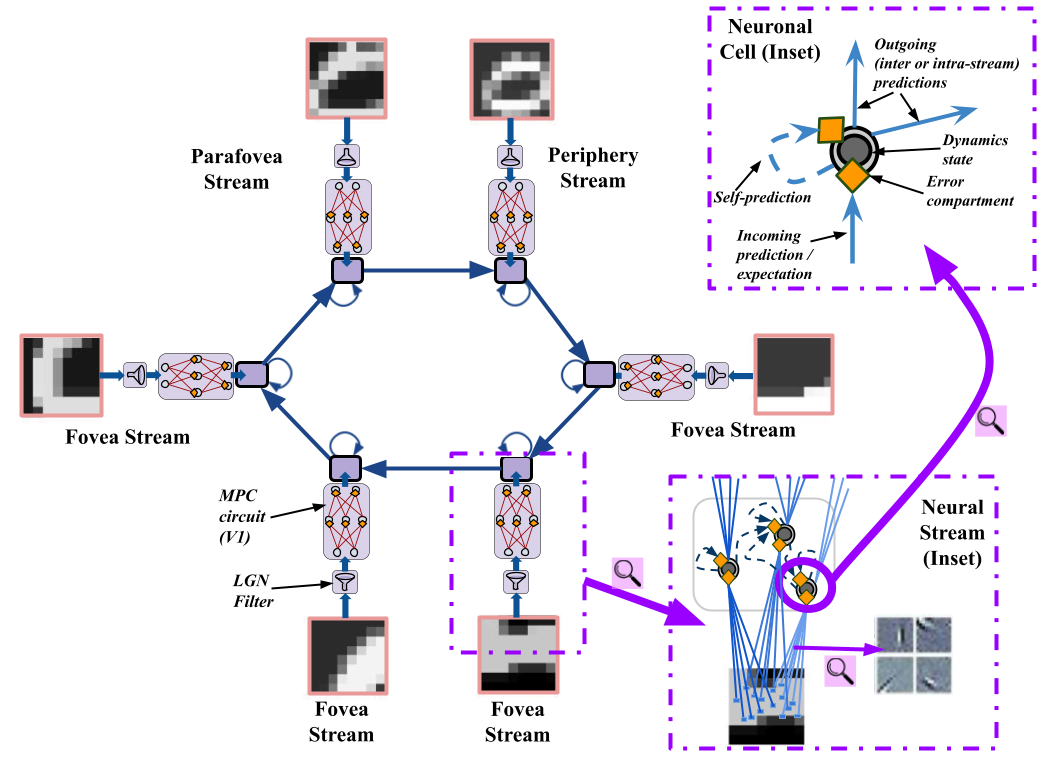}
    \vspace{-0.2cm}
    \caption{ 
    \small{
    \textbf{Visualization of MPC inter- and intra-stream prediction pattern (one layer).} 
    Above is shown, for the MPC encoder studied in this work, the intra- and inter-stream prediction scheme for how the individual streams interact with one another (within the first layer of the architecture; the prediction pattern repeats for higher-level layers); dark blue arrows indicate a prediction direction (blue arrow head ends on prediction target) from which error messages flow backwards. 
    Note that each stream is labeled by the resolution of sensory information that it is specialized to encode (e.g., ``Fovea'', ``Parafovea'', and ``Peripheral''). Each stream is, internally, composed of layers of neurons that make predictions of the activity other neurons (and of their own) and further receive predictions from neurons in other layers of the same circuit or those from other circuits/streams, ultimately encoding the error between predictions they receive and their own current actual activity value. At the bottom of each stream are lateral geniculate nucleus (LGN) inspired units that directly encode sensory stream information.     
    %Each stream is composed of multiple layers of neuronal units that receive predictions from the layer below (or, in the very first layer, these units receive predictions made by the lateral geniculate nucleus (LGN) inspired filter units, embedded at the sensory input level) as well as make predictions of nearby, neighboring population(s) within the same level and of their own activity values.    
    }
    }
    \label{fig:mpc_prediction_pattern}
    \vspace{-0.5cm}
\end{figure}
%%%%%%%%%%%%%%%%%%%%%%%%%%%%%%%%%%%%%%%%%%%%%%%%%%%%%%%%%%%%%%%%%%%%%%%

Within an MPC architecture, each individual stream, arranged in a heterarchical or hierarchical structure, follows a variational free energy (VFE) \cite{friston2010free} 
%\cite{friston2005theory,friston2008hierarchical,friston2010free}
gradient flow that is driven by at least one other stream. In this work, we take this to mean that one stream, driven by a particular view of the sensory stimulus (at one scale/resolution, e.g., a foveal view), seeks to predict the activity values of itself as well as those of another stream that is driven by a different, yet complementary, view of the same stimulus (this could be another view but possibly at a different scale/resolution, e.g., a peripheral view). Based on the sensory views produced by a saccade process (described in Methods), %Section \ref{sec:sensory_saccades}, 
the MPC architecture will consist of several ($C$) ``foveal'' neuronal streams, ($F$) ``parafoveal'' streams, and ($P$) ``peripheral'' streams that seek to predict one another, resulting in a message passing scheme combining intra-stream message passing (internally-communicated mismatch signals) and inter-stream message passing (mismatches between streams). 

For a general architecture made of $V = C + F + P$ streams (all are assumed to have the same number of $L$ layers and $\ell = 0$ indexes the LGN-filtered sensory layer, which receives glimpse input $\mathbf{g}(\tau)$), with the $v$-th stream composed of synaptic parameters 
% $\Theta^v = \{ \mathbf{W}^{\ell,v}, \Sigma^{\ell,v}, \mathbf{A}^{\ell,v,1}, \mathbf{R}^{\ell,v,1}, \Sigma^{\ell,v,1}_C ,...,\mathbf{A}^{\ell,v,V}, \mathbf{R}^{\ell,v,V}, \Sigma^{\ell,v,V}_C \}^L_{\ell=1}$,
$\Theta^v = \big\{ \mathbf{W}^{\ell,v}, \Sigma^{\ell,v}, \{ \mathbf{A}^{\ell,v,q}, \mathbf{R}^{\ell,v,q}, \Sigma^{\ell,v,q}_C \}^V_{q=1} \big\}^L_{\ell=1}$, 
the resulting energy for the $v$-th stream -- which tries to predict the latent representation of any other stream ($q \neq v$) as well as its own ($v=q$) -- is given by: 
% \begin{align}
%     \mathcal{F}_v(\Theta^v, \mathbf{M}^v, \mathbf{C}^v) = \underbrace{\sum^L_{\ell=1} \sum_q \mathcal{N}\Big(\mathbf{z}^{\ell,q}(t); \mu^{\ell,v,q}_C, \Sigma^{\ell,v,q}_C \Big)}_{\text{Cross-Representation Term}}
%     + \underbrace{\sum^L_{\ell=1}\mathcal{N}\big( \mathbf{z}^{\ell,v}(t); \mu^{\ell,v}, \Sigma^{\ell,v} \big)}_{\text{Residual Energy}} + \underbrace{\sum_{p,i,j} \mathcal{N}(\Theta^v[p]_{ij}; 0, \lambda_w)}_{\text{Synaptic prior}} \label{eqn:rpc_vfe}
% \end{align}
\begin{align}
    \mathcal{F}_v(\Theta^v) &= \underbrace{\sum^L_{\ell=1} \sum_q \gamma_{v,q} \mathcal{N}\Big(\mathbf{z}^{\ell,q}(t); \mu^{\ell,v,q}_C, \Sigma^{\ell,v,q}_C \Big)}_{\text{Cross-Representation Term}} 
    + \underbrace{\sum^L_{\ell=1}\mathcal{N}\big( \mathbf{z}^{\ell,v}(t); \mu^{\ell,v}, \Sigma^{\ell,v} \big)}_{\text{Residual Energy}} + 
    \underbrace{ \Omega\Big( \Theta^v \Big) }_{\text{Synaptic prior}}  \label{eqn:rpc_vfe} \\
     &= \sum^L_{\ell=1} \sum_q \gamma_{v,q}  \mathcal{N}\Big(\mathbf{z}^{\ell,q}(t); \mu_C\big(\mathbf{z}^{\ell,v}(t), \mathbf{R}^{\ell,v,q}, \mathbf{A}^{\ell,v,q}\big), \Sigma^{\ell,v,q}_C \Big)  \nonumber \\ 
    & \quad + \sum^L_{\ell=1}\mathcal{N}\Big( \mathbf{z}^{\ell,v}(t); \mu\big(\mathbf{z}^{\ell-1,v}(t); \mathbf{W}^{\ell,v}\big), \Sigma^{\ell,v} \Big) \nonumber + 
    \sum_{p,i,j} \mathcal{N}\Big(\Theta^v[p]_{ij}; 0, \lambda_w\Big) 
    \nonumber 
\end{align}
where $\gamma_{v,q}$ is one for $q=v$ and $0 \leq \gamma_q < 1$ for $q \neq v$ (determined, in this study, based on a power kernel applied to the index coordinates of each stream when they are arranged in a simple chain-like topology; see Figure \ref{fig:mpc_prediction_pattern} and Supplementary Note 7 for details). 
The three terms of the energy depicted above can be understood as follows.  
%% cross-representation energy 
The cross-representation term dictates that, at time $t$, any non-sensory layer $\ell > 0$ with activity  $\mathbf{z}^{\ell,v}(t)$ in the $v$-th stream predicts the activity values $\mathbf{z}^{\ell,q}$ of the corresponding $q$-th target stream -- the prediction connections from stream $v$'s layer $\ell$ convey the mean  $\mu^{\ell,v,q}_C$ and covariance  $\Sigma^{\ell,v,q}_C$ of that layer's prediction of $\mathbf{z}^{\ell,q}$. 
%% residual energy 
The residual energy term captures the fact that each individual stream is hierarchically structured, where every layer $\ell$ of neuronal units attempt to minimize the prediction error between its activity and the prediction of this activity by layer $\ell+1$. 
%% synapse prior term 
The synaptic prior term is synaptic decay or, in other words, a zero-mean Gaussian prior (with standard deviation $\lambda_w$) placed over the $v$-th stream's plastic synapses represented by the parameters $\Theta^v$.\footnote{$\Theta^v$ is a tuple containing all of the synaptic efficacy matrices for the $v$-th MPC stream; $p$ retrieves the $p$-th synaptic parameter matrix from $\Theta^v$, whereas $i$ and $j$ index a particular synapse within the $p$-th matrix, i.e., $\Theta^v[p]_{ij}$ returns a scalar value.}  Finally, for the entire MPC architecture, the ``ensemble free energy'' would be the combination of all of its individual stream-level VFEs: $\mathcal{F}(\Theta) = \sum^V_{v=1} \mathcal{F}_v(\Theta^v)$. 
In Methods, we provide a detailed treatment of the exact neuronal dynamics (gradient flow) and Hebbian synaptic plasticity rules that result from our above model design and its VFE. 

%%%%%%%%%%%%%%%%%%%%%%%%%%%%%%%%%%%%%%%%%%%%%%%%%%%%%%%%%%%%%%%%%%%%%%%%%%%%%%%%%%%%%%%%%%%%
%%%%%%%%%%%%%%%%%%%%%%%%%%%%%%%%%%%%%%%%%%%%%%%%%%%%%%%%%%%%%%%%%%%%%%%%%%%%%%%%%%%%%%%%%%

\paragraph{Active perception through planned saccades.} %\paragraph{Epistemic glimpse planning.} 
In the MPC model designed for this study, the next location of where to look (i.e., the $\tau$-th glimpse $\mathbf{g}(\tau)$) is chosen in accordance to its ``saccade planner'', a free energy-driven process that engages in a reflex-based form of active perception. This planner allows the MPC oculomotor model to act such that it minimizes it long-term epistemic uncertainty and engage in a basic form of information foraging \cite{perrinet2014active,parr2021computational}. 
In effect, this planner operates by taking into account both the local structural changes computed by the LGN units (in effect, the LGN units provide a bottom-up response map) as well as its top-down prediction error (in terms of the model's ensemble energy $\mathcal{F}(\mathbf{g}(\tau);\Theta) = \mathcal{F}(\Theta)$) produced by the MPC circuit streams. At each glimpse $\mathbf{g}(\tau)$, the planner maintains a local history and a spatial inhibition-of-return map driven by the distance between the current center-point of fixation and other possible pixel coordinates to glimpse at, seeking to iteratively update its energy map of the image pixel space and ultimately select a region/point with highest uncertainty. %(or potential information gain). 
In the Methods, we describe the process in detail, including both a deterministic variant (i.e., deterministic ``epistemic saccade generation'') and stochastic version (i.e., ``stochastic foraging'') of this planner. 
Note that, in addition to maintaining and using a latent energy map to select the coordinates of the next glimpse, the MPC model progressively, after each saccade, builds a global latent code vector of the processed visual scene. This is done through the binding of encoded saccade coordinate information (inspired by grid-cells \cite{stensola2016grid}) and each glimpse's top-most neural activity vector. See Methods and Supplementary Note 6 for a full explication of the global latent code construction process. 
%\textcolor{red}{TODO: WRITE ABOUT INTEGRATION AND A BIT MORE FLUFF/HIGH-LEVEL HERE...} 
% In Methods, we provide a full implementation specification of the reflexive MPC saccade planning scheme. 

%%% experimental results
\subsection*{Meta-representational predictive coding learns representations useful for downstream tasks} 
% downstream probes

%%%%%%%%%%%%%%%%%%%%%%%%%%%%%%%%%%%%%%%%%%%%%%%%%%%%%%%%%%%%%%%%%%%%%%%
\begin{table}[tp!]
\begin{center}
\begin{tabular}{c | c | c | c | c} 
 \hline
   &  \multicolumn{2}{c|}{\textbf{MNIST}} & \multicolumn{2}{c}{\textbf{K-MNIST}} \\\
  & \textbf{ACC}  & \textbf{SSIM} & \textbf{ACC} & \textbf{SSIM}  \\
 %% units
   & (\%) & (no units) & (\%) & (no units) \\
 %% measurements
 \hline\hline
 BP-CNN \cite{lecun2004learning} (Impl.) & $99.02 \pm 0.18$ & -- & $93.09 \pm 0.55$ & -- \\
 BP-FNN \cite{glorot2011deep} (Impl.) & $98.04 \pm 0.03$ & -- & $90.57 \pm 0.11$ & -- \\
 \hline 
 I-JEPA \cite{garrido2024learning} & $90.82 \pm 0.02$ & $0.972 \pm 0.003$ & $80.82 \pm 0.01$ & $0.953 \pm 0.001$\\
 GPC-sig \cite{rao1999predictive,ororbia2022ngc} & $97.20 \pm 0.03$ & $0.828 \pm 0.013$ & $90.23 \pm 0.10$ & $0.791 \pm 0.001$ \\ % $91.97 \pm 0.03$ 
 %GPC-\emph{gelu} & $93.78 \pm 0.05$ & $-- \pm --$ & $78.83 \pm 0.32$ & $-- \pm --$ \\
 %\hline 
 GPC-\emph{nwta} & $96.62 \pm 0.20$ & $0.846 \pm 0.002$ & $90.25 \pm 0.14$ & $0.677 \pm 0.015$ \\ %$81.99 \pm 0.03$ 
 %\hline
 \hline
 GPC-\emph{fov} & $98.04 \pm 0.02$ & $0.979 \pm 0.001$ & $92.82 \pm 0.08$ & $0.944 \pm 0.001$ \\ %& $96.83 \pm 0.03$ & $-- \pm --$ & $79.95 \pm 0.09$ & $-- \pm --$ \\
 MPC-epi & $98.10 \pm 0.10$ & $0.974 \pm 0.001$ & $93.10 \pm 0.20$ & $0.910 \pm 0.005$ \\ %$0.908 \pm 0.004$ \\ 
 MPC-stoch & $97.57 \pm 0.13$ & $0.973 \pm 0.002$ & $92.05 \pm 0.21$ & $0.904 \pm 0.006$ \\
 \hline
 \addlinespace[0.5em] 
 %% measurements
 \hline 
 &  \multicolumn{2}{c|}{\textbf{ETH-80} (Leave-2-Out)} & \multicolumn{2}{c}{\textbf{NORB}} \\\
  & \textbf{ACC}  & \textbf{SSIM} & \textbf{ACC}  & \textbf{SSIM}  \\
 %% units
   & (\%) & (no units) & (\%) & (no units) \\
 \hline\hline 
 BP-CNN \cite{lecun2004learning} (Impl.) & $82.99 \pm 4.91$ & -- & $93.42 \pm 0.11$ & -- \\
 BP-FNN \cite{glorot2011deep} (Impl.) & $78.27 \pm 6.11$ & -- & $78.86 \pm 0.46$ & -- \\
 \hline 
 I-JEPA \cite{garrido2024learning}  & $69.09 \pm 6.20$ & $0.857 \pm 0.007$ & $74.68 \pm 0.01$ & $0.805 \pm 0.008$ \\
 GPC-sig \cite{rao1999predictive,ororbia2022ngc}  & $74.33 \pm 5.66$ & $0.832 \pm 0.004$ & $82.04 \pm 0.21$ & $0.790 \pm 0.004$ \\
 %GPC-\emph{gelu}  & $-- \pm --$ & $-- \pm --$ & $-- \pm --$ & $-- \pm --$ \\
 %\hline 
 GPC-\emph{nwta}  & $73.60 \pm 5.15$ & $0.780 \pm 0.005$ & $80.73 \pm 3.68$ & $0.793 \pm 0.002$ \\
 %\hline
 \hline
 GPC-\emph{fov}  & $78.48 \pm 4.47$ & $0.840 \pm 0.008$ & $87.09 \pm 0.07$ & $0.876 \pm 0.002$ \\
 MPC-epi & $84.02 \pm 5.23$ & $0.801 \pm 0.010$ & $88.78 \pm 0.30$ & $0.839 \pm 0.004$    \\ 
 MPC-stoch & $83.45 \pm 4.43$ & $0.788 \pm 0.011$ & $88.41 \pm 0.21$ & $0.832 \pm 0.006$ \\ 

 \hline %
\end{tabular}
\vspace{0.2cm}
\caption{ 
\small{
\textbf{Downstream generalization ability of self-supervised models.} 
Measurements of generalization accuracy (ACC, in terms of \%) and the reconstruction structural similarity index measure (SSIM, no units) 
%decoder mean squared error (Dec-MSE, in terms of nats)
of different types of predictive coding (PC) schemes (mean $\pm$ standard deviation reported for $10$ trials). BP-FFN is a supervised reference model, i.e., a backprop-trained sparse feedforward rectifier neural network \cite{glorot2011deep}, %(with hidden ReLU activations) 
and BP-CNN is an implementation of the classical (with no data augmentation) object-recognition CNN detailed in \cite{lecun2004learning}; both of these BP-based ANNs directly learned a mapping between inputs and labels. GPC is a fully-connected generative PC network, GPC-\emph{fov} is generative PC reformulated to work with our saccade sensory processing scheme, and MPC is our proposed meta-representational PC model (MPC). 
Under two both non-natural (MNIST, K-MNIST) and natural image (ETH-80, NORB) conditions, we explore MPC with two different configurations of its reflex-based saccade planning process: MPC with deterministic epistemic saccade generation (`MPC-epi') and MPC with stochastic foraging (`MPC-stoch'); 
see Methods for details on these two saccade planning schemes. 
I-JEPA is the encoder-only backprop model proposed in \cite{garrido2024learning} adapted to our study (to follow the same training process as well as have the same number of parameters to ensure fair comparison). 
}
}
\label{table:benchmarks}
\end{center}
\end{table}
%%%%%%%%%%%%%%%%%%%%%%%%%%%%%%%%%%%%%%%%%%%%%%%%%%%%%%%%%%%%%%%%%%%%%%%

\paragraph{Downstream evaluation of MPC latent codes.} The neural encoding framework that we have described thus far would be primarily useful for acquiring distributed representations of data that could then be used in subsequent statistical learning tasks, including those that make use of annotations/labels \cite{geiping2023cookbook}. Here, we examine the utility of MPC-acquired latent representations in the context of downstream tasks such as classification and reconstructing (decoding). We compare across eight models: six baseline/reference models and two MPC circuits on four datasets: 
\textbf{1)} the MNIST digit recognition database \cite{lecun1998mnist};  
\textbf{2)} the Kuzushiji-MNIST (K-MNIST) character recognition database \cite{clanuwat2018deep}; 
\textbf{3)} the NYU NORB natural image dataset \cite{lecun2004learning}; and, 
\textbf{4)} the ETH-80 natural image collection \cite{leibe2003analyzing} %\cite{leibe2003analyzing,chen2020covariance}  
(five-fold ``leave-two-objects-out'' task format). 
Of the six baselines, two are fully-supervised -- the backpropagation-trained feedforward network (BP-FNN) and the convolutional network (BP-CNN) -- and offer ideal (or soft upper bounds on) performance targets for downstream discrimination. The other four are SSL models: a comparable joint-embedding SSL model (I-JEPA) \cite{garrido2024learning} and three variants of self-supervised generative PC (GPC); one GPC model uses sigmoidal activations \cite{rao1999predictive,ororbia2022ngc} (GPC-sigm), another uses a competitive N-winner-take-all design (GPC-\emph{nwta}), and, finally, another model was structured to match the multi-stream design of our MPC circuits (GPC-\emph{fov}). %

\begin{figure}[!t]
    \begin{subfigure}[b]{0.595\textwidth}
    \centering
    \includegraphics[width=1.0\textwidth]{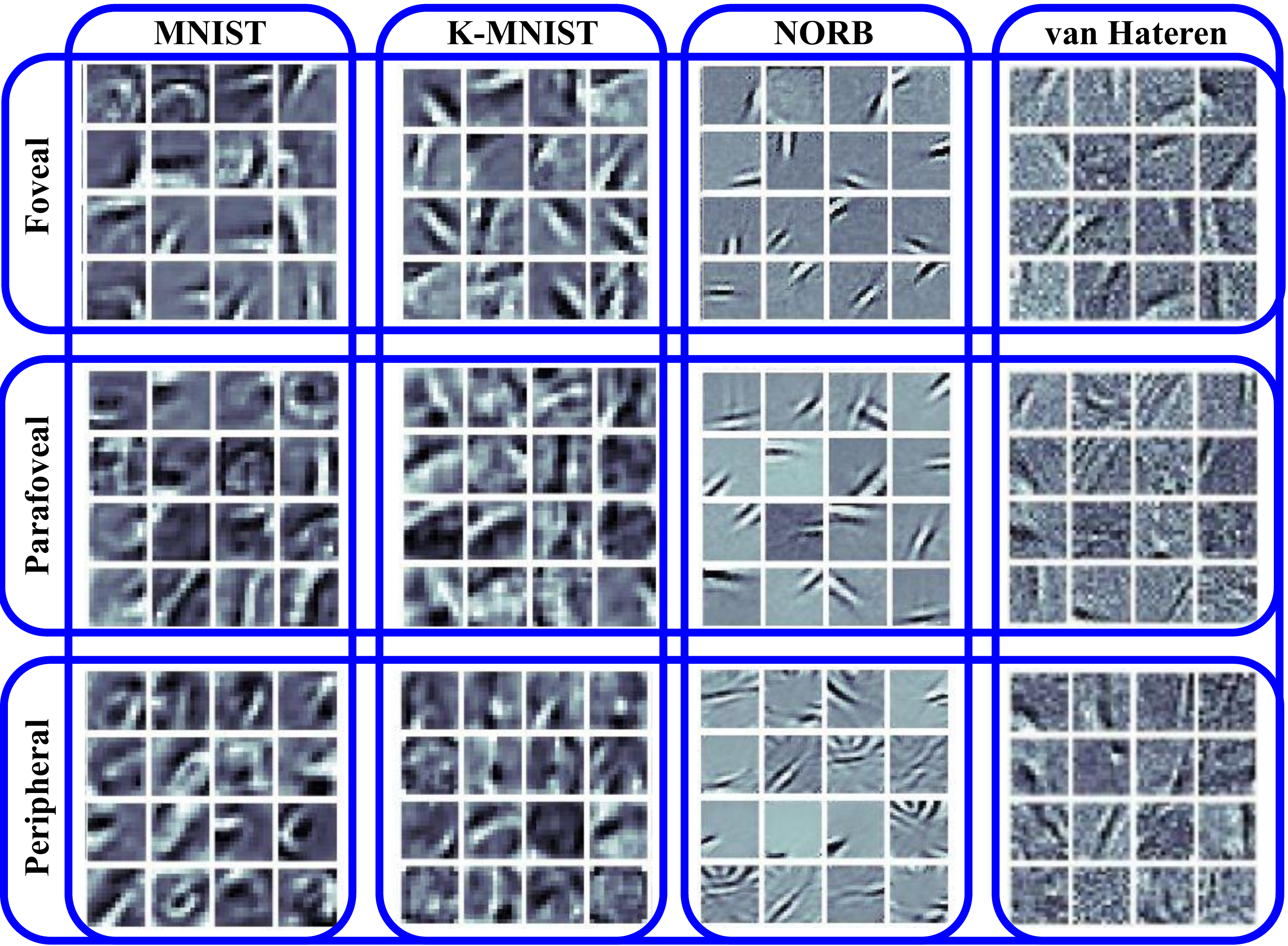}
    \caption{MPC receptive fields across datasets.}
    \label{fig:mpc_recfields}
    \end{subfigure}
     \begin{subfigure}[b]{0.4\textwidth}
            \centering
            \setlength{\tabcolsep}{5pt} 
            \begin{tabular}{cccc}
            \toprule
            % Row 1 Header: Column 1 is empty, then three 2-column wide headers
            & \multicolumn{3}{c}{\textbf{NORB}}  \\
            % Optional: Adds elegant visual separation bars under each grouped header
            \cmidrule(lr){2-4} 
            
            \midrule
            
            % --- Row 1 ---
            \adjustbox{valign=c}{\rotatebox{90}{{\footnotesize\textbf{Source}}}} 
            & \adjustbox{valign=c}{\includegraphics[width=1.4cm]{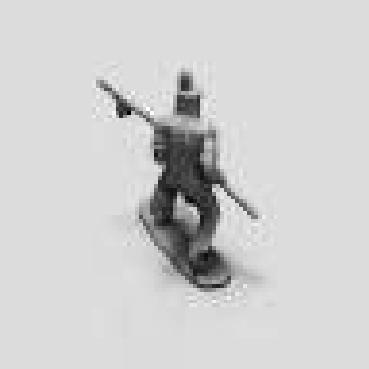}} 
            & \adjustbox{valign=c}{\includegraphics[width=1.4cm]{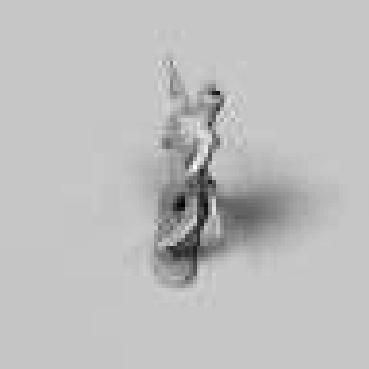}} 
            & \adjustbox{valign=c}{\includegraphics[width=1.4cm]{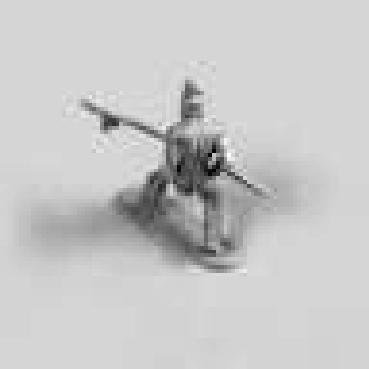}}  \\
            \cmidrule{2-4}
            
            % --- Row 2 ---
            \adjustbox{valign=c}{\rotatebox{90}{{\footnotesize\textbf{GPC}}}} 
            & \adjustbox{valign=c}{\includegraphics[width=1.4cm]{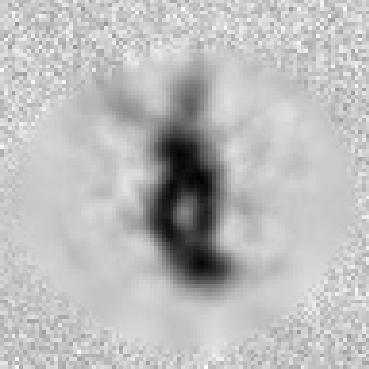}} 
            & \adjustbox{valign=c}{\includegraphics[width=1.4cm]{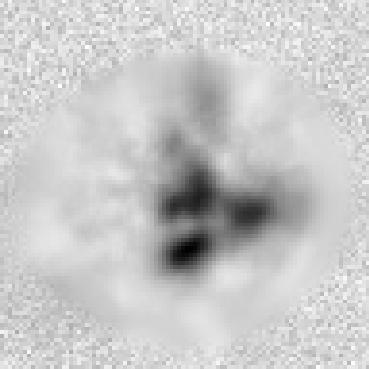}} 
            & \adjustbox{valign=c}{\includegraphics[width=1.4cm]{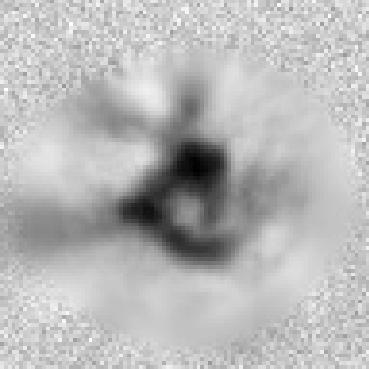}} \\
            \cmidrule{2-4}
            
            % --- Row 3 ---
            \adjustbox{valign=c}{\rotatebox{90}{{\footnotesize\textbf{I-JEPA}}}} 
            & \adjustbox{valign=c}{\includegraphics[width=1.4cm]{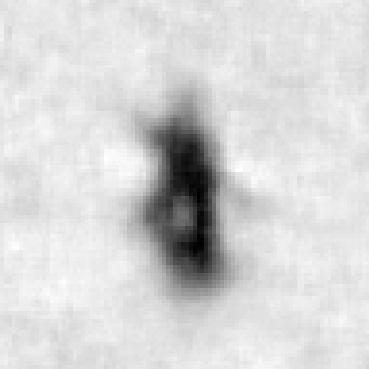}} 
            & \adjustbox{valign=c}{\includegraphics[width=1.4cm]{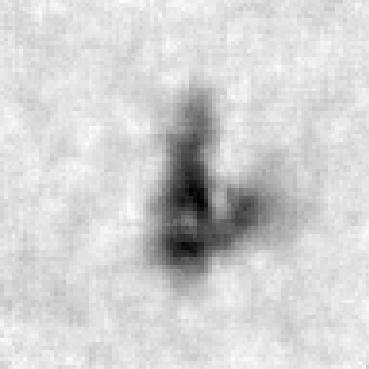}} 
            & \adjustbox{valign=c}{\includegraphics[width=1.4cm]{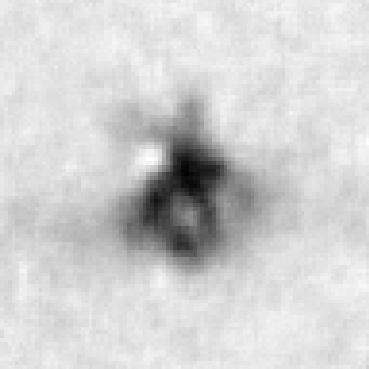}} \\
            \cmidrule{2-4}
            
            % --- Row 4 ---
            \adjustbox{valign=c}{\rotatebox{90}{{\footnotesize\textbf{MPC}}}} 
            & \adjustbox{valign=c}{\includegraphics[width=1.4cm]{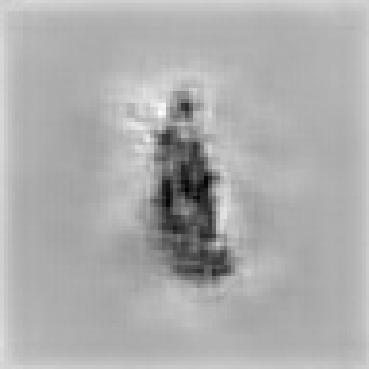}} 
            & \adjustbox{valign=c}{\includegraphics[width=1.4cm]{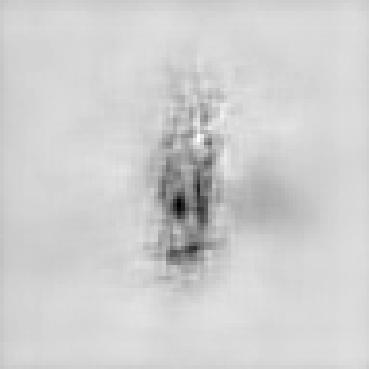}} 
            & \adjustbox{valign=c}{\includegraphics[width=1.4cm]{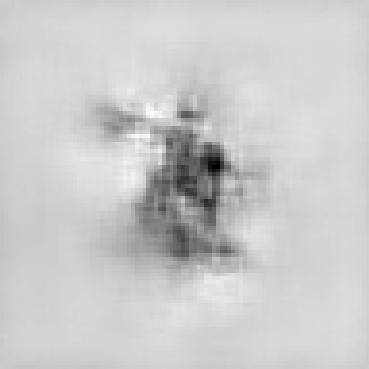}} \\
            \bottomrule
        \end{tabular}
        \caption{
        Downstream decoded NORB natural images.
        }
        \label{tab:downstream_reconstructions}
    \end{subfigure}

    \caption{ 
    \small{
    \textbf{Bottom-layer receptive fields acquired by and reconstructed patterns produced from MPC circuits.} 
    (Panel \textbf{a}) 
    Shown are sampled foveal, parafoveal, and peripheral receptive fields of the bottom-most neurons closest to the sensory input (i.e., layer $\ell = 1$) of a meta-representational predictive coding (MPC) encoder models, trained across four datasets, i.e., MNIST, K-MNIST, NORB, and a collection of van Hateren natural images  \cite{olshausen1997sparse,vanhaterenRealOptimalNeural1992}. 
    %Shown are the receptive fields for each of the four MPC encoder circuit's different neural streams: 
    Filters are shown, by row, as follows: 
    (First Row) one of the four foveal streams; 
    (Second Row) the parafoveal stream; and, 
    (Third Row) the peripheral stream. 
    %\textbf{Sampled downstream reconstruction of NORB natural images. } 
    (Panel \textbf{b}) 
    Shown are representative samples of the downstream decoder's input reconstruction (from frozen latent representations) of NORB test-set instances. 
    \textit{Row 1, `Source')} depicts original test-time images (containing detailed 3D object geometries, fine contours, etc.); 
    \textit{Row 2, `GPC')} depicts global/full-image predictive coding (GPC) samples; 
    \textit{Row 3, `I-JEPA')} samples of a trained-from-scratch joint-embedding predictive architecture (I-JEPA); and, 
    \textit{Row 4, `MPC')} samples of the proposed MPC. 
    Note that, for GPC, to ensure stability in the training process, input data was mean-centered prior to prediction and baseline mean values were add post-GPC predictions. For GPC, the resulting reconstructions indicate distorted object topologies and high-frequency pixel noise; I-JEPA samples show a collapse to smooth, low-frequency blurred outputs (discarding high-contrast structural edges); whereas MPC, through its active binding of foveated sensory primitives to grid-cell phase-centroids, is able to produce sharper, spatially coherent edge contours as well as global object geometry, thus allowing it match the authentic visual archetype of the unseen test patterns.
    }
    }
    \label{fig:patch_viz}
    \vspace{-0.5cm}
\end{figure}

The empirical results gathered from these experiments are summarized in Table \ref{table:benchmarks}. Specifically, we report the mean and standard deviation (over $10$ uniquely-seeded experimental trials) of the test-set accuracy (ACC; higher is better) for a downstream classification probe and the structural similarity index measure (SSIM; higher is better) for a downstream decoder/reconstruction probe for all models; see Methods for experimental setup details. 
Observe that, although all of the self-supervised models (generative and encoder-centric) do not exceed the performance of the best purely discriminative, fully-supervised reference model (BP-CNN), the proposed encoder-only MPC scheme gets quite close, with the MPC models offering competitive performance, notably outperforming the fully-supervised MLP (BP-FNN). MPC further outperforms I-JEPA and the GPC models in terms of KNN discriminative performance.  
In terms of reconstruction, we observe that MPC facilitates the effective learning of a downstream decoder, yielding SSIM scores that are better than the whole-image GPC models, although I-JEPA offers the best downstream SSIM along with GCP-\emph{fov} (it makes sense that GCP-\emph{fov} would offer excellent decoding ability given that it is trained directly to reconstruct input pixels). 
To further examine other differences in how each model affects downstream reconstruction, on NORB, we qualitatively compare several decoder samples to the original data patterns in Table \ref{tab:downstream_reconstructions}; we observe that MPC appears to provide more coherent, sharper/clearer reconstructions (from frozen latent codes) as compared to GPC and even I-JEPA. Given the reported measurements of Table \ref{table:benchmarks}, we observe that, overall, MPC circuits produce representations that facilitate high downstream classifier performance and good decoder output quality. 

Qualitatively, we next examined the receptive fields acquired by the neuronal units of one of our best-performing MPC circuits, i.e., MPC-epi, by visualizing them in terms of 2D images; the results of this examination are shown in Figure \ref{fig:patch_viz}. Note how the foveal receptive fields (the sampled ones in the first row of grids of $12 \times 12$ fields) look very similar to those acquired by classical PC/sparse coding models  \cite{olshausen1997sparse,rao1999predictive}, representing a variety of different possible strokes. These strokes are aggregated by higher-level layers of the scheme as it assembles higher-level representations of sensory input. 
The upper-level neuronal units, we further hypothesize, are learning a latent command structure (much as what was tested for in \cite{ororbia2022ngc}) that produces combinations of the low-level (level $1$) features -- implicitly learning contour detectors, complex weighted combinations that represent shape parts, etc. 
The parafoveal and peripheral receptive fields, i.e., the last two rows of $12 \times 12$ squares of receptive fields, appear to encode larger, more zoomed-out versions of (or maps that contain) strokes/atomic features or possibly low-resolution objects/chunks of objects (e.g., digits/Kanji characters, NORB \cite{lecunLearningMethodsGeneric2004NORB} objects). In the rightmost column, we further present the receptive fields (of each stream type) of an MPC trained on Van Hateren natural images \cite{vanhaterenRealOptimalNeural1992}; %\cite{vanhaterenRealOptimalNeural1992,van1998independent}; 
here we see that MPC is capable of extracting useful features even from complex, natural scene imagery. 

% Beyond examining receptive fields, we visualized the relationship between the latent codes that a trained MPC circuit (MPC-epi) produced across various test-sets. The results of this qualitative analysis used t-SNE \cite{van2008visualizing} to visualize the latent codes formed with respect to the test-set samples; the results of this visualization is presented in Figure \ref{fig:latent_viz}. Notice that, despite never having access to the labels nor using any kind of supervisory signal, the MPC codes for MNIST, K-MNIST, and NORB tend to cluster rather well (albeit a bit noisily in some spots), yielding distinct major representational groupings with each corresponding (item/object) class 
% (labels were only used to color the t-SNE mapped latent code points in the t-SNE sub-figures). For K-MNIST, we see that MPC latent codes form groupings as well but there many more groups/clusters or ``sub-groupings''  than the ten identified Kanji character categories; this behavior makes sense, given that the Kanji characters are more complex and exhibit a higher degree of variety than the handwritten digits in the MNIST database. For NORB, we see that clusters representing the five categories emerge though two of these still overlap; this is indicative of the challenge offered by a natural image dataset. 

%%%%%%%%%%%%%%%%%%%%%%%%%%%%%%%%%%%%%%%%%%%%%%%%%%%%%%%%%%%%%%%%%%%%%%%
%%%%%%%%%%%%%%%%%%%%%%%%%%%%%%%%%%%%%%%%%%%%%%%%%%%%%%%%%%%%%%%%%%%%%%%

\subsection*{Encoder-centric predictive coding facilitates sample efficiency and zero-shot representational transfer} % sample efficiency and zero-shot representational transfer

\begin{figure}[!t]
    %%
    % sample efficiency
     \centering
     \begin{subfigure}[b]{0.485\textwidth}
         \centering
         \includegraphics[width=0.85\textwidth]{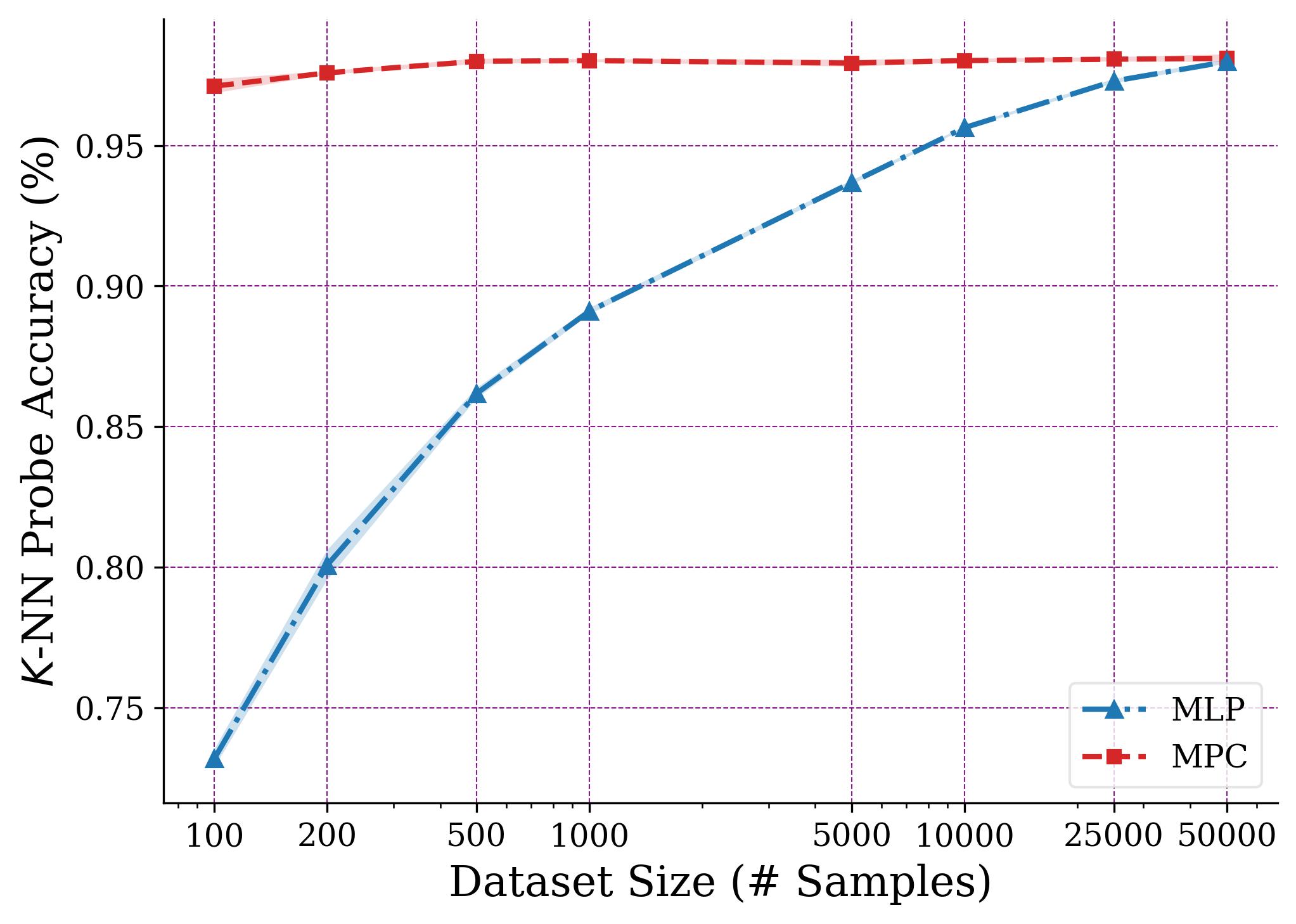}
         \caption{KNN Accuracy vs. MNIST Sample Size.}
         \label{fig:mnist_subset_exp}
     \end{subfigure}
     \begin{subfigure}[b]{0.485\textwidth}
         \centering
         \includegraphics[width=0.85\textwidth]{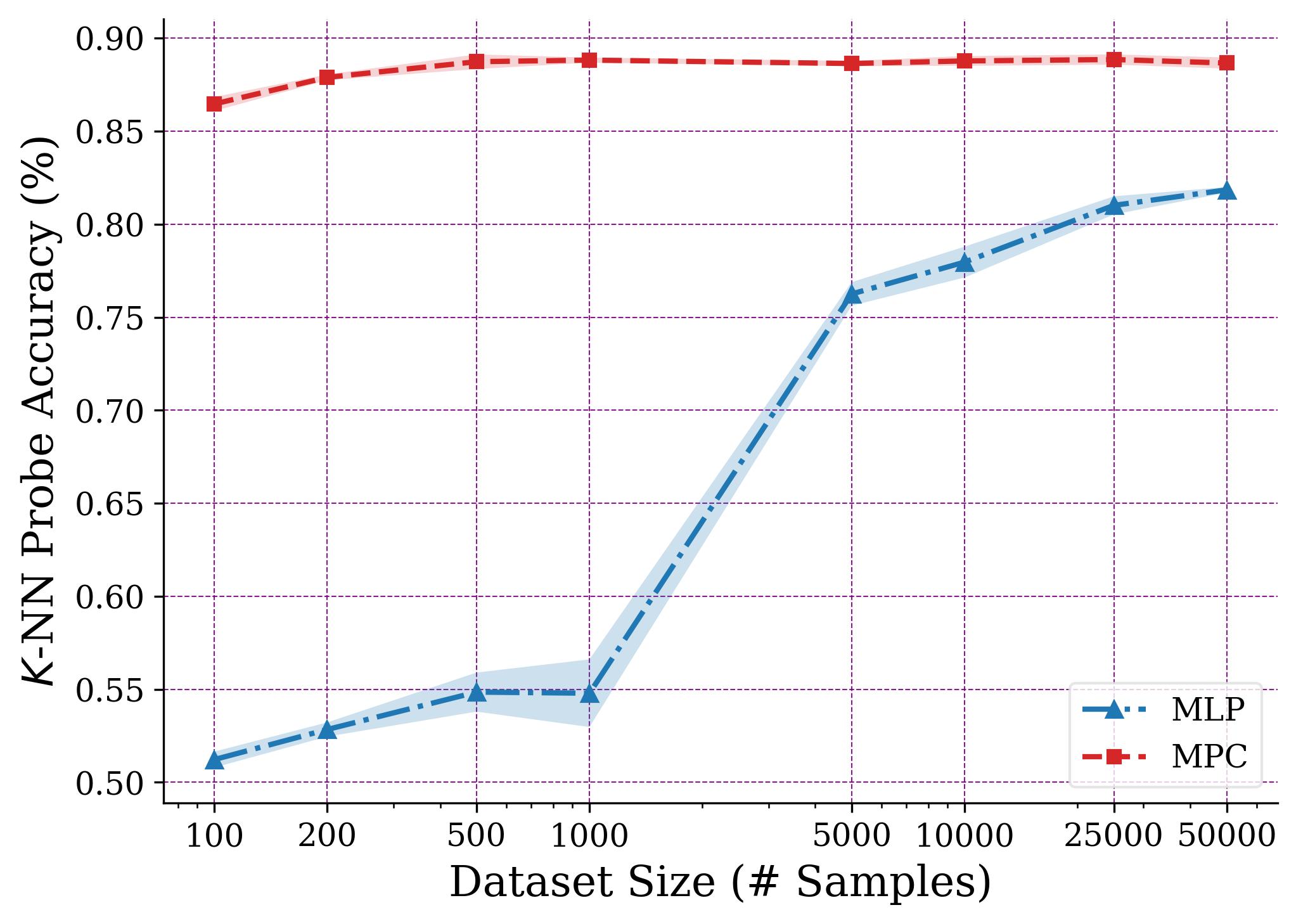}
         \caption{KNN Accuracy vs. NORB Sample Size.}
         \label{fig:norb_subset_exp}
     \end{subfigure}
     %% 
     % glimpse budget
     \begin{subfigure}[b]{0.485\textwidth}
         \centering
         \includegraphics[width=0.85\textwidth]{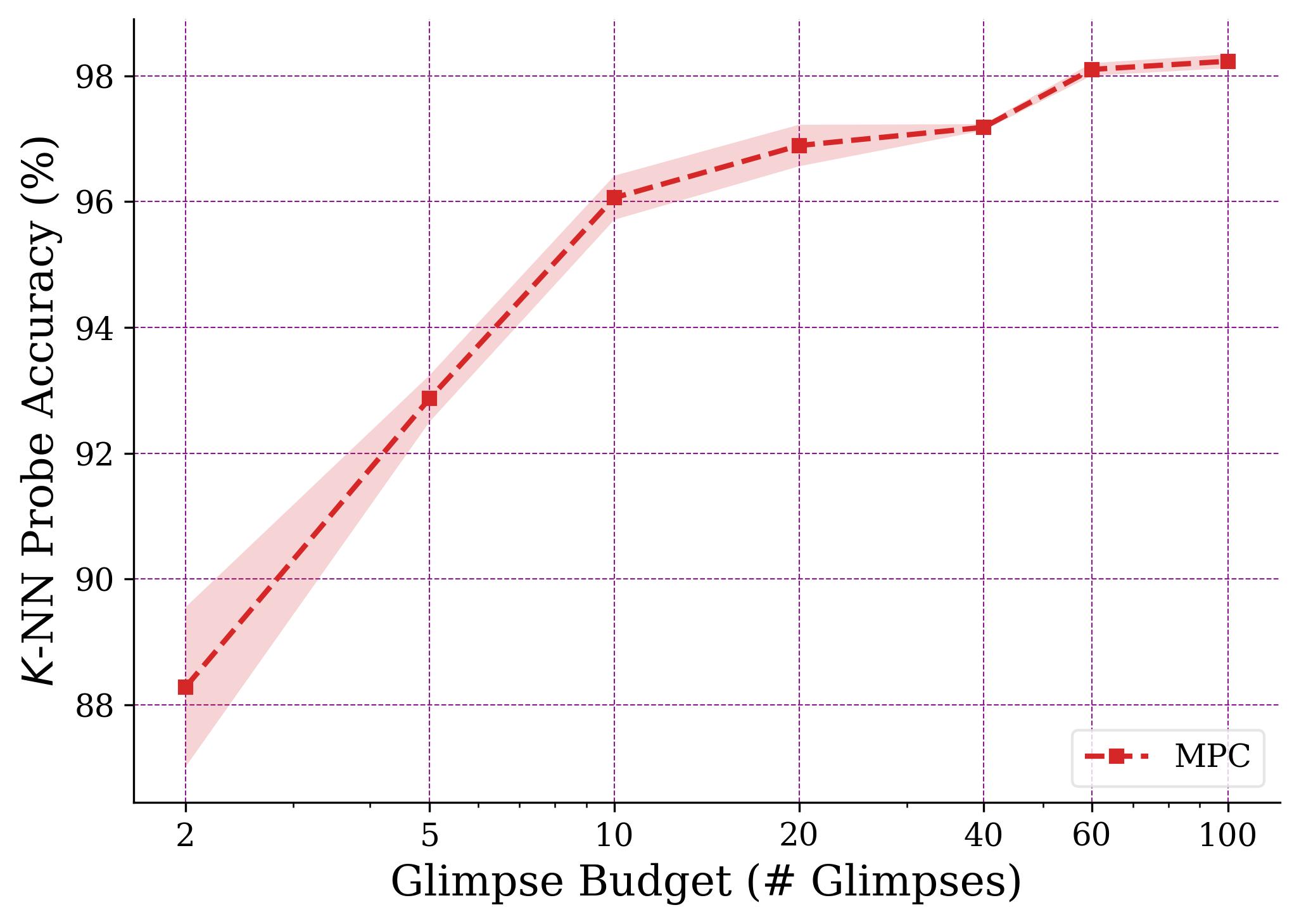}
         \caption{KNN Accuracy vs. MNIST Glimpse Budget.}
         \label{fig:mnist_glimpse_budget_exp}
     \end{subfigure}
     \begin{subfigure}[b]{0.485\textwidth}
         \centering
         \includegraphics[width=0.85\textwidth]{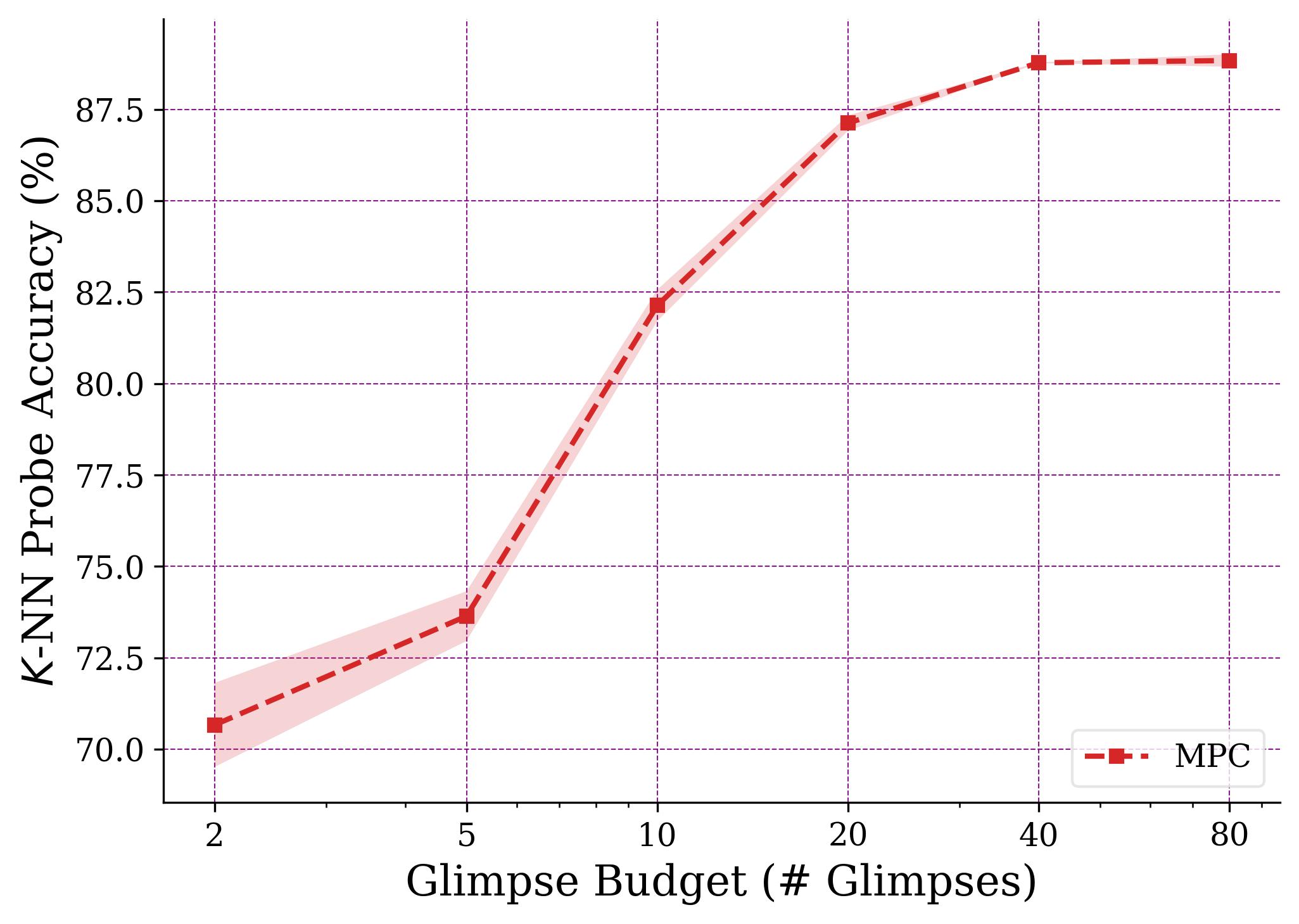}
         \caption{KNN Accuracy vs. NORB Glimpse Budget.}
         \label{fig:norb_glimpse_budget_exp}
     \end{subfigure}
    \begin{subfigure}[b]{0.95\textwidth}
        \centering
        \includegraphics[width=0.675\textwidth]{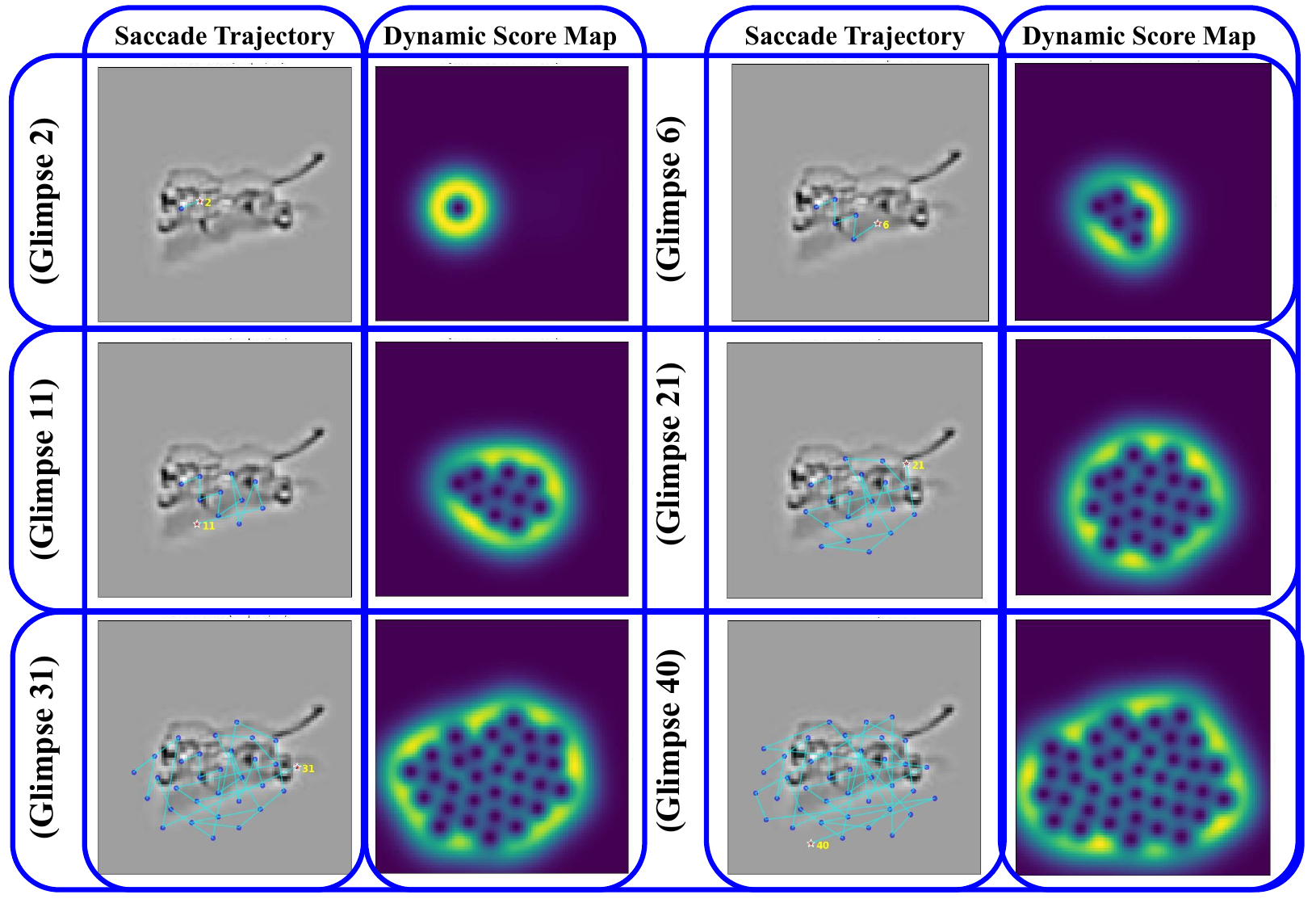}
        \caption{A sampled MPC saccade trajectory over a NORB image.}
        \label{fig:mpc_saccade_trajectory}
    \end{subfigure}
    \caption{ 
    \small{
        \textbf{Sample Efficiency on Image Databases.} Here, we plot test-set classification accuracy ($10$-trial mean values; accuracy values normalized to the range of $[0,1]$) on MNIST (Panel \textbf{a}) and NORB (Panel \textbf{b}) of MPC (MPC-epi) against the reference BP-FNN (MLP) as a function of the number of samples used to train each model. %\emph{Note}: x-axis was plotted on a logarithmic scale to help visualize the generalization curve. 
        %\textbf{Downstream generalization as a function of glimpse budget.} 
        We furthermore plot, for our best MPC model, the downstream test-set classification accuracy ($10$-trial mean values) %; accuracy values normalized to the range of $[0,1]$) 
        on MNIST (Panel \textbf{c}) and NORB (Panel \textbf{d}) of MPC (MPC-epi) as a function of the number of the maximum of saccades it is allowed to when constructing its global representations of sensory inputs (i.e., its ``glimpse budget'')
        %the ``glimpse budget'' (i.e., number of maximum allowed glimpses that the MPC is allowed to take when processing an image). 
        %\textbf{A sampled MPC saccade trajectory over a NORB image.} 
        (Panel \textbf{e}) 
        Depicted here is one sampled saccade trajectory taken by an MPC model on a (post-LGN-processed) test-image after it has been trained on NORB. Discrete snapshots (its particular glimpse step indicated by an integer between $1$ and $40$, the maximum number of glimpses) within this trajectory are taken, where the order proceeds left-to-right, then top-to-bottom. Paired alongside each glimpse image is the dynamic (free-energy) score map maintained by MPC as it forages for information from the image (brighter/higher intensity colors indicate higher free energy regions that MPC will likely explore in its next planned glimpses). 
        \emph{Note}: For plots of Panels \textbf{a}-\textbf{d}, the x-axis was plotted on a logarithmic scale to help visualize curves. 
        }
    }
    \label{fig:mpc_sample_efficiency_budget}
    \vspace{-0.5cm}
\end{figure}

In Figures \ref{fig:mnist_subset_exp} and \ref{fig:norb_subset_exp}, 
%Figure \ref{fig:mpc_sample_efficiency_budget} (Panels \textbf{a} and \textbf{b}), 
we show the results of a test conducted to examine MPC's sample efficiency on MNIST (non-natural images) and NORB (natural images). Specifically, we re-fit MPC and the reference BP-FNN on differently-sized training datasets -- dataset sizes 
\footnote{Samples selected from the originally training database were randomly sampled without replacement.}  $|\mathcal{D}|$ included ${\{100,200,500,1000,5000,10000,25000,50000\}}$ -- and then re-evaluated their generalization accuracy on the full test-set (mean values over $10$ trials are plotted). Notice that, as the number of available samples declines towards $100$ for both non-natural and natural image cases, the supervised BP-FNN's generalization degrades significantly (as expected for a learning scheme that requires labels) whereas MPC's downstream performance remains relatively consistent. 
We hypothesize that MPC's sample efficiency likely comes from the fact that it treats each data point as a sort of ``mini sensorium'' via its saccade-driven sampling process. This means that MPC models work to extract reusable simpler features, e.g., strokes/arcs and object chunks, that appear to more readily generalize (in MPC, lower layers capture these atomic features whereas upper layers produce weighted combinations of lower-level atomic features).

%%%%%%%%%%%%%%%%%%%%%%%%%%%%%%%%%%%%%%%%%%%%%%%%%%%%%%%%%%%%%%%%%%%%%%%%%%%%%%%%%%%%
%% sample efficiency figure - KNN probe versus sample size
%%%%%%%%%%%%%%%%%%%%%%%%%%%%%%%%%%%%%%%%%%%%%%%%%%%%%%%%%%%%%%%%%%%%%%%%%%%%%%%%%%%%

%%%%%%%%%%%%%%%%%%%%%%%%%%%%%%%%%%%%%%%%%%%%%%%%%%%%%%%%%%%%%%%%%%%%%%%
%% cross/zero-shot transfer results
\begin{table}[tp!]
\begin{center}
\begin{tabular}{c | c | c | c | c} 
 \hline
   &  \multicolumn{2}{c|}{\textbf{K-MNIST}$\Rightarrow$\textbf{MNIST}} & \multicolumn{2}{c}{\textbf{MNIST}$\Rightarrow$\textbf{K-MNIST}} \\\
  & \textbf{ACC} (\%) & SSIM (no units) & \textbf{ACC} (\%) & SSIM (no units) \\
 %% measurements
 \hline\hline
 MPC &  $97.53 \pm 0.12$ & $0.968 \pm 0.002$ & $92.75 \pm 0.05$ & $0.918 \pm 0.001$ \\
 % MNIST-MPC & $--$ & $--$ & $92.75 \pm 0.05$ & $0.918 \pm 0.001$ \\
 % K-MNIST-MPC & $97.53 \pm 0.12$ & $0.968 \pm 0.002$ & $--$ & $--$\\
\hline 
&  \multicolumn{2}{c|}{\textbf{t-SNE Visualization}} & \multicolumn{2}{c}{\textbf{t-SNE  Visualization}} \\
\hline
&  \multicolumn{2}{c|}{\includegraphics[width=0.315\textwidth]{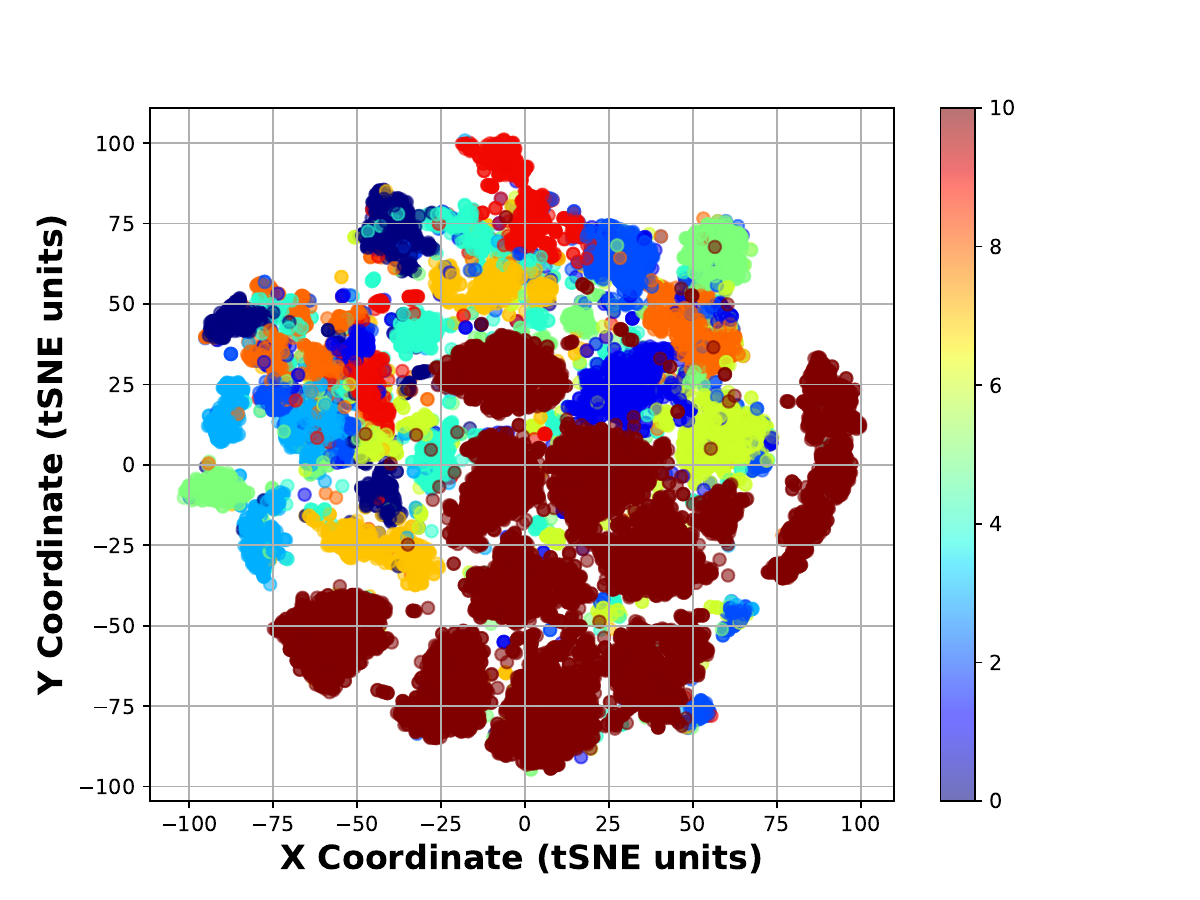}} & \multicolumn{2}{c}{\includegraphics[width=0.315\textwidth]{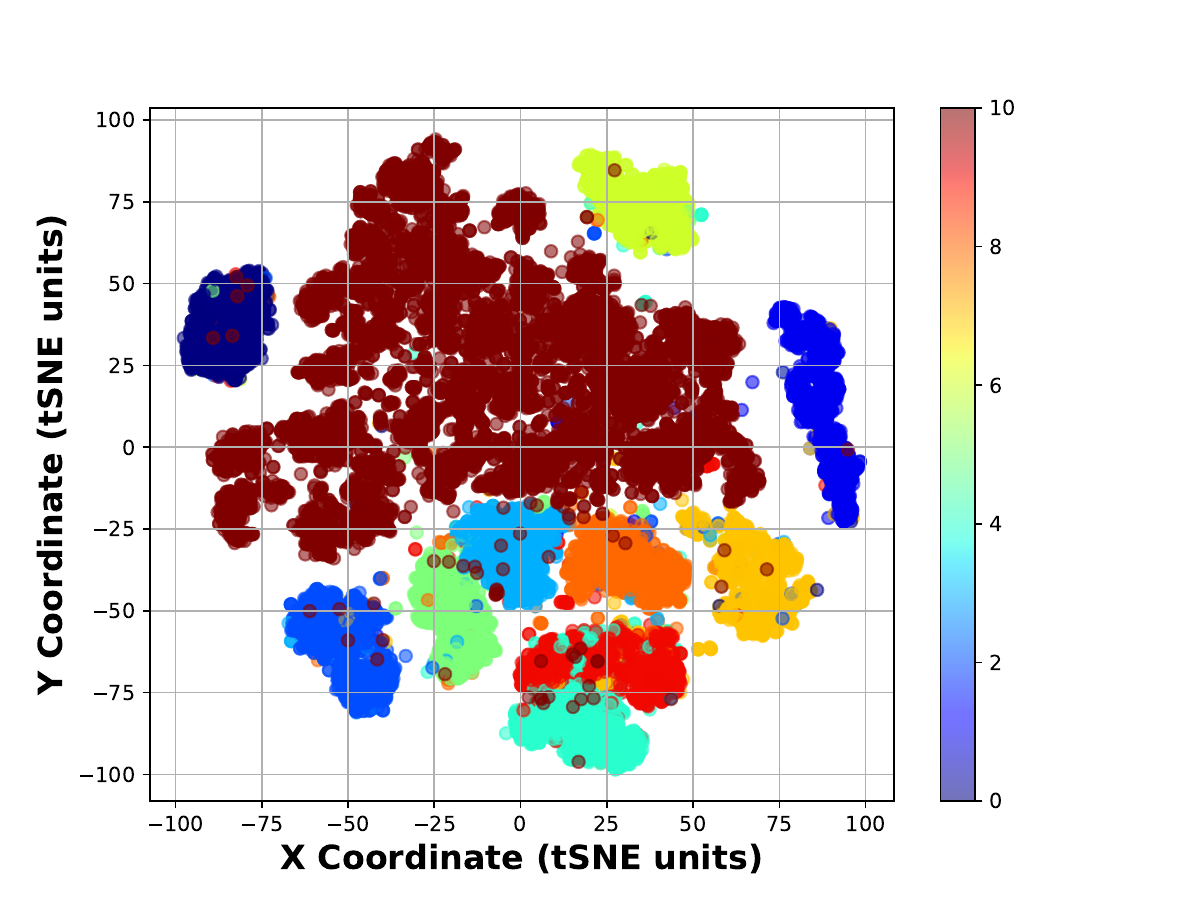}} \\
\hline %
\end{tabular}
\vspace{0.2cm}
\caption{ 
\small{
\textbf{Cross-dataset zero-shot generalization.} 
Here, we evaluate and report ($10$ trial mean \& standard deviation) downstream accuracy (ACC) and decoding ability (SSIM) to characterize the unsupervised domain generalization capability of MPC, with no cross-data fine-tuning. 
(Left; K-MNIST $\Rightarrow$ MNIST) downstream performance (KNN accuracy and decoder fidelity) when the MPC encoder is trained only on K-MNIST and zero-shot evaluated on the MNIST test set. 
(Right: MNIST $\Rightarrow$ K-MNIST) downstream performance when the MPC encoder is trained exclusively on K-MNIST and zero-shot evaluated on the K-MNIST test set. 
Underneath each set of measurements, we show a 2D t-SNE projection of the test-set manifolds across the $10$ sources classes with, crucially, a distinct $11$-th out-of-distribution category (rendered in red-brown) to visualize the spatial integration of the target/transfer dataset. Quantitatively and qualitatively, it appears that MPC uncovers more universal geometric primitive information (e.g., related to strokes, edges, orientations, etc.) that remains rather stable regardless of semantic data boundaries.
}
}
\label{tab:cross_data_transfer_results}
\end{center}
\end{table}
%%%%%%%%%%%%%%%%%%%%%%%%%%%%%%%%%%%%%%%%%%%%%%%%%%%%%%%%%%%%%%%%%%%%%%%

Next, given the MPC encoder's goal of producing general latent activity responses of visual sensory input, in Table \ref{tab:cross_data_transfer_results}, we present the results of an additional experiment investigating MPC's ability to conduct zero-shot representational generalization. Empirically, we find that our best-performing MPC model (MPC-epi) is able to meaningfully encode even data patterns that it has never seen before, i.e., it is capable of producing MNIST latent codes that cluster even though it was only trained on K-MNIST patterns and vice versa (as seen in the bottom-row t-SNE plots). The zero-shot KNN accuracy of the MPC latent codes on MNIST (after training on K-MNIST) are less than a percentage point lower than when trained and evaluated directly on MNIST; the same holds for K-MNIST (when trained on MNIST). Similarly, the reconstruction SSIM is only slightly lower than when trained directly (for both MNIST and K-MNIST).  

%\paragraph{Assembling representations through the MPC saccade process.} 
Finally, an interesting feature of our MPC model is that it represents stimuli by encoding portions of input across saccades. This means that multiple glimpses of the input are needed to obtain a decent ``bigger picture'' encoding. To investigate how the number of glimpses (a ``glimpse budget'') affects an MPC scheme's ability to form useful representations, we measure the performance of the model -- in terms of classification accuracy -- in response to the number of saccades allowed. In Figures \ref{fig:mnist_glimpse_budget_exp} and \ref{fig:norb_glimpse_budget_exp}, we examine the generalization of MPC (measured in terms of downstream, test-set classification accuracy) as a function of the number of glimpses that it is allowed to take (up to a maximum of $80$-$100$) when processing sensory input. 
Empirically, we notice that generalization improves with a larger glimpse budget, up to $>98$\% on average for MNIST ($100$ glimpses) and $>88$\% on average for NORB ($80$ glimpses). 
Furthermore, in Figure \ref{fig:mpc_saccade_trajectory}, we provide a qualitative sample of the MPC's saccade process (``MPC-epi'') to observe the trajectory it takes (budget of $40$ glimpses) when processing a natural image stimulus. 

\section*{Discussion}

%% conclusions
In this study, we proposed a new formulation of predictive coding, \emph{meta-representational predictive coding} (MPC), which is a neuroscience-informed form of self-supervised learning (Neuro-SSL) in service of acquiring distributed representations of sensory input. Our framework, grounded in the free energy principle, inverts the standard premise of PC from one of adapting a generative model to explain raw sensory inputs to an encoder-centric scheme where representations of distinct data features predict each other. Empirically, our results demonstrate that casting neuronal dynamics and synaptic plasticity as message passing -- induced by within and between stream prediction -- offers a mechanistic explanation of how representations of sensory stimuli might emerge. In the context of visual perception, MPC is instantiated by an architecture of visual streams that are each concerned with processing foveal/parafoveal (high-resolution) or peripheral (low-resolution) sensory data features. The resulting neuronal dynamics are driven by intra- and inter-stream message passing of predictions and prediction errors. 

Note that the resulting framework engages with the relatively challenging problem of self-supervised learning (SSL) of representations, side-stepping the need for positive and negative data samples (as in contrastive SSL schemes, which require a process for producing negative samples to pair with positive data points) through cross visual stream prediction, showing how abstract encodings might emerge using only predictions and prediction error alone (as opposed to schemes that use similarity metrics and complementary views of data through random transformations). In  Supplemental Notes 2 and 13, we provide a detailed review of the space of SSL and related neuroscience/biomimetic schemes to further contextualize MPC and its implications as well as limitations (and future directions to explore). 
Our experimental simulations show that self-supervised MPC was capable of learning representations that were useful for downstream discriminative learning and downstream reconstruction, without requiring great quantities of data and even exhibiting promising zero-shot generalization capabilities. Simulation results demonstrated that MPC was competitive with standard state-of-the-art PC and backprop-trained supervised learning, even though the framework never uses label information nor does it need to predict high dimensional sensory data (e.g., pixels). 
%Future efforts should investigate how the various biological mechanisms under-pinning MPC circuitry, such as its mechanisms for inducing lateral competition or topology-weighting of its inter-stream predictions of innovations, aid in mitigating potential dimensional collapse of the underlying neuronal activities. 

%%%%%%%%%%%%%%%%%%%%%%%%%%%% 
\section*{Methods} %% does not count towards 3500 word count
%\subsection*{Experimental Setup}

%\paragraph{Simulation setup and datasets. } 
\subsection*{Datasets} 
To demonstrate the efficacy of our MPC Neuro-SSL framework, we simulate models across four datasets of increasing complexity: 
\textbf{1)} MNIST \cite{lecun1998mnist};  
\textbf{2)} the Kuzushiji-MNIST (K-MNIST)  database \cite{clanuwat2018deep}; 
\textbf{3)} NYU NORB \cite{lecun2004learning}; and, 
\textbf{4)}  ETH-80 \cite{leibe2003analyzing}. %\cite{leibe2003analyzing,chen2020covariance}. 
MNIST and K-MNIST contain gray-scale $28\times28$ images from $10$ categories; NORB contains $96\times96$ images from $5$ categories and ETH-80 contains $256\times256$ images (down-scaled to shape $64\times64$) over $8$ classes. MNIST contains images of handwritten digits while Kuzushiji-MNIST is a challenging drop-in replacement for MNIST, containing images depicting hand-drawn Japanese Kanji characters; in K-MNIST, each class (out of $10$ classes) corresponds to the character's modern hiragana counterpart. The only pre-processing applied to the images in all of these datasets was to normalize the pixel intensities to lie in the range of $[0,1]$. Further note that, whenever an image patch is extracted for patch-level models (GPC-\emph{fov} and MPC circuits), we only center it (i.e., subtract the mean value of patch from the group of pixels). 
For the relevant models (GPC-\emph{fov} and MPC), we process each sample pattern according to the scheme described in Methods (in ``Sensory stimuli processing details''). 

%For NORB, we use all stereo-views (yielding a training dataset of $48,600$ patterns and a test-set of $48,6000$ samples) and for ETH-80 we constructed a difficult ``leave two objects out'' multi-fold data benchmark (over five folds). 
For the NORB dataset, we utilize the binocular stereo-views, yielding a training set of $48,600$ stereo-pair patterns and an independent test-set of $48,600$ stereo-pair patterns. The NORB dataset serves as an ideal benchmark because it isolates pure 3D object geometry under varying elevations, azimuths, and lighting directions against uniform backgrounds. This control allows us to evaluate the network's intrinsic capacity to learn viewpoint and lighting invariant characteristics through active, binocular foveated trajectories without the confounding effects of natural scene clutter. 

For the ETH-80 dataset, we construct a strict, five-fold ``leave-two-objects-out'' cross-validation benchmark. 
Unlike standard random image data splits that inadvertently place varying angles of the same object instance across both training and test subsets, the ``leave-two-objects-out'' protocol enforces strict category-level geometric generalization. By holding out entire physical object topologies and all $41$ of their respective viewpoints per fold \cite{leibe2003analyzing}, any given model is prevented from exploiting local pixel-coordinate ``shortcuts''; this setup forces a model to demonstrate true zero-shot abstraction on entirely novel physical structures. 
For the van Hateren natural images used to visualize filters on complex natural scenes in Figure \ref{fig:patch_viz} (which do not come with discrete labels), we used the same (pre-)processing as in the case of NORB; the van Hateren dataset, often used for models of unsupervised efficient coding and early visual receptive field development, contains thousands of varied, high-resolution images -- in our figure, we only used the same subset of images employed in \cite{olshausen1997sparse,rao1999predictive}. 

\subsection*{Model baseline descriptions}

%\paragraph{Model baselines.} 
We compare several circuit models (with $L=3$ layers) in our experimental simulations: 
\textbf{1)} generative predictive coding (GPC) circuits that process the entire input image (the more traditional PC model), including a full-image version of the classical model in \cite{rao1999predictive,ororbia2022ngc} (GPC-sigm)  
%, a variant of this model using ReLU activations GPC-\emph{relu}), 
and a variant of this GPC model using an N-WTA activation (GPC-\emph{nwta}) with the number of winners scaled to match the total number of winners across all neural columns in the GPC-\emph{fov} and MPC models; 
\textbf{2)} the GPC-\emph{fov} described earlier, which processes the same sensory information as our MPC architecture; and,  
\textbf{3)} two variants of our proposed MPC circuit (MPC-epi and MPC-stoch; the latter of which uses stochastic foraging). 
Furthermore, we report performance of a BP-centric encoder-only SSL model, i.e., the joint-embedding architecture known as (I-)JEPA \cite{assran2023self,garrido2024learning}. 
For the two MPC circuits, we study one possible cross-stream pattern (we remark that many others are possible), as depicted in Figure \ref{fig:mpc_prediction_pattern}. 
Note that, even though the GPC models work in an self-supervised fashion, they are all ``decoder-centric'' whereas all MPC circuits are ``encoder-centric''. Our experimental goal is to determine if our MPC scheme offers representations as useful (or possibly better than) generative PC models, demonstrating that we can effectively construct Neuro-SSL encoder models that acquire useful distributed representations of sensory input data without having to predict raw, low-level data features (e.g., pixel values). 

\paragraph{Experimental task and evaluation setup.} We train all models on each database for up to $100$ epochs\footnote{The patch-level models were trained in a streaming/episodic fashion; a mini-batch was sampled, randomly with replacement, from the dataset in one iteration. All such models were trained for $20,000$ iterations.} with gradient ascent (using mini-batches of length $200$). The learning rate of the gradient ascent optimization of parameters was tuned for each model using the validation subset of each database (we generally found that the iterative patch-level models preferred higher rates, whereas whole-image models worked better with lower rates). Unless stated otherwise, all saccade-driven models, i.e., GPC-\emph{fov} and MPC, were given a glimpse budget between $K=15$-$60$ saccades (it was found that training could make use of smaller budgets, e.g., $K=15$, whereas test-time inference benefited from larger budgets, e.g., $K=60$). 

To assess each model's efficacy, we train each under the same experimental conditions and data. We allow each to process the data and adapt parameters according to their specific dynamics and plasticity mechanisms. Once a model has completed its adaptation (parameter-fitting) phase, we fix its synaptic connection strengths (i.e., we disable its plasticity) and then allow it to process the training data, validation data, and test data once, extracting its latent representation for each data sample. If a model iteratively processes one input (e.g., the GPC-\emph{fov} and MPC models), we construct their resultant global latent codes in accordance with the process explained in the sub-section ``Latent glimpse path integration'' (also see Supplementary Note 5). %concatenated the sub-representations it produces across the $K$-length saccade trajectory.  
The resulting representations of the data samples are then used in the two following down-stream analyses:
\begin{itemize}[noitemsep,nolistsep]
    \item A K-nearest neighbor (K-NN) classifier ($K=3$ neighbors with a $p=2$ Minkowski distance) is fit to the latent codes of the training set (with validation latent codes used for hyperparameter-tuning) and then evaluated on the test-set latent codes. We report the test-set error measurements in Table \ref{table:benchmarks} and compare performance against a reference, supervised backprop-trained MLP (BF-FNN) and an object-recognition convolutional neural network  \cite{lecun2004learning} (BP-CNN). 
    \item A three-layer (de)convolutional decoder with Gaussian error linear units (GeLUs), trained with gradient descent (using Adam to apply the gradients of an Elastic-net loss function) was retro-fit to the training-set latent codes of our MPC scheme (the decoder was also tuned using validation-set latent codes). 
    This decoder's reconstruction efficacy was evaluated using test-set latent codes. We report mean squared error and compare the decoder's down-stream reconstruction against the natural reconstruction ability of the full-image GPC circuits. % and PSNR?
\end{itemize}

\subsection*{Meta-Representational Predictive Coding Procedures}

% \subsection{Eye Structure and Movements: Sensory Stimuli Processing Details}
% \label{sec:sensory_saccades}
\paragraph{Eye structure and movements: sensory stimuli processing details. }
Formally, for our glimpse processing scheme, we treat an observation at time $t_g$, i.e., $\mathbf{o}(t_g) \in \mathbb{R}^{O_H \times O_W \times Z}$ (a single static image or a frame sampled from a video; $t_g$ marks global time in milliseconds, $Z$ denotes number of channels and $O_H$ and $O_W$ denotes the image's height and width in pixels, respectively), as a small temporary environment from which our neural models extract a fixed-length trajectory of glimpses produced as the result of saccades, either randomly generated or produced by an active reflex process (or, in future work, a motor-control policy \cite{rao2023active}), yielding the sequence $\mathcal{S} = \Big\{ \big(\mathbf{g}(0), \mathbf{a}(0)\big), \big(\mathbf{g}(1), \mathbf{a}(1)\big),..., \big(\mathbf{g}(k), \mathbf{a}(k)\big), ... , \big(\mathbf{g}(K), \mathbf{a}(K)\big) \Big\}$ where $\mathbf{g}(k)$ is the $k$-th saccade-driven ``glimpse'' of the sensory input, $\mathbf{a}(k) \in [-1,1]^{2 \times 1}$  is a saccadic action or x-y coordinate vector recording the chosen center of the fixation-point of glimpse $\mathbf{g}(k)$, and $K$ is the maximum number of steps taken.

Each glimpse vector $\mathbf{g}(k)$ is made up of several groups (pixel patches) sampled from the observation $\mathbf{o}(t_g)$. Specifically: a combination of $C$ ``foveal'' views ($8\times8$ pixel patches), $F$ ``parafoveal'' views ($16\times16$ patches), and $P$ ``peripheral'' views  ($24\times24$ patches). We choose $C = 4$ (four overlapping foveal views, arranged in a $2 \times 2$ grid), $F = 1$ (one parafoveal view), and $P = 1$ (one peripheral view).\footnote{In Supplementary Note 12, we present the results of preliminary experimentation justifying this particular arrangement of streams.} 
Figure \ref{fig:saccade_scheme_example} illustrates the result of a two-step ($K=2$) saccade sequence over an image (of the digit zero, extracted from the MNIST database): the red dot-dashed box shows the four foveal and the single parafoveal and peripheral patches used to create $\mathbf{g}(k)$. 

To obtain the final glimpse vector, all foveal, parafoveal, and peripheral views (centered around the same center-point) are first average pooled to always be the shape of $S \times S$ pixels, i.e., $\mathbf{g}^v(k) \in \mathbb{R}^{S \times S}$. 
These are then vectorized (i.e., flattened), and concatenated to obtain $\mathbf{g}(k) \in \mathbb{R}^{( (C + F + P) * (S * S) ) \times 1}$. 
This means that the vector $\mathbf{g}(k)$ produced by the $k$-th saccade is as follows:
\begin{align}
    \mathbf{g}(k) = (<\text{Flat}\big(\mathbf{g}^1(k)\big),...,\text{Flat}\big(\mathbf{g}^v(k)\big),...,\text{Flat}\big(\mathbf{g}^V(k)\big)  >)^T \label{eqn:glimpse_vector_create}
\end{align}
where $V = C + F + P$ %, brackets $<\cdot>$ denote vector concatenation, 
and, specifically, indices $v = 1, 2, 3, 4$ would correspond to flattened foveal views 
%$\mathbf{g}^v(k) = \text{Flat}(\mathbf{p}^v)$ where $\mathbf{p}^v$ is the original $S \times S$ extracted patch, 
index $v = 5$ would correspond to a flattened parafoveal view 
%$\mathbf{g}^v(k) = \text{Flat}(\mathbf{p}^v)$, 
and index $v = 6$ would correspond to a flattened peripheral view 
%$\mathbf{g}^v(k) = \text{Flat}(\mathbf{p}^v)$
. See Supplementary Note 10 for additional technical details on constructing $\mathbf{g}(k)$. 

The above process means that any $\mathbf{g}(k)$ is a essentially collection of sampled sub-spaces of the observation $\mathbf{o}(t_g)$, represented in terms of several higher-resolution (smaller/close-up) views and several lower-resolution (larger/zoomed out) features. 
Although this scheme was designed for visual perception, a similar patch extraction process could be considered for other sensory domains, e.g., raw audio waveform representations, where relevant anatomical knowledge, e.g., the ear, spiral ganglion neurons in the cochlea, could be used to motivate the sampling scheme.

\subsection*{Neural Circuit Model Implementation and Specification}

\noindent 
\textbf{LGN-relay input filtering units.} At the (sensory) input layer of our MPC model, our LGN filter first spatially smoothens the incoming stream input $\mathbf{x} \in \mathbb{R}^{H \times W \times Z}$ (where $Z$ denotes the number of channels; in this work, we examine gray-scale images thus $Z=1$) via two independent two-dimensional Gaussian kernels. Note that the input to this set of filtering units could be, as in the case of each stream of the MPC model $\mathbf{x} = \mathbf{g}^v(k)$ ($H=W=S$ and $Z=1$; note that this means each stream's LGN units would need access to the pre-flattened $\mathbf{g}^{v}(k) \in \mathbb{R}^{S \times S \times 1}$) or, as in the case of the saccade planner described later, $\mathbf{x} = \mathbf{o}$ ($H=O_H$, $H=O_W$, and $Z=1$).  
One kernel represents a narrow excitatory receptive field center and a broader inhibitory receptive field surround \cite{rodieck1965quantitative} as follows:  
\begin{align}
\kappa_{\text{cen}}(u,v) = \frac{1}{2\pi \sigma _{\text{cen}}^{2}}\exp \left(-\frac{u^{2}+v^{2}}{2\sigma _{\text{cen}}^{2}}\right), \; \text{and}, \;
\kappa_{\text{sur}}(u,v) = \frac{1}{2\pi \sigma _{\text{sur}}^{2}}\exp \left(-\frac{u^{2}+v^{2}}{2\sigma _{\text{sur}}^{2}}\right)
\end{align}
where $(u, v) \in [-K, K]^2$ denotes the spatial window defined by the kernel of radius $K$. 
The depthwise, channel-independent spatially-blurred outputs are extracted via a cross-correlation  applied to the padded input image, i.e., 
$\mathbf{b}_{\text{cen}} = \mathbf{x} * \kappa_{\text{cen}}$ (center-processed input) and $\mathbf{b}_{\text{sur}} = \mathbf{x} * \kappa_{\text{sur}}$ (surround-processed input). %, where $*$ denotes convolution. 

To emulate the parallel processing streams of the early visual pathway, the blurred, center-surround outputs $\mathbf{b}_{\text{cen}}$ and $\mathbf{b}_{\text{sur}}$ are next mapped into two separate, half-wave rectified \emph{ON} and \emph{OFF} channels \cite{schiller1992and}. 
Specifically, each of these particular channels undergoes a form of non-linear shunting inhibition\footnote{This prevents MPC from allocating its capacity on modeling large shifts in ambient lighting; the smooth scaling of pixel values induced by this mechanism aids MPC in extracting features that are robust to extremes in lighting variation and global contrast.} that is governed/modulated by the (square of the) sem-saturation coefficient $\alpha_{\text{sat}}$ as well as scaling exponent values $\gamma_e$ (the excitation control value) and $\gamma_i$ (the inhibition control value). 
The \emph{ON} stream output activities (which captures local increments in luminance) and \emph{OFF} stream output activities (which captures local decrements in luminance) are computed as follows:
\begin{align}
    \mathbf{x}_{\text{ON}}=\frac{\left[\max (0,\mathbf{b}_{\text{cen}})\right]^{\gamma _{e}}}{\left[\max (0,\mathbf{b}_{\text{sur}})\right]^{\gamma _{i}}+\alpha^{2}}_{\text{sat}} + \eta_b, 
    \quad \text{and,} \quad 
    \mathbf{x}_{\text{OFF}}=\frac{\left[\max (0,\mathbf{b}_{\text{sur}})\right]^{\gamma _{e}}}{\left[\max (0,\mathbf{b}_{\text{cen}})\right]^{\gamma _{i}}+\alpha^{2}_{\text{sat}}}  + \eta_b
\end{align} 
where $\eta_b$ is a small, non-negative scalar that models baseline neuronal activity (which keeps the \emph{ON}/\emph{OFF} processing pathways tonically active). 
%%The ON/OFF streams preserve the spatial phase transitions underlying visual borders, treating dark boundaries on lighter backgrounds and light boundaries on darker backgrounds symmetrically, yielding more stable and organized input signals for both the MPC V1 component as well as the saccade planner. 
Finally, in our specific model of LGN filter units, we construct a single, signed contrast-invariant feature map $\mathbf{x}_{\text{lgn}} \in \mathbb{R}^{H \times W \times Z}$ 
by linearly integrating %(combining) 
the parallel \emph{ON}/\emph{OFF} streams via: $\mathbf{x}_{\text{lgn}} = \mathbf{x}_{\text{ON}} - \mathbf{x}_{\text{OFF}}$. 
The gain-control, edge-enhanced feature map produced by the LGN filter units ultimately protects the MPC encoding structure from disruptive shifts in global illumination. Note that the full glimpse vector $\mathbf{g}(k)$, after the LGN filter units process incoming signals, is a composition of the filtered sub-spaces, i.e., we apply Equation \ref{eqn:glimpse_vector_create} but over $\mathbf{g}^v(k) = \mathbf{g}^v_{\text{lgn}}(k)$. 

%\paragraph{Neuronal architecture and dynamics.} 
\noindent 
\textbf{Neuronal architecture and dynamics.} We next provide a mechanistic description (see Supplementary material for a visual depiction) %Figure \ref{fig:pc_flows} for a depiction) 
of the $V$ streams ($v \in \{1, 2,...,v,...,V\}$) that compose an MPC architecture, which optimize the free energy of 
Equation \ref{eqn:rpc_vfe}. Each layer of the $v$-th stream encodes expectations (mean values) that are parameterized as neuronal (population) activity: 
\begin{align}
    \mu^{\ell,v} = \mu\big(\mathbf{z}^{\ell-1,v}(t); \mathbf{W}^{\ell,v}\big) = \mathbf{W}^{\ell,v} \cdot \phi^{\ell-1}\big(\mathbf{z}^{\ell-1,v}(t)\big) %+ \mathbf{V}^{\ell,v} \cdot \phi\big(\mathbf{z}^{\ell,v}(t-1)\big)
\end{align}
where $\phi^{\ell-1}()$ denotes the element-wise nonlinearity applied to the $\ell$-th layer's state values $\mathbf{z}^{\ell-1,v}(t)$.\footnote{Neurobiologically, in this work, we refer to values as pre- and post-synaptic depending on their relationship to the following linear algebraic transformation: $\mathbf{a} = \mathbf{W} \cdot \mathbf{b}$; $\mathbf{a}$ contains the post-synaptic values, $\mathbf{b}$ contains the pre-synaptic values, and $\mathbf{W}$ contains the synaptic efficacies themselves.}  
$\mathbf{W}^{\ell,v}$ is a matrix that contains the (intra-stream) predictive synaptic efficacies for the $v$-th stream. %while $\mathbf{V}^{\ell,v}$ is a matrix which houses the recurrent/temporal synaptic efficacies. 
Note that the bottom (sensory) layer $\ell=0$ of an MPC stream has no nonlinearity, i.e., $\phi^0(\mathbf{z}^{0,v}) = \mathbf{z}^{0,v}$ (the identity), and is clamped to the relevant portion of the (LGN-filtered) sensory glimpse input, i.e., this means that $\mathbf{z}^{0,v}(t) = \mathbf{g}^v(k)$, where the $v$-th stream is provided with the $v$-th view of the glimpse vector $\mathbf{g}(k)$. We furthermore set the intra-stream covariance parameters to be scaled identity matrices $\Sigma^{\ell,v} = \sigma \mathbf{I}^{\ell,v}$ (where $\sigma > 0$), which simplifies our model's energy-based optimization.\footnote{We remark that incorporating dynamics inherent to neuronal implementations of precision-weighting %, e.g., such as the precision implementation in \cite{tang2023recurrent}, 
would be useful for more complex (modeling) tasks. }

In order for the $v$-th stream to make predictions of the $q$-th stream, each layer $\ell > 0$ of neurons is further equipped with lateral synaptic connections. This means that the cross-representation mean $\mu^{\ell,v,q}_C$ emitted by the $v$-th stream is parameterized as follows: 
\begin{align}
    \mu^{\ell,v,q}_C = \mu_C\big(\mathbf{z}^{\ell,v}(t); \mathbf{R}^{\ell,v,q}, \mathbf{A}^{\ell,v,q}\big)  = \mathbf{R}^{\ell,v,q} \cdot \phi^\ell\Big(\mathbf{z}^{\ell,v}(t)\Big) + \mathbf{A}^{\ell,v,q} \cdot \mathbf{a}(t)
\end{align}
where $\mathbf{R}^{\ell,v,q}$ contains the cross-stream prediction synapses (emitting from stream $v$ to stream $q$) and $\mathbf{A}^{\ell,v,q}$ contains the action-conditional, afferent synapses. $\mathbf{a}(t) \in [-1,1]^{2}$ is the action taken by the MPC model at time $t$; specifically, it is a two-dimensional vector encoding the chosen coordinates (positional coordinates of the sensory input relative to the MPC scheme, produced in the manner described in Methods). % in the later ``Epistemic glimpse planning'' sub-section). 
Furthermore, for additional simplicity, we set inter-stream covariances to be scaled identity matrices, i.e., $\Sigma^{\ell,v}_C = \sigma\mathbf{I}^{\ell,v}$ where $\sigma > 0$. 
Inspired by \cite{ororbia2022ngc} -- which demonstrated that lateral competition is useful for learning generative models -- the activation function $\phi()$ that we chose induces a fast form of lateral competition (without requiring physical lateral synapses) in the internal layers. Specifically, we chose $\phi()$ to promote high levels of sparsity within each stream. %, similar in spirit to the part-whole spiking model proposed in \cite{gebhardt2024time}. 
%We implement a recurrent form of the $N$-winners-take-all (NWTA) function \cite{ahmad2019can} by constructing an additional 
Given neuronal activities $\mathbf{z}^{\ell,v}$ as its argument, the activation $\phi(\mathbf{z}^{\ell,v})$ can be written out as follows:
\begin{align}
    \text{NWTA}(\mathbf{z}^{\ell,v}) &= 
    \begin{cases} 
      z^{\ell,v}_j & z^{\ell,v}_j \in \{N_w \text{ largest elements of } \mathbf{z}^{\ell,v} \} \\
      0 & \text{otherwise}
   \end{cases}
\end{align}
which is the $N$-winners-take-all (N-WTA) function \cite{ahmad2019can} where only the $N_w$ neurons with highest values within the layer/group $\mathbf{z}^{\ell,v}$ emit a non-zero firing rate (the rest that lose this cross-neuron competition will emit a zero).
\footnote{Although this winner-take-all function worked well for the experiments carried out in this study, future work would benefit by considering extensions, such as incorporating a duty-cycle \cite{ahmad2019can} to promote cooperation and the sharing of firing responsibilities among neurons within a group.} 
We further remark that the above N-WTA function can be implemented via lateral inhibition and self-excitation in the format of recurrently-wired lateral connections, much as in \cite{ororbia2022ngc}. 

Crucially, embedded within each layer $\ell > 0$ of an MPC stream is a set of prediction error neurons that compute the mismatch signals reporting how far off each layer's predictions are from their corresponding targets. There are two kinds of error neurons, which result from the free energy functional of Equation \ref{eqn:rpc_vfe}, for every layer -- intra-stream error units $\mathbf{e}^{\ell,v} $ and inter-stream error units $\mathbf{e}^{\ell,v,q}_C$. These two kinds of error neurons can be written down as:
\begin{align}
    \mathbf{e}^{\ell,v} &= \mathbf{z}^{\ell,v}(t) - \mu^{\ell,v}, \quad \text{// Intra-stream  mismatch signals} \label{eqn:intra_error} \\
    \mathbf{e}^{\ell,v,q}_C &= \mathbf{r}^{\ell,q}(t) - \mu^{\ell,v,q}_C \quad \text{// Inter-stream mismatch signals} \label{eqn:inter_error}
\end{align}
where we notice that mismatch signals will be produced as a result of either local predictions of intra-stream activity (within $v$), i.e., $\mu^{\ell,v}$ attempting to guess $\mathbf{z}^{\ell,v}(t)$, or inter-stream activity between $v$ and $q$, i.e., $\mu^{\ell,v,q}_C$ attempting to guess $\mathbf{z}^{\ell,q}(t)$. 
$\mathbf{r}^{\ell,q}(t)$ itself, the target for the cross-stream predictions, can take on several forms, e.g., a portion of neural activities of another stream or statistics thereof; here, in this study's model, we found that predicting the ``innovation'' of layer $\ell$ of stream $q$, i.e., $\mathbf{r}^{\ell,q}(t) = \mathbf{e}^{\ell,q}$, worked best.\footnote{
Predicting the cross-stream innovations (errors), as opposed to their direct state values, has a stabilizing effect on the MPC's overall dynamics; once a neighboring target stream's features have been successfully anticipated (i.e., when the target layer $\ell$'s error units have been reduced due to its own dynamics), the innovation collapses to zero which prevents continuous cross-talk from driving the source stream's (i.e., the one making the predictions) dynamics into runaway or oscillatory behavior. This design choice means that, in addition to the topology-weighting applied to inter-stream synapses, the inter-stream communication (contextually) regularizes a given stream's local error-driven updates. This further can be though of as a form of the precision-weighting \cite{friston2010free} %\cite{friston2005theory} 
that is sometimes built into hierarchical PC architectures (to ensure stability in its message-passing process).
}
\begin{comment}
%%%%%%%%%%%%%%%%%%%%%%%%%%%%%%%%%%%%%%%%%%%%%%%%%%%%%%%%%%%%%%%%%%%%%%%
%% Improvement on MPC
To improve the stability of the learning process of MPC (and further mitigate against representational/dimensional collapse \cite{geiping2023cookbook}, a well-known problem in SSL), we modify Equations \ref{eqn:intra_error} and \ref{eqn:inter_error} slightly as follows: 
\begin{align}
    \mathbf{e}^{\ell,v} &= \Big((1 - \beta) \mathbf{r}^{\ell_v} + \beta \mathbf{z}^{\ell,v}(t)\Big) - \mu^{\ell,v}, \quad \text{// Intra-stream  mismatch signals} \label{eqn:intra_error_targeted} \\
    \mathbf{e}^{\ell,v,q}_C &= \Big((1 - \beta) \mathbf{r}^{\ell_v} + \beta \mathbf{z}^{\ell,q}(t) \Big) - \mu^{\ell,v,q}_C \quad \text{// Inter-stream mismatch signals} \label{eqn:inter_error_targeted}
\end{align}
where $\mathbf{r}^{\ell_v}$ is the representation of stream $v$ produced by the initial (random) conditions of the network circuit; $\beta \in \in [0,1]$ is the interpolation coefficient between the random target projection and the current neural activity state (target). The integration of this random projection component is inspired by the use of random projections in the reinforcement learning literature \cite{burda2018exploration} and was found to serve, in the case of our MPC SSL, as a small attractive force towards vector space where individual codes are approximately orthogonal.
%%%%%%%%%%%%%%%%%%%%%%%%%%%%%%%%%%%%%%%%%%%%%%%%%%%%%%%%%%%%%%%%%%%%%%%
\end{comment}

Driven by the error neurons of Equations \ref{eqn:intra_error} and \ref{eqn:inter_error} %(or Equations \ref{eqn:intra_error_targeted} and \ref{eqn:inter_error_targeted}), 
the dynamics of the neuronal cells within each layer of an MPC stream follow the gradient flow of free energy; this flow is presented by the following (vectorized) ordinary differential equation (ODE):
\begin{align}
    \frac{\partial \mathcal{F}(\Theta)}{\partial \mathbf{z}^\ell(t)} = \tau_z \frac{\partial \mathbf{z}^{\ell,v}(t) }{\partial t} = -\mathbf{e}^{\ell,v} + \Big( \mathbf{E}^{\ell,v}_W \cdot \mathbf{e}^{\ell+1,v}  + \mathbf{E}^{\ell,v,q}_R \cdot \mathbf{e}^{\ell,v,q}_C\Big) \odot \frac{\partial \phi\big(\mathbf{z}^{\ell,v}(t)}{\partial \mathbf{z}^{\ell,v}(t)}\big)  \label{eqn:rpc_estep}
\end{align}
where $\tau_z$ is the neural cell membrane time constant (in milliseconds; ms) and $\frac{\partial \phi(\mathbf{z}^{\ell,v})}{\partial \mathbf{z}^{\ell,v}}$ is the partial derivative of the activation function with respect to the neural state activities at $t$. 
$\mathbf{E}^{\ell,v}_W$ contains the $v$-th stream's intra-stream message-passing synapses whereas $\mathbf{E}^{\ell,v,q}_R$ contain its inter-stream message-passing synapses. Note that these message-passing synapses can either be set to be the transposes of the corresponding forward connections, i.e., $\mathbf{E}^{\ell,v}_W = (\mathbf{W}^{\ell,v})^T$ and $\mathbf{E}^{\ell,v,q}_R = (\mathbf{R}^{\ell,v,q})^T$, or can usefully be learned 
%One more simplification -- that could be applied to any MPC stream -- is to set its feedback connection matrices to $\mathbf{E}^{\ell,v}_W = (\mathbf{W}^{\ell,v})^T$ and $\mathbf{E}^{\ell,v,q}_R = (\mathbf{R}^{\ell,v,q})^T$; note that these can, alternatively, be learned 
with Hebbian rules, as in \cite{ororbia2022ngc}. %\cite{rao1999visionresearch,ororbia2022ngc}

%% plasticity / learning (M-step)
\paragraph{Synaptic plasticity.} Learning in an MPC stream follows the gradient flow of Equation \ref{eqn:rpc_vfe} and synaptic connection strengths are updated according to Hebbian plasticity rules. The intra-stream synapses $\mathbf{W}^{\ell,v}$, the inter-stream synapses $\mathbf{R}^{\ell,v,q}$, and the action-conditional afferent synapses $\mathbf{A}^{\ell,v,q}$ of any MPC stream are updated as follows: 
\begin{align}
    \tau_w \frac{\partial \mathbf{W}^{\ell,v}}{\partial t} &= -\lambda_w\mathbf{W}^{\ell,v} + \mathbf{e}^{\ell,v} \cdot \big( \phi(\mathbf{z}^{\ell-1,v}) \big)^T, \label{eqn:intra_syn_update} \\
    \tau_w \frac{\partial \mathbf{R}^{\ell,v,q}}{\partial t} &= -\lambda_w\mathbf{R}^{\ell,v,q} + \mathbf{e}^{\ell,v,q}_C \cdot \big( \phi(\mathbf{z}^{\ell,v}) \big)^T, \label{eqn:inter_syn_update}\\ 
    \tau_w \frac{\partial \mathbf{A}^{\ell,v,q}}{\partial t} &= -\lambda_w\mathbf{A}^{\ell,v,q} + \mathbf{e}^{\ell,v,q}_C \cdot \big(\mathbf{a}^{\ell,v} \big)^T \label{eqn:action_syn_update}
\end{align}
where $\tau_w$ is a synaptic plasticity time constant (in ms) and $\lambda_w$ is a synaptic decay modulation coefficient. 
The inference and learning steps in an MPC scheme are scheduled according to an expectation-maximization \cite{dempster1977maximum} (EM) like process: 
\textbf{1)} inference (E-step) is carried out in an MPC stream by applying Equation \ref{eqn:rpc_estep} using Euler integration, for all layers $\ell > 0$, several times over a stimulus window length $E = T/\Delta t$ \footnote{$T$ is the length of stimulus presentation time for examining an input view $\mathbf{z}^{0,v}(t) = \mathbf{g}^v(k)$ and $\Delta t$ is the integration time constant; both are in milliseconds (ms).}; 
\textbf{2)} then synaptic learning (M-step) is performed by applying Euler integration of Equations \ref{eqn:intra_syn_update}, \ref{eqn:inter_syn_update}, and \ref{eqn:action_syn_update}, to the parameters of all layers $\ell > 0$, once. After the M-step is performed, updated synaptic matrices are finally constrained to have unit column Euclidean norms. % unit rows in Python-ic implementations

In Figure \ref{fig:fov_rpc}, we illustrate what a simple MPC architecture with $V=2$ two streams (each with $L=4$ layers) would look like; in this example, a single foveal stream predicts the activities (at each layer) of a single parafoveal stream and \textit{vice versa}; hence the 'meta' aspect of the ensuing representations. 
Notice that, our free energy framework complements self-supervised representation learning \cite{ericsson2022self} schemes, such as those that work to avoid (dimensional) collapse\footnote{Dimensional collapse refers to the case where the embeddings produced by a (neural) encoder model % such as the wo branches of a dual-embedding architecture, e.g., a Siamese neural network \cite{bromley1993signature}, 
degenerate to producing identical and constant output vectors. This is a degenerate outcome indicating that the model ends up learning to simply ignore the input data.} 
including information-maximization  \cite{bardes2021vicreg} %\cite{ermolov2021whitening,zbontar2021barlow,bardes2021vicreg} 
and regularization approaches \cite{assran2023self}. %\cite{assran2023self,drozdov2024video}
Specifically, MPC espouses a meta-representational narrative for SSL; representations of data features should be able to predict one another in an internally consistent fashion. In some sense, the MPC architecture speaks to the notion that complementary views of input patterns should yield embeddings that are 'close' to one another, as in some forms of masked prediction \cite{pathak2016context}. %\cite{chen2020simple,pathak2016context}
However, instead of focusing on input data, MPC operates exclusively in latent space, in the same spirit as \cite{bardes2021vicreg,assran2023self}, where parallel streams encoding input effectively learn to resonate with one another as a result of continual prediction and message passing. 

\paragraph{Epistemic glimpse and saccade planning.} Central to MPC's information processing is its in-built reflex-based saccade planner, which brings the model from just a purely action-conditioned one to a self-driven engine, much in the spirit of active perception \cite{mnih2014recurrent}  %\cite{mnih2014recurrent,sharafeldin2024active} 
and selection through the framework of planning-as-inference \cite{botvinick2012planning}. %\cite{botvinick2012planning,friston2016active,friston2017active}
Here, we formally specify the MPC saccade planner. Let the spatial grid domain of the environment, e.g., the image canvas, be $\Omega = \{0, 1, \dots, H-1\} \times \{0, 1, \dots, W-1\}$, where a discrete absolute coordinate is denoted as $(u, v) \in \Omega$. 
The saccade planner chooses the next coordinate vector $\mathbf{c}_{k+1} = [y_{k+1}, x_{k+1}]^T \in [-1, 1]^2$ by optimizing a continuous ``focus'' map $\mathbf{A}$ that balances reflexive bottom-up features with accumulated top-down (epistemic) prediction errors.

%bottom-Up saliency estimation
The bottom-up focus landscape $\mathbf{S}_{\text{BU}}(u,v)$ is computed from the local structural changes contained within the LGN response map $\mathbf{o}_{\text{lgn}}$ (i.e., the LGN filter applied to the whole image $\mathbf{o}$) via a (bottom-up) local feature energy extraction mechanism, i.e., $\mathbf{S}_{\text{BU}}=\mathcal{G}_{\text{BU}}(\mathbf{o}_{\text{lgn}})$. 
Specifically, the operator $\mathcal{G}_{\text{BU}}(\mathbf{o}_{\text{lgn}})$ transforms the map $\mathbf{o}_{\text{lgn}} \in \mathbb{R}^{H \times W \times C}$ by summing its absolute local amplitudes to produce a total (rectified) energy map $\mathbf{U}_{\text{energy}} \in \mathbb{R}^{H \times W}$ as follows:
\begin{align}
    \mathbf{U}_{\text{energy}}(u,v)=\sum _{c=1}^{C}\big|\mathbf{o}_{\text{lgn}}(u,v,c)\big| . 
\end{align}
%To simulate non-local dendritic spatial pooling and structural integration
This raw energy field is then smoothened via convolution with a normalized (2D) isotropic Gaussian kernel $\kappa_{\text{sal}}(x, y)$: 
\begin{align}
    \kappa_{\text{sal}}(x,y)=\frac{1}{Z}\exp \left(-\frac{x^{2}+y^{2}}{2\sigma _{\text{sal}}^{2}}\right)
\end{align}
where $(x,y) \in [-R, R]^2$ defines a localized pixel search matrix bounded by truncation radius $R = \lfloor 3\sigma_{\text{sal}} \rfloor$ ($Z = \sum_{x,y} \kappa_{\text{sal}}(x, y)$ is a normalization constant based on area). The final %un-gated 
bottom-up saliency map \(\mathbf{S}_{\text{BU}} \in \mathbb{R}^{H \times W}\) is then prepared through a cross-correlation (with a valid boundary convolution) over an edge-padded input matrix through: $\mathbf{S}_{\text{BU}} = \mathbf{U}_{\text{energy}} * \kappa_{\text{sal}}$. 
% primary visual cortex (V1) local energy models 
In a sense, the $\mathbf{U}_{\text{energy}}$ energy map emulates classical local energy \cite{adelson1985spatiotemporal} observed in primate V1 complex cells; complex cells achieve phase-invariant edge detection through the accumulation of rectified absolute output values of quadratic pairs of simple cells (e.g., the outputs of the LGN filter units). The spatial pooling of this energy, via kernel $k_{\text{sal}}$, produces a useful baseline map of local structural boundaries and visual textures \cite{findlay1999model} %\cite{itti1998model,findlay1999model} 
before any top-down context (provided by $\mathbf{S}_{\text{TD}}$) is applied.

%top-down epistemic error \& inhibition of return accumulation
The above bottom-up energy map $\mathbf{S}_{\text{BU}}$ is next combined with a top-down MPC energy-driven map. Specifically, let us define the history/trajectory of saccade-produced glimpses taken up until step $k$, under normalized coordinates (in range of $[-1,1]$), as a set of three sequences: 
\textit{(1)} the (normalized) coordinate\footnote{These are mapped back to absolute pixel coordinates $(y_\tau, x_\tau)$ via: $y_{\tau } = \frac{y_{\tau ,\text{norm}}+1}{2}(H-1),\quad x_{\tau }=\frac{x_{\tau ,\text{norm}}+1}{2}(W-1)$.} 
history $\{\mathbf{c}_\tau\}_{\tau=1}^k$, 
\textit{(2)} the MPC energies tracked thus far $\{\delta_\tau\}_{\tau=1}^k$ (where $\delta_\tau = -\mathcal{F}(\mathbf{g}(\tau); \Theta)$), and, 
\textit{(3)} the ``step validity'' masks $\{m_\tau\}_{\tau=1}^k \in \{0, 1\}$.

At current step $k$, the MPC model computes a top-down energy map $\mathbf{S}_{\text{TD}}(u,v)$ (the ``epistemic'' map) and a mask, which we label as the spatial inhibition-of-return \cite{findlay1999model} (IOR) map $\mathbf{I}(u,v)$. Let $d_\tau^2(u,v) = (u - y_\tau)^2 + (v - x_\tau)^2$ be the (squared Euclidean) distance from any pixel $(u,v)$ to the historical fixation site at $\tau$. 
The map $\mathbf{S}_{\text{TD}}(u,v)$ and IOR multiplier map $\mathbf{I}(u,v)$ are produced by an integration/accumulation through time as follows: 
\begin{align}
    \mathbf{S}_{\text{TD}}(u,v) &= \sum _{\tau =1}^{k}m_{\tau }\cdot \delta _{\tau }\cdot \exp \left(-\frac{d_{\tau }^{2}(u,v)}{2\sigma _{\text{err}}^{2}}\right) \\ 
    \mathbf{I}(u,v) &= \prod _{\tau =1}^{k}\left[1-m_{\tau }\cdot \exp \left(-\frac{d_{\tau }^{2}(u,v)}{2\sigma _{\text{IOR}}^{2}}\right)\right]
\end{align}
where $\sigma _{\text{err}}$ controls the spatial spread of the prediction error feedback and $\sigma _{\text{IOR}}$ establishes the spatial radius of neural foveal inhibition. % 
Notice that the $\mathbf{S}_{\text{TD}}$ map operationalizes this planner's epistemic uncertainty reduction goal: when the MPC circuit fixates on a region and encounters high error (negative free-energy) $\delta _{\tau }$, this error projected back down as an epistemic demand onto the focus field $\mathbf{A}$. This construction will actively pull subsequent saccades back towards zones of unresolved structural uncertainty \cite{wolfe2015saccadic}. 
% inhibition-of-return + oculomotor foraging efficiency 
The multiplicative map $\mathbf{I}(u,v)$, on the other hand, implements a neurophysiological IOR profile. In effect, in primates, it is observed that transient neural activity in the SC (and parietal circuits) corresponding to a spatial address can be actively suppressed \cite{findlay1999model}; this prevents the oculomotor loop from entering the potential ``trap'' of staring endlessly at the same high-contrast region of input space (in our model, the IOR gating forces MPC to explore novel geometric contours systematically across a perceived object).

% integrated epistemic value field
Ultimately, the epistemic map $\mathbf{S}_{\text{TD}}$ is combined with the bottom-up energy map $\mathbf{S}_{\text{BU}}$ (and their linear blend is dynamically gated by $\mathbf{I}$) through a convex combination to produce the necessary focus map: 
\begin{align}
    \mathbf{A}(u,v) = \Big[(1-\beta )\mathbf{S}_{\text{BU}}(u,v)+\beta \mathbf{S}_{\text{TD}}(u,v)\Big]\cdot \mathbf{I}(u,v)
\end{align}
where $\beta$ is the top-down ``focus'' coefficient. 
% superior colliculus + attentional blending
We remark that the structure of $\mathbf{A}$ embodies a form of the sensorimotor integration that happens within the intermediate and deep layers of superior colliculus (SC) \cite{isa2021tectum}. %\cite{gandhi2011motor,isa2021tectum}
The SC, in effect, receives rapid, reflexive bottom-up retinal input signals from the superficial layers (emulated by the bottom-up $\mathbf{S}_{\text{BU}}$ pathway). It then integrates these with top-down, goal-directed (or predictive) motor command signals that originate from the frontal eye fields (FEF) \cite{thompson1999detection} and the visual cortex (emulated by the $\mathbf{S}_{\text{TD}}$ pathway).

%to restrict attention within a valid spatial foraging boundary
Note that, in the model crafted for this study, we further incorporate a constraint to the focus map to enforce hard foraging boundaries (to restrict the coordinates of what the MPC saccade planner can choose) using an indicator mask $\mathds{1}_{\text{bound}}(u,v)$ that is evaluated over the restricted normalized grid space (e.g., $[-0.95, 0.95]^2$) as follows: 
\begin{align}
\mathbf{A}_{\text{bounded}}(u,v) = 
    \begin{cases}
    \mathbf{A}(u,v) &\text{if\ } \mathds{1}_{\text{bound}}(u,v) = 1 \\ 
    -10^{5}&\text{otherwise} . 
    \end{cases} 
\end{align}
The next action taken by MPC, i.e., the next coordinate selected by its saccade planner, is done by taking the `argmax' of the focus map $\mathbf{A}_{\text{bounded}}$ to obtain the final coordinates $(u^{*},v^{*})$ of the next glimpse: 
\begin{align}
(u^{*},v^{*})=\arg \max _{(u,v)\in \Omega }\left[\mathbf{A}_{\text{bounded}}(u,v)\right]. \label{eqn:epistemic_foraging}
\end{align}

%stochastic foraging via the Gumbel-max trick / Gumbel-Max stochastic exploration (saccadic noise)
While the above process permits MPC to choose where it is to glimpse next effectively, it does in a deterministic fashion. To encourage an additional degree of exploration, we studied a modification of the above scheme that we call \emph{stochastic foraging}. 
Specifically, this variation replaces the final coordinate selection step described above by instead sampling the next discrete pixel target $(u^*, v^*)$, under a temperature schedule $\mathcal{T}$, using independent and identically distributed standard Gumbel noise \cite{maddison2014sampling} samples $\xi(u,v)$. These samples are drawn, for every coordinate site, using a uniform distribution $U(u,v) \sim \mathcal{U}(10^{-7}, 1)$ as follows: $\xi (u,v) = -\log \big( -\log U(u,v) \big)$. 
The focus map is then perturbed using this noise before finally choosing the coordinates with maximum energy value:
\begin{align}
    (u^{*},v^{*}) = \arg\max _{(u,v)\in \Omega }\left[ \frac{\mathbf{A}_{\text{bounded}}(u,v)}{\mathcal{T}}+\xi (u,v)\cdot \mathds{1}(\mathcal{T}>0) \right] . \label{eqn:stochastic_foraging}
\end{align}
Note that the above injected Gumbel-Max noise \cite{maddison2014sampling} works to broaden MPC's visual exploration footprint (by making it occasionally select sub-optimal locations to break out of potential loops induced by patterns in sensory input values). This stochasticity further emulates the fact that biological eye movements are not deterministic and are affected by the (neuronal) noise of the brainstem and FEF \cite{sparks2002brainstem}, resulting in more stochastic saccadic target selections. (In Supplementary Note 11, we describe how the temperature $\mathcal{T}$ is scheduled.) 

Finally, note that chosen pixel coordinates, using any variant of the planner, are normalized back to the continuous range $[-1, 1]^2$, yielding the next action to be taken by the MPC circuit: $\mathbf{c}_{k+1}=\left[\frac{2u^{*}}{H-1}-1,\quad \frac{2v^{*}}{W-1}-1\right]^{T}$. This also serves as the input to condition MPC's processing of the next glimpse. 

% Central to MPC's information processing is its in-built reflex-based saccade planner, which brings the model from just a purely action-conditioned one to a self-driven engine, much in the spirit of active perception \cite{mnih2014recurrent,sharafeldin2024active} and selection through the framework of planning-as-inference \cite{botvinick2012planning,friston2016active,friston2017active}. 
To understand how an MPC circuit processes sensory information through its reflex-based saccade planning, we examine its information foraging behavior quantitatively in Supplementary Note 9 (where we carried out an experimental analysis of MPC under different foraging conditions). In the earlier Figure \ref{fig:mpc_saccade_trajectory}, we provided a sampled experimental trajectory of the MPC's saccade process (under the deterministic epistemic planner, i.e., ``MPC-epi'') to observe the trajectory taken (with a constrained budget of $40$ glimpses) when MPC processes a complex natural image stimulus (in the Supplementary Material, we provide two further qualitative examples on the MNIST database). 
As seen in Figure \ref{fig:mpc_saccade_trajectory}, the MPC model engages in carefully chosen micro- and macro-saccades, effectively creating a mesh around the central object/figurine within the natural image, wasting little time in the outer regions where no meaningful information exists. Motivated by the effectiveness of our reflexive, active perceptual saccade planner, future work should develop a complementary motor circuit to drive the selection of the MPC sampling of the input (with a bias towards policies that entail the minimum number of saccades and consider a better balance of representational efficiency-efficacy). This would bring our model from an action-conditioned, reflexive one to a self-driven model, much in the spirit of active perception \cite{mnih2014recurrent}  %\cite{mnih2014recurrent,sharafeldin2024active}
and selection through the framework of planning-as-inference \cite{botvinick2012planning}. %\cite{botvinick2012planning,friston2016active,friston2017active}. 

%%%%%%%%%%%%%%%%%%%%%%%%%%%%%%%%%%%%%%%%%%%%%%
\paragraph{Latent glimpse path integration.} In biological oculomotor systems, eyes execute rapid saccades % (three to fours times per second, e.g., in the primate visual system)
in service of extracting foveal ``snapshots'' of different parts of a visual scene. However, these raw snapshots are not explicitly buffered and are, instead, incorporated into a progressively built, stable non-retinotopic representation of a global object/item within a working memory \cite{bays2008dynamic,melcher2015nonretinotopic} %\cite{bays2008dynamic,melcher2003spatiotopic,melcher2005spatiotopic,melcher2015nonretinotopic}
(over a finite window of time). 
To obtain global latent codes (one code per visual scene; $\mathbf{z}_K$, i.e., the code at the end of $K$ glimpses) from the glimpse latent vectors of any of our saccade-based, iterative models (e.g., MPC, GPC-\emph{fov}), we designed an aggregation scheme, inspired by principles of spatial mapping in biological neurons/grid-cells \cite{stensola2016grid},  % \cite{fuhs2006spin}, 
that integrates both neural activity (of the layer/population farthest away from the sensory input layer) and glimpse coordinates. % 
This process was designed to be iterative in nature -- yielding the current state of the global code $\mathbf{z}_k$ after any saccade $k$ -- in order to avoid storing histories of glimpse pathways in (computer) memory. 
See Supplementary Note 6 for a full formal treatment of our proposed latent glimpse path integration model. 
%%%%%%%%%%%%%%%%%%%%%%%%%%%%%%%%%%%%%%%%%%%%%%

\subsection*{Programming}

The experiments, the baseline models, and the proposed computational framework were written in Python, using the NGC-Learn \cite{ororbia2022ngc}, Numpy, and JAX libraries. Experiments were sped up using a GPU card.

\section*{Data Availability}

The datasets, i.e., MNIST \cite{lecun1998mnist}, Kanji-MNIST \cite{clanuwat2018deep}, NORB \cite{lecunLearningMethodsGeneric2004NORB}, and ETH-80 \cite{leibe2003analyzing}, %\cite{leibe2003analyzing,chen2020covariance}, 
used by this study are publicly available.

\section*{Code Availability}
The GitHub repository that supports this work can be found at the following URL: \hyperlink{https://github.com/NACLab/encoder-only-predictive-coding}{https://github.com/NACLab/encoder-only-predictive-coding}.

%% funding, acknowledgements
\subsection*{Acknowledgements}

We thank Viet Nguyen for writing the custom (I-)JEPA baseline model used for this paper's experiments. 
This research was funded in whole, or in part, by the Cisco Research Gift Award \#26224 (AO). This research was funded in whole, or in part, by the Wellcome Trust [203147/Z/16/Z] (KF), the National Science Foundation (NSF) (EFRI grant no.\ 2223495) (RPNR) (Science of Learning and Augmented Intelligence grant no.\ 2618151) (AO), and a Frameworks grant from the Templeton World Charity Foundation (RPNR). 
For the purpose of Open Access, the authors have applied a CC BY public copyright license to any Author Accepted Manuscript version arising from this submission.

\bibliographystyle{acm}
\bibliography{ref}

\newpage
\section*{Appendix and Supplementary Material}

\tableofcontents

\section{Supplementary Material Overview}
\label{sec:supplmenent_overview}

In this supplement/appendix, we present a review of the relevant self-supervised learning and neuro-mimetic intelligence literature, limitations and directions for future work related to Neuro-SSL, additional details related to our experimental protocols and other important modeling details related to the meta-representational predictive coding (MPC) framework and related model baselines developed in the main paper. Furthermore, we provide additional experiments and analyses that were conducted (that are referenced in the main manuscript text) in service of the main manuscript empirical inquiries.  

%\newpage 

\section{Related Work}
\label{sec:lit_review}

\nocite{gebhardt2024time,strasburger2011peripheral,wang2017central,sasaki2001local,malach2002topography,hasson2003large,krauzlis2005control,findlay2009saccadic,efron2019lidwiper,ballard2009modelling,krauzlis2005control,findlay2009saccadic,rensink2000dynamic,sanocki2003representation,elias1955predictive,srinivasan1982predictive,friston2008hierarchical,clark2015surfing,seth2021predictive,mishkin1982contribution,zeki2005ferrier,wurtz2011thalamic,ognibene2014ecological,rumelhart1986learning,sharafeldin2024active,ermolov2021whitening,zbontar2021barlow}
\noindent 
\textbf{Self-supervised learning and representations.} Self-supervised learning (SSL), specifically self-supervised representation learning \cite{ericsson2022self} (SSRL), which can be viewed as a special case of unsupervised learning, strives to learn features or (abstract) representations of data without using supervisory annotation, e.g., labels of semantic categories. As opposed to unsupervised learning which centers around density estimation or (input data) reconstruction, SSRL relies on what are known as ``pretext tasks'' or artificial tasks that exploit knowledge related to a particular input modality (or modalities). Pretext SSL tasks can take a wide variety of forms, ranging from counting visual primitives in a scene \cite{noroozi2017representation} to in-painting \cite{zhang2022inpaint2learn} one part of (masked out) input using other (non-masked) portions of the input. SSRL methodology in machine intelligence research seeks to develop approaches that acquire representations (or those that learn general features) of data that facilitate strong downstream supervised learning performance without requiring time-consuming effort from human annotators (to produce the labels required by supervised learning). The chosen pretext task(s) tend to be be less complex than full, raw data generative modeling, possibly yielding representations of input that are less ``distracted'' by noise or irrelevant data details, with many motivating cases coming from reinforcement learning research \cite{burda2018exploration,laskin2020curl,mazzaglia2021contrastive,nguyen2024r}.

In learning useful representations \cite{bengio2013representation} or latent embeddings of input data, a wide variety of SSL/SSRL methods have been studied and developed \cite{geiping2023cookbook}. Recent approaches, particularly those that can be labeled as ``joint-embedding architectures'' \cite{bardes2021vicreg}, can be broken down roughly into a few general categories: contrastive approaches, information-maximizing / regularization approaches, or those which are driven by particular heuristics \cite{grill2020bootstrap,richemond2020byol,chen2021exploring}. Contrastive methods focus on constructing objectives that pull/attract embeddings of similar inputs (e.g., images) closer to one another and push embeddings of dissimilar inputs away from each other. These schemes strongly rely on either: 
\begin{itemize}[noitemsep,nolistsep]
    \item \textbf{1)} an effective mining process for uncovering dissimilar images  within a batch \cite{chen2020simple} or memory bank \cite{he2020momentum};  
    \item \textbf{2)} the design of a synthetic process that creates out-of-distribution or negative data examples \cite{hinton2002training,hinton2022forward,ororbia2023predictive,ororbia2024contrastive};  
    or, 
    \item \textbf{3)} a quantization/clustering scheme \cite{caron2018deep} that assigns embeddings of dissimilar data patterns to different clusters within a unit sphere. 
\end{itemize}
Information-maximization SSL methods use objective functions that decorrelate and orthogonalise variable/dimension pairs within an joint-embedding of latent vectors; this is argued to (indirectly) maximize the information content of embedding vectors \cite{bardes2021vicreg,geiping2023cookbook}. Generally, these methods focus on either: 
\textbf{1)} driving the normalized cross-correlation matrix of the two embeddings (of two different, yet complementary inputs, e.g., two transformations of an image) produced by the architecture towards the identity \cite{zbontar2021barlow,bardes2021vicreg}, or 
\textbf{2)} whitening and spreading out embedding vectors (of input) across the unit sphere \cite{ermolov2021whitening}. Although MPC does not encode explicit objectives that directly maximize information, its focus on encouraging ``resonant'', predictive sub-representations of different, dynamically-selected (temporally and spatially adjacent) subsets of input via a cross-stream message passing scheme brings it closest to regularization-based SSL approaches, e.g., the JEPA family \cite{garrido2024learning,bardes2024revisiting}, as well as latent predictive learning (LPL)  \cite{halvagal2023combination}. Future efforts should investigate how the various biological mechanisms under-pinning MPC circuitry, such as its mechanisms for inducing lateral competition or topology-weighting of its inter-stream predictions of innovations, aid in mitigating potential dimensional collapse \cite{jing2021understanding} of the underlying neuronal activities.  

%Our work complements regularization-based setups including information-maximizing schemes since MPC can be distilled to a scheme where embeddings (at all levels of abstraction) of dynamically-selected subsets of the input learn to predict one another and share (error/mismatch) messages over small windows of processing time to settle on complementary latent activities. This allows us to write down a free energy functional in terms of a cross-representation attraction problem, giving rise to our proposed representational predictive coding framework.

% Biomimetics, PC, MPC/MFA
\noindent 
\textbf{Biomimetic self-supervised (representation) learning.} With respect to biomimetic intelligence, there have been several approaches to construct schemes that are (essentially) encoder-centric. Some approaches focus on Hebbian plasticity \cite{grinberg2019local,miconi2021hebbian,moraitis2022softhebb}; while these models have the benefit of operating with only local pre- and post-synaptic statistics -- to drive various forms of associative learning -- it is difficult to write down the cost/energy functionals that they are optimizing, resulting in an obscure optimization-success tracking experience.\footnote{Work is ongoing for uncovering the implicit objectives that these forms of plasticity might be approximating \cite{pehlevan2017similarity,melchior2019hebbian}.} See \cite{isomura2020reverse} %Isomura et al. 
for a worked example with recurrent neural networks. More recent efforts include approaches that fall under the banner of forward-only learning \cite{kohan2023signal,ororbia2024review}, with particular approaches such as (the unsupervised forms of) forward-forward \cite{hinton2022forward} and predictive forward-forward \cite{ororbia2023predictive} learning offering useful ways of conducting SSRL. Scientific inquiry along this direction has led to insights into how systems of spiking neuronal cells could engage in self-supervised forward-only learning \cite{ororbia2024contrastive,merkel2024contrastive}, building strong connections between machine learning, computational neuroscience, and neuromorphic engineering. Nevertheless, these forms of neuro-mimetic learning inherit the same limitations as related machine learning SSL contrastive methodology, i.e., they require the generation or mining of negative samples to facilitate the proper organization of latent embeddings. 

Predictive coding (PC), which is a promising biomimetic learning-and-inference scheme that has emerged from theoretical/computational neuroscience \cite{elias1955predictive,olshausen1997sparse,rao1999predictive} and been continuously developed in neuroscience-inspired machine learning research \cite{salvatori2023brain,ororbia2023brain}, primarily takes on the form of an auto-associative memory \cite{salvatori2021associative} structure (thus decoder-focused) in modeling sensoria. The goal of any unsupervised PC model is to optimize its (variational) free energy (VFE) \cite{friston2009predictive,friston2010free}, leading to neuronal dynamics and message passing that follow the (gradient) flow of this VFE, where synaptic connection strengths minimize the same VFE objective, resulting in error-guided Hebbian plasticity. Most PC models are formulated as unsupervised associative memory engines (such as this work's GPC baseline models), which predict raw sensory inputs. These models acquire their distributed representations, which constitute their underlying models-of-the-world, as a consequence of optimizing VFE \cite{rao1999predictive,friston2010free,chalasani2013deep,salvatori2021associative,ororbia2022ngc,salvatori2022learning,salvatori2023brain}. Beyond unsupervised reconstruction, PC has been formulated for supervised learning, e.g., classification \cite{whittington2017approximation,salvatori2022learning,ororbia2022convolutional,millidge2022predictive}, and for reinforcement learning/active inference \cite{ororbia2023active}; although these PC formats often do not take on a decoder/associative memory format, they generally require annotation/human supervision to provide desired target signals or priors. This work directly recasts PC in terms of a non-auto-associative, self-supervised learning framing -- meta-representational PC (MPC) -- showing that it is possible, from a free energy principle perspective, to learn distributed representations of a sensorium without reconstructing raw sensory inputs. In essence, MPC entails a kind of generative learning of latent causes, guided by afferent synaptic connections that supply motor-action information to the circuit's neuronal units (actions, in our case, are the coordinates of saccades/glimpses), further elaborated by leveraging conditional independencies \cite{friston2016functional} to reproduce the computational architecture found in the visual system, e.g., central (foveal and parafoveal) and peripheral sensing, and associated parvocellular and magnocellular streams in the visual cortical hierarchy.

\textbf{Active predictive coding:} A recent variant of PC called active predictive coding (APC) \cite{rao2024natureneuro,rao2023active}, which is inspired by the primacy of actions in the neocortex, utilizes a hierarchy of coupled sensory prediction and action policy modules for solving a variety of sensory-motor tasks such as active perception, learning part-whole hierarchies, and spatial navigation. In APC, higher level latent sensory states and abstract actions change the lower-level state prediction function and policy function respectively according to the current task using hypernetworks \cite{ha2016hypernetworks} or top-down modulation of lower-level networks. MPC shares with APC the emphasis on using actions (saccadic eye movements) to generate a sequence of glimpses, which are then used for self-supervised learning. While APC learns an eye movement policy to intelligently sample the scene according to the task at hand, the current implementation of MPC uses a reflex-based scheme to generate movements. On the other hand, APC relies on prediction errors from predicting raw inputs for learning while MPC avoids input prediction and relies on prediction errors from predicting latent states within and across multiple visual streams. An obvious direction for future research is to combine the strengths of APC and MPC by learning hierarchical policy networks for intelligently sampling the sensorium while learning task-specific and hierarchical latent state representations via predictions across visual or more broadly, multimodal sensory streams. 

\section{Generative Predictive Coding Baselines}

%\paragraph{Generative predictive coding.} 
To contrast the proposed MPC framework with standard PC \cite{rao1999predictive,ororbia2022ngc,salvatori2023brain} (which we will refer to as generative PC\footnote{Even though all free energy-centric models learn probabilistic generative models, we emphasize the word ``generative'' to emphasize the decoder-centric nature of most standard PC models.}; GPC), we provide a brief explication of GPC's requisite free energy functional and its resultant dynamics. Specifically, when processing a clamped sensory input $\mathbf{z}^0(t) = \mathbf{o}(t_g)$, a GPC circuit works -- under a dynamic expectation-maximization \cite{dempster1977maximum} scheme -- to optimize the following VFE:
\begin{align}
    \mathcal{F}(\Theta) = \sum^{L-1}_{\ell=0} \mathcal{N}\Big( \mathbf{z}^\ell(t); \mu^\ell, \Sigma^\ell\Big) + \sum_{p,i,j} \mathcal{N}(\Theta[p]_{ij}; 0, \lambda_w) \label{eqn:gpc_vfe}
\end{align} 
where, depending on the distribution that one assumes over sensory inputs, one can modify the above functional to use other likelihoods at specific layers, e.g., a multivariate Bernoulli distribution for $\ell=0$ as was done in \cite{ororbia2022ngc}. Note that the above VFE has been expressed such that it also includes the same synaptic prior used in the MPC circuit in Equation 1. %\ref{eqn:rpc_vfe}. 
Given its goal to learn how to synthesize sensory inputs, a GPC scheme usually focuses on processing and predicting the entire sensory input $\mathbf{o}(t_g)$. Much as in the MPC circuit, the expectation $\mu^\ell$ at each layer is produced via a transformation, i.e., $\mu^\ell(t) = \mathbf{W}^\ell \cdot \phi(\mathbf{z}^{\ell+1}(t))$. The covariance parameters, as in the MPC model, are simplified to the scaled identity matrix $\Sigma^\ell = \sigma \mathbf{I}^\ell$ ($\sigma > 0$).

In the above VFE (Equation \ref{eqn:gpc_vfe}) and in the structure for GPC, error/mismatch neurons can be represented as a subtractive difference, i.e., $\mathbf{e}^\ell = \mathbf{z}^\ell(t) - \mu^\ell(t)$, and message passing emerges as a consequence of using feedback synaptic connections in tandem with these error units. Specifically, the free energy gradient flow -- that the neuronal units adhere to -- is the following ODE: 
\begin{align}
    \frac{\partial \mathcal{F}(\Theta)}{\partial \mathbf{z}^\ell(t)} = \tau_z \frac{\partial \mathbf{z}^\ell(t) }{\partial t} = -\mathbf{e}^\ell + \Big( \mathbf{E}^\ell \cdot \mathbf{e}^{\ell-1} \Big) \odot \frac{\partial \phi\big(\mathbf{z}^\ell(t)\big)}{\partial \mathbf{z}^\ell(t)} \label{eqn:gpc_step}
\end{align}
where $\mathbf{E}^\ell = (\mathbf{W}^\ell)^T$ . Note that Equation \ref{eqn:gpc_step} (and the VFE of Equation \ref{eqn:gpc_vfe}) can be further modified to incorporate additional constraints such as kurtotic priors that encourage sparsity in the latent states (we use a Laplacian prior in the GPC models studied in main manuscript, 
%Section \ref{sec:experiments}, 
to recover the modeling setups of \cite{rao1999predictive} and \cite{olshausen1997sparse}). 
After solving Equation \ref{eqn:gpc_step} for $T/\Delta t$ steps (examining an observation $\mathbf{z}^0 = \mathbf{o}(t_g)$), Hebbian adjustments may be made to the synaptic weight matrices as follows:
\begin{align}
    \tau_w \frac{\partial \mathbf{W}^{\ell}}{\partial t} = -\lambda_w\mathbf{W}^{\ell} + \mathbf{e}^{\ell} \cdot \big(\mathbf{z}^{\ell-1} \big)^T \label{eqn:gpc_update}
\end{align}
where the first term (right-hand side of the equality) constitutes the controllable weight decay that emerges from the synaptic prior introduced in Equation \ref{eqn:gpc_vfe}; this recovers the Gaussian prior over synapses in \cite{rao1999predictive}). 

\begin{figure}[!t]
     \centering
     \begin{subfigure}[b]{0.45\textwidth}
         \centering
         \includegraphics[width=0.3775\textwidth]{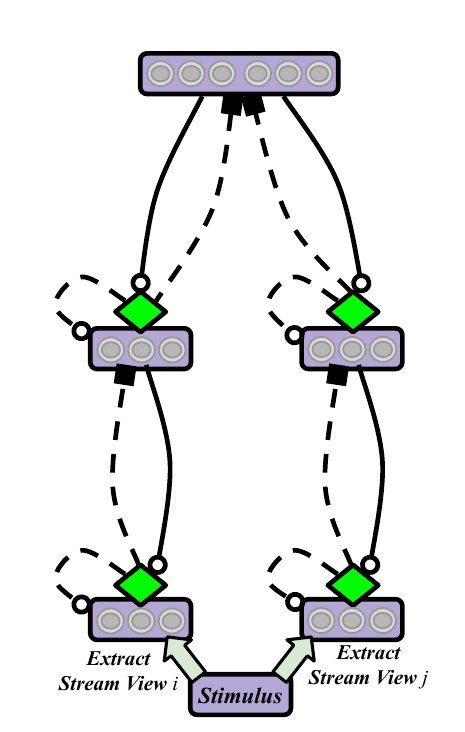}
         \caption{GPC-\emph{fov}.}
         \label{fig:gpcfov_circuit_mp}
     \end{subfigure}
     \begin{subfigure}[b]{0.45\textwidth}
         \centering
         \includegraphics[width=0.88\textwidth]{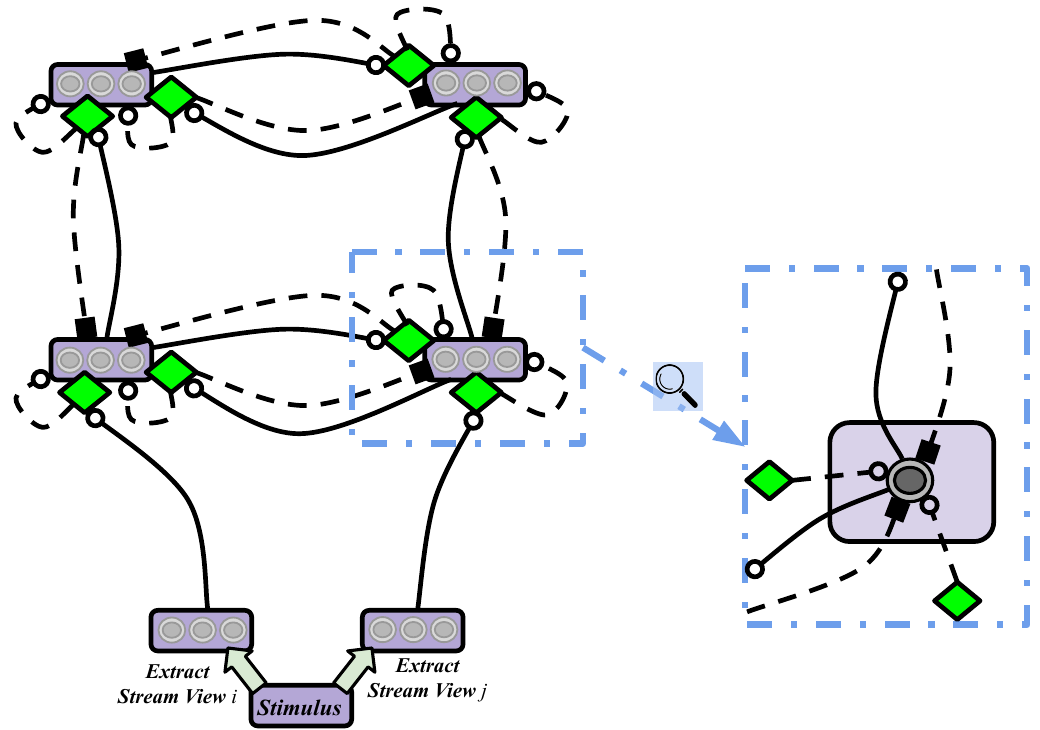}
         \caption{MPC.}
         \label{fig:mpc_circuit_mp}
     \end{subfigure}
     \vspace{-0.25cm}
        \caption{
        \small{
        \textbf{Structures of generative and meta-representational predictive coding schemes.} Depicted are two proposed variants of predictive coding -- a ``field-of-view'' form of generative predictive coding  (GPC-\emph{fov}) and meta-representational predictive coding (MPC) -- that process the same input stimulus. In this graphical example, two portions of visual input, at a point in time, are extracted (yielding a dual view, each containing different patches of information) %, possibly containing foveal, parafoveal, or peripheral patch pixel information) 
        by an eye movement process (represented by the pale green fat arrows), such as the saccade planning scheme 
        described in the ``Epistemic glimpse planning'' sub-section. %involuntary saccades described in Section \ref{sec:sensory_saccades}. 
        Specifically, we show:  
        (\textbf{a}) the proposed GPC-\emph{fov} (with two streams) processing a dual view of the input, i.e., a generalization of GPC that uses the same information as our MPC encoder; and 
        (\textbf{b}) the proposed MPC (with two streams) processing a dual view of the sensory input. Note that solid (black) arrows with open circles denote inhibitory (predictive) synapses, dashed (black) arrows with solid squares denote a population of excitatory (message-passing) synapses, purple boxes indicate a population of neurons encoding latent states, and green diamonds denote a group of error neurons for a specific layer. The zoomed-in inset for sub-Figure \ref{fig:mpc_circuit_mp} shows the incoming and outgoing wired connections to a single neuron within a population. 
        }
        }
        \label{fig:pc_circuit_types}
        \vspace{-0.5cm}
\end{figure}

Note that, in this work, we further modified the above GPC model—which we named the ``GPC-\emph{fov}'' (field-of-view GPC) model—to (iteratively) process glimpses as in MPC. In order to do so, we changed the input that the GPC model predicts to be $\mathbf{z}^0 = \mathbf{g}(k)$ and converted the GPC model's synaptic matrices $\{\mathbf{W}^1, \mathbf{W}^2,..., \mathbf{W}^L\}$ %up to but not including $\mathbf{W}^L$ ($\mathbf{W}^L$ was configured to be a dense, fully-connected matrix) 
to be block matrices; the bottom matrix $\mathbf{W}^1$ contained a number of blocks set equal to $V$ (number of streams) whereas upper-level matrices would contain fewer, wider blocks (e.g., $\mathbf{W}^2$ contains $V-1$ overlapping blocks, $\mathbf{W}^3$ contains $V-2$ blocks, and so on). This means that, after every synaptic update (Equation \ref{eqn:gpc_update}), we would constrain this block structure through application of a binary mask. This stream-based form of GPC, the GPC ``field-of-view'' (``GPC-\emph{fov}'') model, is our generalization of the full patch-level model of \cite{rao1999predictive} to learning how to generate dynamically-extracted patches of different resolutions; the GPC-\emph{fov} will use the same saccade planning scheme described earlier as well as use the same glimpse path integration scheme (described next). 
In Figure \ref{fig:pc_circuit_types}, we depict and visually compare the structure of the GPC-\emph{fov} and MPC studied in this work 
%of the various kinds of predictive coding models we study in this work, including the classical GPC as well as the GPC-\emph{fov} and MPC 
(both shown, for simplicity, processing two streams of the input). 

%%%%%%%%%%%%%%%%%%%%%%%%%%%%%%%%%%%%%%%%%%%%%%%%%%%%%%%%%%%%%%%%%%%%%%%%%%%%%%%%
%%%%%%%%%%%%%%%%%%%%%%%%%%%%%%%%%%%%%%%%%%%%%%%%%%%%%%%%%%%%%%%%%%%%%%%%%%%%%%%%

\section{Visual Depiction of MPC Inference Dynamics}

%%%%%%%%%%%%%%%%%%%%%%%%%%%%%%%%%%%%%%%%%%%%%%%%%%%%%%%%%%%%%%%%%%%%%%%%%%%%%%%%%%%%%%%%%%%
%% prediction and message-passing flow diagram
\begin{figure}[!t]
     \centering
     \begin{subfigure}[b]{0.485\textwidth}
         \centering
         \includegraphics[width=0.85\textwidth]{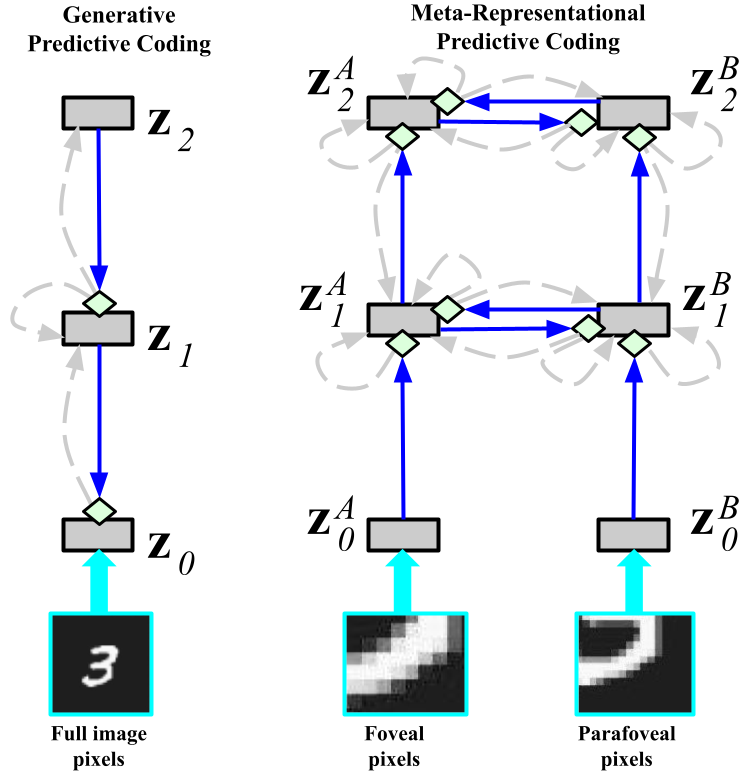}
         \caption{Flow of predictions (solid blue arrows).}
         \label{fig:prediction_flow}
     \end{subfigure}
     %\hfill
     \begin{subfigure}[b]{0.485\textwidth}
         \centering
         \includegraphics[width=0.855\textwidth]{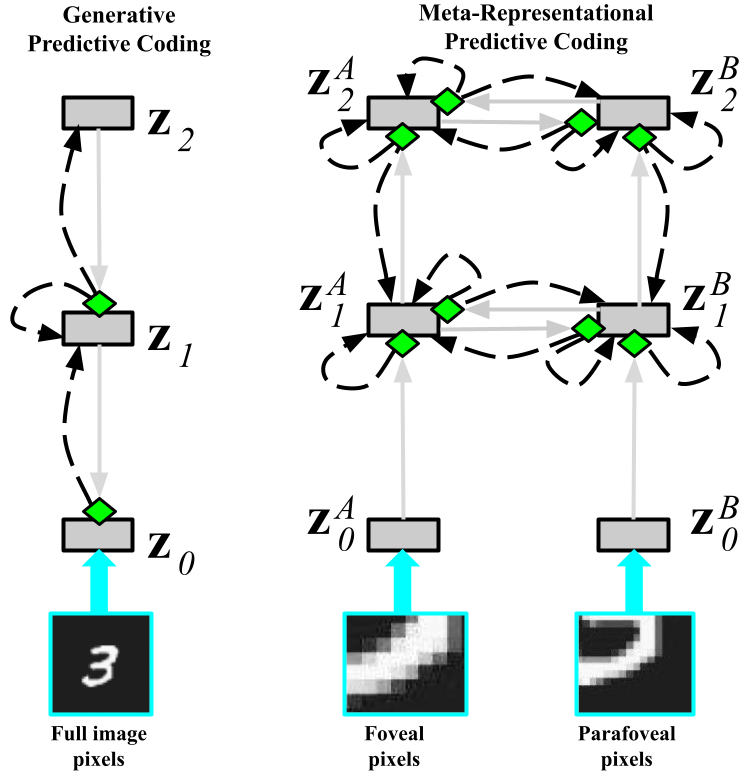}
         \caption{Message passing (dashed black arcs) flow.}
         \label{fig:message_passing_flow}
     \end{subfigure}
        \caption{
        \small{
        \textbf{Illustration of the message passing in MPC.} For generative predictive coding (GPC) and (a dual-stream) meta-representational predictive coding (MPC), depicted is: 
        (\textbf{a}) the flow/directional pattern of predictions made (solid blue arrows, which indicate neuronal populations that produce a prediction) in GPC versus MPC, and 
        (\textbf{b}) the flow/direction of message passing (dashed black arcs, which indicate feedback pathways that carry prediction errors) that result from GPC versus MPC prediction patterns (in sub-figure \text{a}). 
        Solid gray boxes indicate neuronal populations encoding latent states, while green diamonds indicate populations of error neurons. Both types of PC represent the same number of latent states; the MPC shown is an architecture of two streams where stream $v=$``A'' is shown processing foveal sensory information and stream $v=$''B'' processes peripheral sensory information.
        }
        }
        \label{fig:pc_flows}
        \vspace{-0.5cm}
\end{figure}
%%%%%%%%%%%%%%%%%%%%%%%%%%%%%%%%%%%%%%%%%%%%%%%%%%%%%%%%%%%%%%%%%%%%%%%%%%%%%%%%%%%%%%%%%%%

\section{Visualization of MPC Latent Codes}

\begin{figure}[tp!]
     \centering
     \begin{subfigure}[b]{0.325\textwidth}
         \centering
         \includegraphics[width=0.75\textwidth]{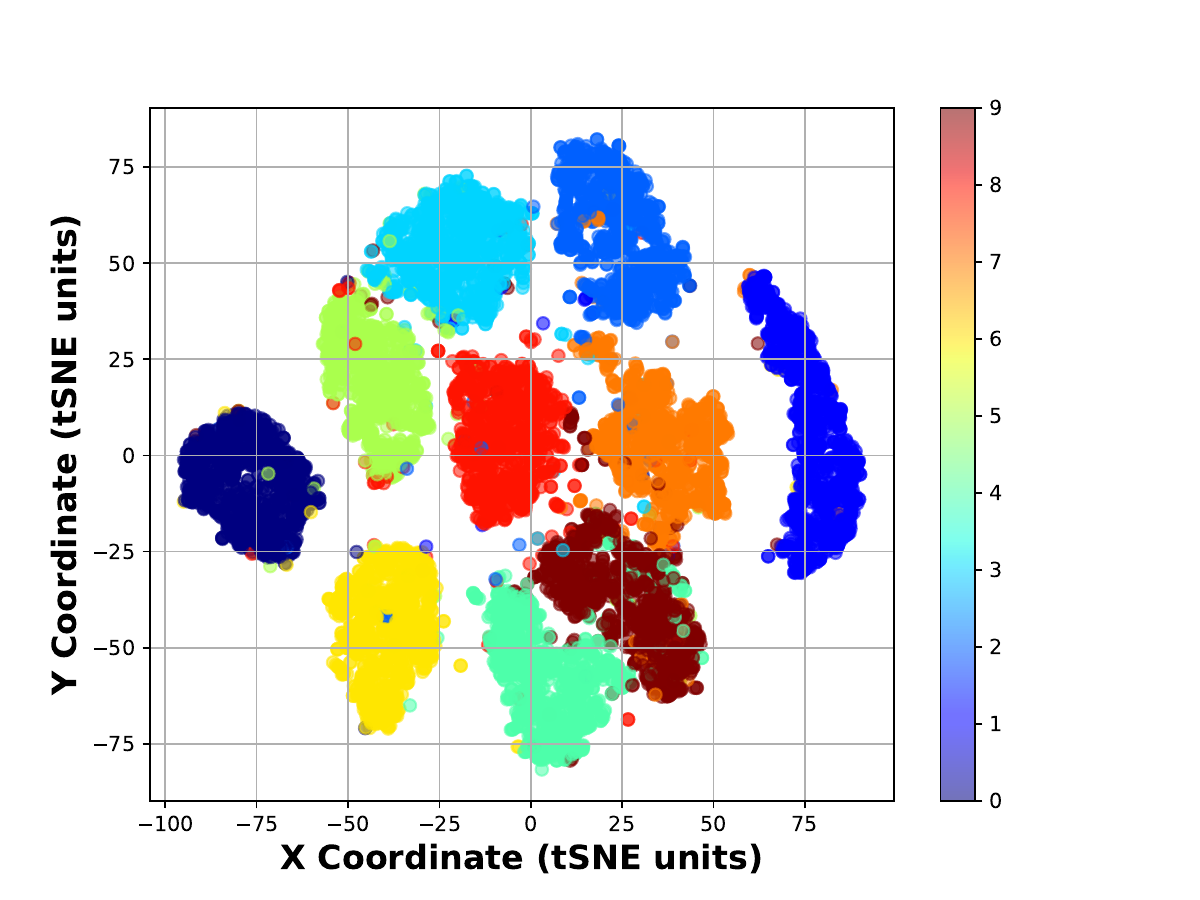}
         \caption{t-SNE of MPC MNIST latents.}
         \label{fig:mpc_mnist_tsne}
     \end{subfigure}
     \begin{subfigure}[b]{0.325\textwidth}
         \centering
         \includegraphics[width=0.75\textwidth]{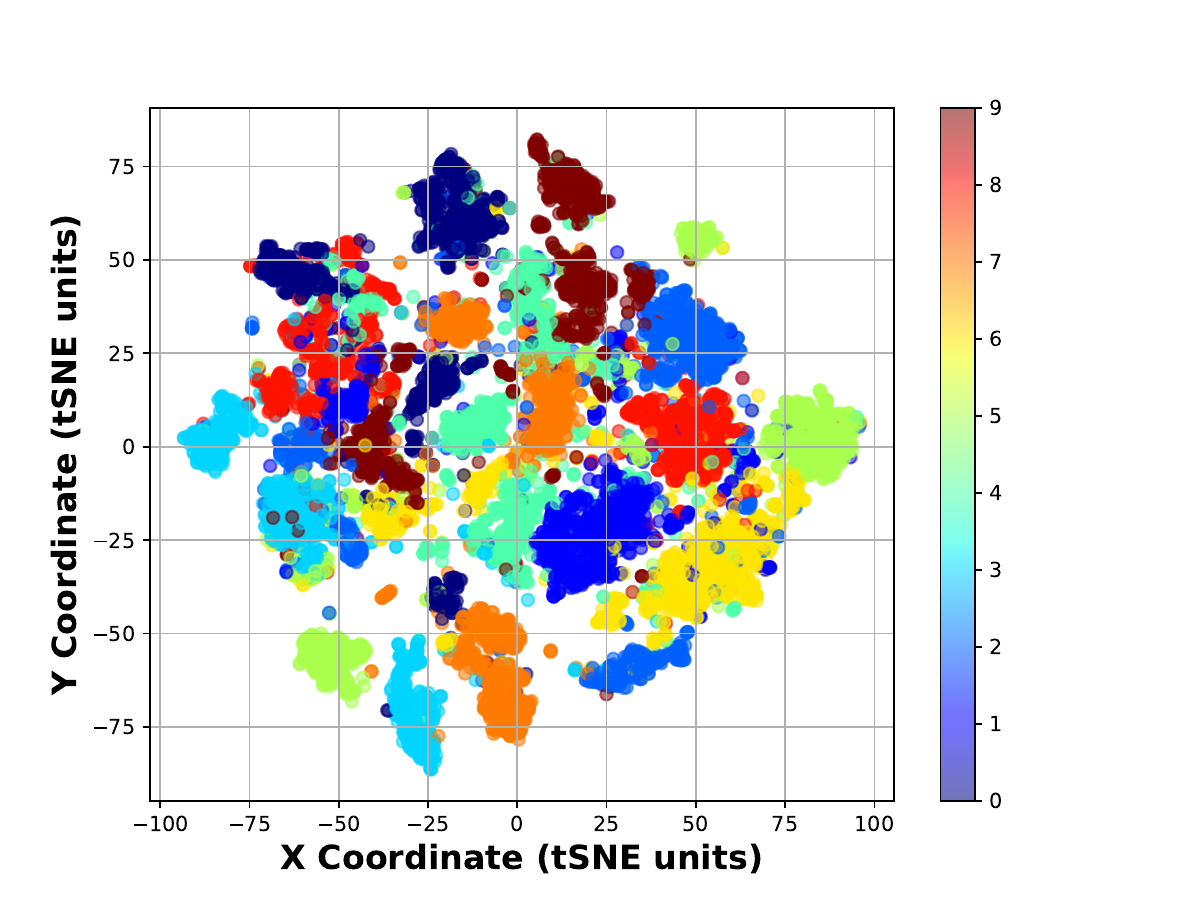}
         \caption{t-SNE of MPC K-MNIST latents.}
         \label{fig:mpc_kmnist_tsne}
     \end{subfigure}
     \begin{subfigure}[b]{0.325\textwidth}
         \centering
         \includegraphics[width=0.75\textwidth]{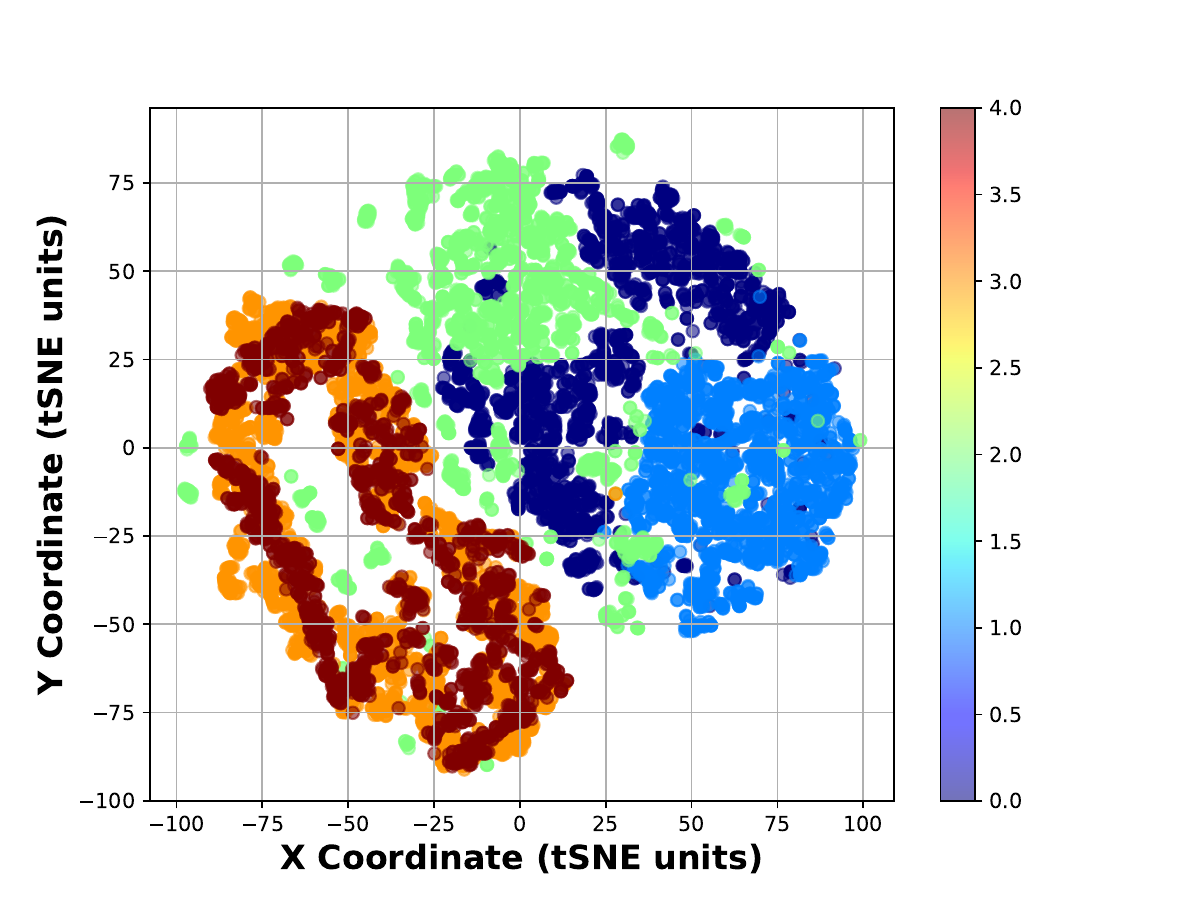}
         \caption{t-SNE of MPC NORB latents.}
         \label{fig:mpc_norb_tsne}
     \end{subfigure}
     \vspace{-0.2cm}
        \caption{ \textbf{Visualization of MPC-acquired latent activity codes.} 
        t-SNE plots of the latent space induced by meta-representational predictive coding (MPC). Rate codes are shown for: 
        (\textbf{A}) MPC on MNIST, and 
        (\textbf{B}) MPC K-MNIST, 
        (\textbf{C}) MPC K-NORB. 
        \emph{Note}: t-SNE coordinate units are dimensionless and are thus denoted as ``tSNE units''). 
        }
        \label{fig:latent_viz}
        \vspace{-0.5cm}
\end{figure}

Beyond examining receptive fields, we visualized the relationship between the latent codes that a trained MPC circuit (MPC-epi) produced across various test-sets. The results of this qualitative analysis used t-SNE \cite{van2008visualizing} to visualize the latent codes formed with respect to the test-set samples; the results of this visualization is presented in Figure \ref{fig:latent_viz}. Notice that, despite never having access to the labels nor using any kind of supervisory signal, the MPC codes for MNIST, K-MNIST, and NORB tend to cluster rather well (albeit a bit noisily in some spots), yielding distinct major representational groupings with each corresponding (item/object) class 
(labels were only used to color the t-SNE mapped latent code points in the t-SNE sub-figures). For K-MNIST, we see that MPC latent codes form groupings as well but there many more groups/clusters or ``sub-groupings''  than the ten identified Kanji character categories; this behavior makes sense, given that the Kanji characters are more complex and exhibit a higher degree of variety than the handwritten digits in the MNIST database. For NORB, we see that clusters representing the five categories emerge though two of these still overlap; this is indicative of the challenge offered by a natural image dataset. 

%%%%%%%%%%%%%%%%%%%%%%%%%%%%%%%%%%%%%%%%%%%%%%%%%%%%%%%%%%%%%%%%%%%%%%%
%%%%%%%%%%%%%%%%%%%%%%%%%%%%%%%%%%%%%%%%%%%%%%%%%%%%%%%%%%%%%%%%%%%%%%%

\section{Latent Glimpse Path Integration and Global Scene Code Creation: Formal Specification}

For any single image, to produce a final, global latent code $\mathbf{z}_K$ (as mentioned in the main text sub-section ``Latent glimpse path integration''), at the end of $K$ glimpses produced by MPC's saccade planner, we designed an iterative path integration scheme that updates a global code (working memory) $\mathbf{z}_k$ with the concatenation of the activities of each stream at step $k$ (meaning that, at any step $k$, we have access to the current state of the global code, i.e., $\mathbf{z}_k$). Our process notably binds, at each step, the content (``what''), i.e., activity magnitudes of the neurons, of each stream with the location they're representing (``where''), i.e., current (input space) location/coordinates.

Let the current state of a glimpse at step $k$ be tracked by an accumulator $S_k$, which receives the current step's encoding vector $\mathbf{z}_k \in \mathbb{R}^{D \times 1}$ (extracted from the MPC's $k$-th glimpse) and its corresponding spatial (2D) coordinate vector $\mathbf{a}_k = \mathbf{c}_k \in \mathbb{R}^{2 \times 1}$, where $\mathbf{c}_k \in [-1, 1]^2$. 
We track and increment, online, the first (mean) and second order (standard deviation) statistical moments via the following running linear and square latent feature (vector) equations below:
\begin{align}
    \mathbf{\Sigma }_{f,k} &= \mathbf{\Sigma }_{f,k-1}+\mathbf{f}_{k} \\ 
    \mathbf{\Sigma }_{f^{2},k} &= \mathbf{\Sigma }_{f^{2},k-1}+\mathbf{f}_{k}^{2}. 
\end{align}
From these accumulation variables, we may dynamically calculate the running sample mean $\mu_k$ and standard deviation $\sigma_k$ in the following manner:
\begin{align}
    \mathbf{\mu }_{k} = \frac{\mathbf{\Sigma }_{f,k}}{k}, \; \text{and}, \; 
    \mathbf{\sigma }_{k} = \sqrt{\max \left(0,\frac{\mathbf{\Sigma }_{f^{2},k}}{k}-\mathbf{\mu }_{k}^{2}\right)}. 
\end{align} 
Note that, within our aggregation scheme, by tracking running vector statistics ($\mathbf{\Sigma}_f$ and $\mathbf{\Sigma}_{f^2}$), MPC dynamically composes and maintains a time-invariant, compressed representation across a saccade trajectory \cite{melcher2015nonretinotopic}.

% multi-scale grid-cell coordinate mapping
We next project the original normalized spatial coordinates $\mathbf{c}_k$ into a high-dimensional, multi-scale periodic phase-space that resembles that induced by neurobiological grid cells \cite{fuhs2006spin}. 
For $M$ geometric frequency scales, a static set of spatial frequencies $\omega_m$ is generated as follows:
\begin{align}
    \omega _{m}=2^{m}\pi ,\quad \text{for\ } m= 0 , 1 , \dots , M-1
\end{align}
and the phase angles along the horizontal and vertical axes are mapped through sinusoidal and cosinusoidal projections in order to construct a flat grid-cell coordinate vector $\mathbf{m}_k \in \mathbb{R}^{G \times 1}$, where $G = 4M$, as follows:
\begin{align}
    \mathbf{m}_{k}=\Big< \big\{ \big<\sin (\omega_{m} \mathbf{c}_{k}),\,\cos (\omega_{m} \mathbf{c}_{k})\big> \big\}_{m=0}^{M-1} \Big>_{\text{col}}
\end{align}
where $<\cdot>_{\text{col}}$ denotes the concatenation (along the column dimension) of all elements in a set $\{\cdot\}$. 
% grid-cells and path integration
Note that the production of $\mathbf{m}_k$ emulates, at a high-level, the periodic, multi-scale firing fields of grid-cells found within the medial entorhinal cortex (MEC) \cite{hafting2005microstructure,fuhs2006spin,stensola2016grid}. These grid-cell units functionally can be viewed as providing the MPC circuit with an internal ``GPS'' within the current scene; this maps spatial trajectories through periodic, hexagonal tessellations across multiple spatial frequency scales. % update? 
Through using geometric powers of two, i.e., $2^m\pi$, our aggregator implements a multi-scale spatial frequency map that structures/organizes the relational distances with the MPC's visual coordinate space; this, in effect, acts a visual path integrator. 

%%%%%%%%%%%%%%%%%%%%%%%%%%%%%%%%%%%%%%%%%%%%%%%%%%%%%%%%%%%%%%%%%%%%%%%%%%%%%%%%%%%%
%% glimpse sequence and foveal/peripheral views
\begin{wrapfigure}{r}{0.5\textwidth}
\vspace{-0.4cm}
  \begin{center}
    \includegraphics[width=0.3\textwidth]{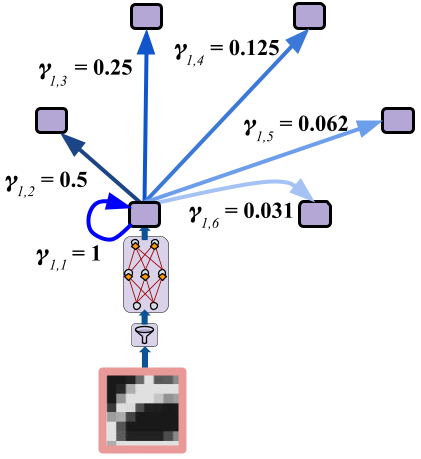}
  \end{center}
  \caption{
  \small{
  Illustration of the MPC (chain) topology weighting for $V=6$ streams; here we show the weighting coefficients for stream $v=1$ out of the set $q = \{1,2,3,4,5,6\}$, yielding its set of relative weights $\{ \gamma_{1,1}, \gamma_{1,2}, \gamma_{1,3}, \gamma_{1,4}, \gamma_{1,5}, \gamma_{1,6} \}$.
  }
  }
  \label{fig:mpc_topology}
  %\vspace{-0.3cm}
\end{wrapfigure}
%%%%%%%%%%%%%%%%%%%%%%%%%%%%%%%%%%%%%%%%%%%%%%%%%%%%%%%%%%%%%%%%%%%%%%%%%%%%%%%%%%%%  

% spatio-temporal phase-centroid binding
To combine the latent visual features with their specific spatial origins, our aggregation process further maintains a continuous, outer-product weight accumulator matrix $\mathbf{C}_k \in \mathbb{R}^{D \times G}$ (along with a normalization/divisive factor  $\Sigma_{w,k} \in \mathbb{R}^{D \times 1}$) through an incremental outer product step. 
This operation works to systematically bind the magnitude of each (extracted) latent feature to its corresponding multi-scale periodic grid phase coordinate (across the MPC circuit's saccadic horizon)
This step is performed as follows:
\begin{align}
    \mathbf{\Sigma }_{w,k} &= \mathbf{\Sigma }_{w,k-1}+|\mathbf{f}_{k}| \\ 
    \mathbf{C}_{k} &= \mathbf{C}_{k} + |\mathbf{f}_k| \cdot (\mathbf{m}_k)^T
    %\mathbf{C}_{i,j} &= \mathbf{C}_{i,j} + |[\mathbf{f}_k]_i| \cdot [\mathbf{m}_k]_j
    % \mathbf{C}_{i, j} &= \mathbf{C}_{i, j} + \vert{}\mathbf{f}_{k}\vert{} \cdot \mathbf{m}_{i, \cdot} .
\end{align}
where $|\cdot|$ is the absolute value operator. 
The above running phase-centroids are next normalized % (by their cumulative feature weights) 
and flattened into a final vector $\mathbf{\psi}_k \in \mathbb{R}^{B \times (D \cdot G)}$: 
\begin{align}
    \mathbf{\psi }_{k}=\text{Flatten}\left(\frac{\mathbf{C}_{k}}{\mathbf{\Sigma }_{w,k}+\epsilon }\right) . 
\end{align}
% feature-location binding (ventral-dorsal stream integration)
Our aggregator's running calculation of phase-centroids, by modulating the multi-scale (grid) vector $\mathbf{m}_k$ with the (absolute value of) latent feature magnitudes $|\mathbf{f}_k|$, creates an internal coordinate map \cite{hafting2005microstructure} for the MPC circuit, binding specific latent features (``what'') directly to the very spatial coordinates from which they were observed (``where''). Note that this scheme offers a computational solution to the well-known what-where binding problem \cite{treisman1996binding} of connecting an object's visual appearance, processed in the ventral ``what'' pathway, to its precise location in space, processed in the dorsal ``where'' pathway. 

% final global code assembly
Finally, each of these three aggregate vector (sub-)codes are divided by their Euclidean norms (so as to ensure scale invariance across these three sources of information before final global code composition) as in below:
\begin{align}
    \hat{\mathbf{\mu }}_{k} &= \frac{\mathbf{\mu }_{k}}{\|\mathbf{\mu }_{k}\|{}_{2}+\epsilon },\quad \hat{\mathbf{\sigma }}_{k}=\frac{\mathbf{\sigma }_{k}}{\|\mathbf{\sigma }_{k}\|{}_{2}+\epsilon },\quad \hat{\mathbf{\psi }}_{k}=\frac{\mathbf{\psi }_{k}}{\|\mathbf{\psi }_{k}\|{}_{2}+\epsilon },  \\ 
    &\text{and, finally, } \mathbf{z}_{k} = \left[\hat{\mathbf{\mu }}_{k},\,\hat{\mathbf{\sigma }}_{k},\,\hat{\mathbf{\psi }}_{k}\right] . 
\end{align} 
%% conclusion
Overall, our aggregation dynamically updates an organized, structural map of the geometry of visual input stimuli as opposed to compiling a loose collection of visual features.

\section{Inter-Stream Topology-Weighting Details}

In the main definition of MPC's ensemble free-energy (Equation 1), %\ref{eqn:rpc_vfe}),
across $V$ total different neural streams, each inter-stream prediction -- stream $v$'s prediction of stream $q$ -- is weighted by a coefficient $\gamma_{v,q}$. 
While the coefficient for stream $v$ predicting its own latent activity is $\gamma_{v,v}=1$, the values for the other streams relative to stream $v$ are computed via the following (wrapped) power kernel:
\begin{align}
    \gamma_{v, q} = \alpha^{(q - v) \bmod V}
    %\alpha_{v,q} = \alpha^{(k - k_q)}, \text{where } \alpha=0.5
\end{align}
where we observe that for each stream $v$, its own intra-stream prediction weight is $1$ while each subsequent inter-stream prediction gradually decays towards $0$ the ``farther away'' or more distal the target stream $q \neq v$ becomes. This down-weighting of inter/cross-stream prediction energies improves overall stability of the MPC evolution process.

% are produced by applying a Gaussian kernel to the coordinates (within a topology) assigned to each stream. We choose to construct a simple ring topology, where each stream was placed at one particular point in the ring.  
% In other words, if stream $v$ has Cartesian coordinates $(a, b)$ and stream $q$ has Cartesian coordinates $(c, d)$, we calculate the inter-stream weighting coefficient to be a function of their Euclidean distance, formally as follows: 
% \begin{align}
%     \gamma_q = k(d_y, d_x) = \exp\Big( -\frac{d^2_y + d^2_x}{2 \sigma^2_{\text{top}}} \Big)
% \end{align}
% and $d_y = b - d$ and $d_x = a - c$.

\section{Saccade Planning: Qualitative MNIST Samples}

%%%%%%%%%%%%%%%%%%%%%%%%%%%%%%%%%%%%%%%%%%%%%%%%%%%%%%%%%%%%%%%%%%%%%%%
%% MPC fov glimpse sequence for digit 7
\begin{figure}[!t]
    \begin{center}
    \tabcolsep=0.05cm
    \begin{tabular}{ c c c c c c} 
    \midrule
    & \multicolumn{1}{l}{\includegraphics[width=0.15\textwidth]{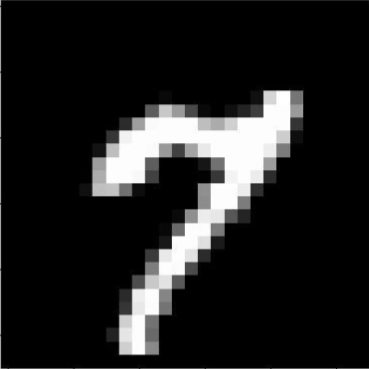}} &  & \multicolumn{1}{c}{\includegraphics[width=0.3\textwidth]{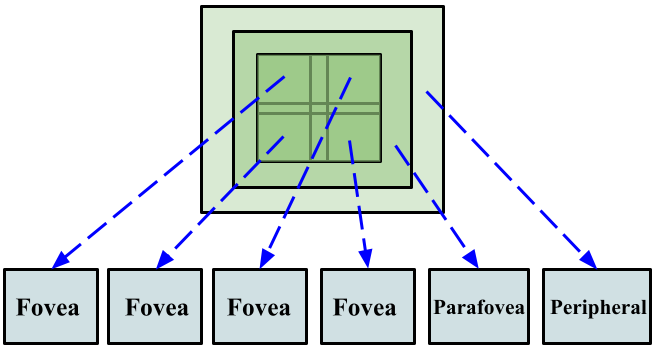}} & & \\
    \multicolumn{1}{p{0.055cm}}{\vspace{-0.9cm}\textbf{1}} &  \includegraphics[width=0.3\textwidth]{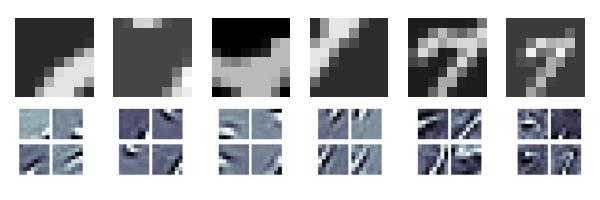} & 
    \multicolumn{1}{p{0.055cm}}{\vspace{-0.9cm}\textbf{2}} &
    \includegraphics[width=0.3\textwidth]{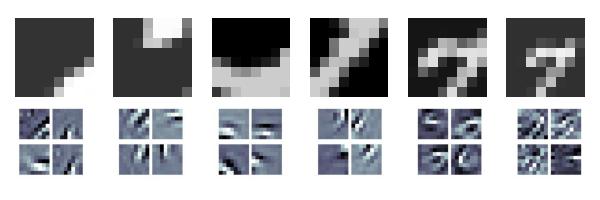} & 
    \multicolumn{1}{p{0.055cm}}{\vspace{-0.9cm}\textbf{3}} & 
    \includegraphics[width=0.3\textwidth]{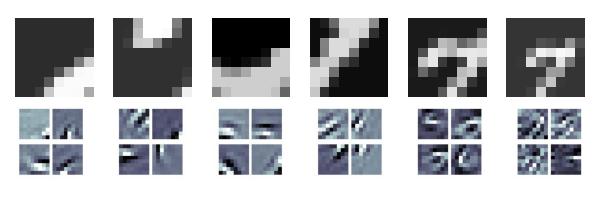} \\ 
    \multicolumn{1}{p{0.055cm}}{\vspace{-0.9cm}\textbf{4}} &  \includegraphics[width=0.3\textwidth]{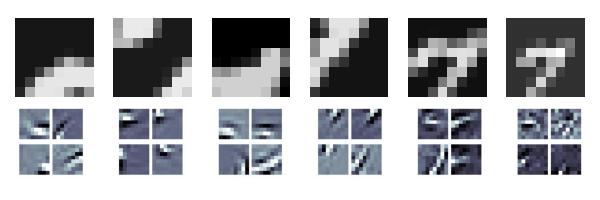} & 
    \multicolumn{1}{p{0.055cm}}{\vspace{-0.9cm}\textbf{5}} &
    \includegraphics[width=0.3\textwidth]{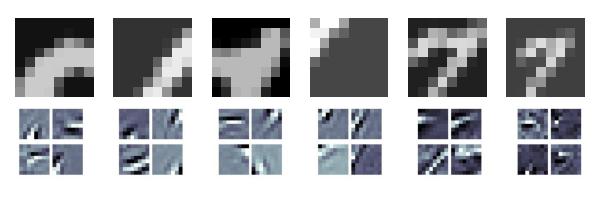} & 
    \multicolumn{1}{p{0.055cm}}{\vspace{-0.9cm}\textbf{6}} & 
    \includegraphics[width=0.3\textwidth]{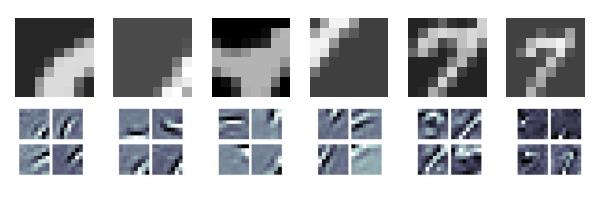} \\ 
    \multicolumn{1}{p{0.055cm}}{\vspace{-0.9cm}\textbf{7}} &  \includegraphics[width=0.3\textwidth]{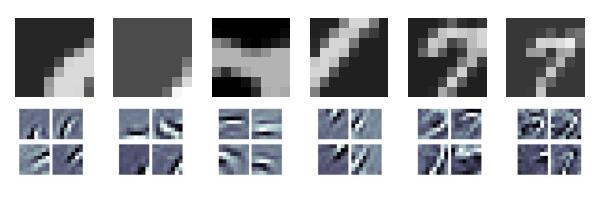} & 
    \multicolumn{1}{p{0.055cm}}{\vspace{-0.9cm}\textbf{8}} &
    \includegraphics[width=0.3\textwidth]{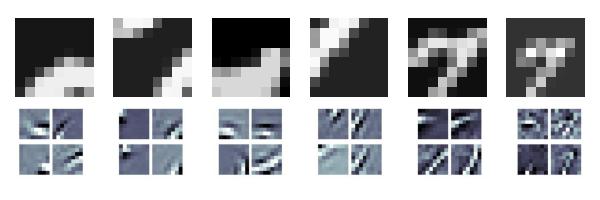} & 
    \multicolumn{1}{p{0.055cm}}{\vspace{-0.9cm}\textbf{9}} & 
    \includegraphics[width=0.3\textwidth]{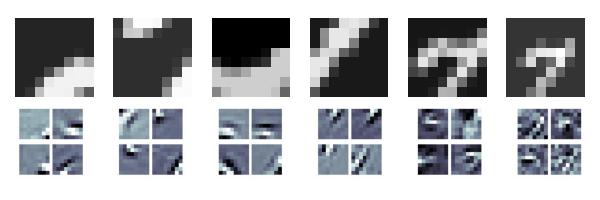} \\
    \midrule
    & \multicolumn{1}{l}{\includegraphics[width=0.15\textwidth]{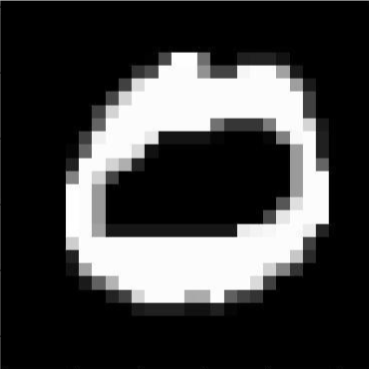}} & & \multicolumn{1}{c}{\includegraphics[width=0.3\textwidth]{figs/mpc_fov_reference.png}} & & \\ 
    \multicolumn{1}{p{0.055cm}}{\vspace{-0.9cm}\textbf{1}} &  \includegraphics[width=0.3\textwidth]{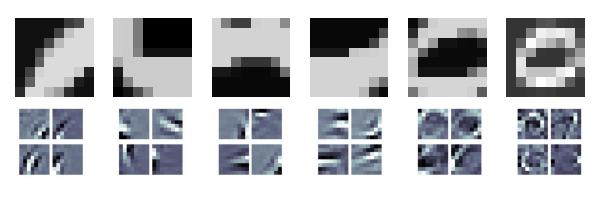} & 
    \multicolumn{1}{p{0.055cm}}{\vspace{-0.9cm}\textbf{2}} &
    \includegraphics[width=0.3\textwidth]{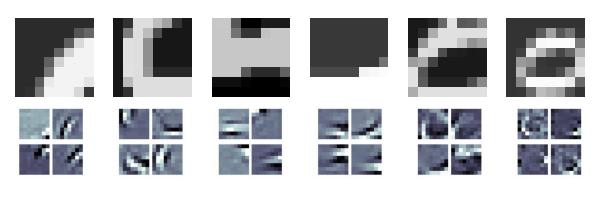} & 
    \multicolumn{1}{p{0.055cm}}{\vspace{-0.9cm}\textbf{3}} & 
    \includegraphics[width=0.3\textwidth]{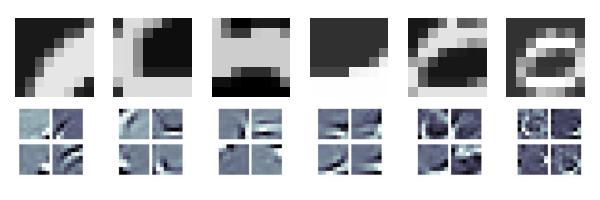} \\ 
    \multicolumn{1}{p{0.055cm}}{\vspace{-0.9cm}\textbf{4}} &  \includegraphics[width=0.3\textwidth]{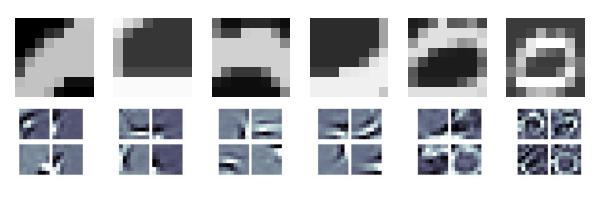} & 
    \multicolumn{1}{p{0.055cm}}{\vspace{-0.9cm}\textbf{5}} &
    \includegraphics[width=0.3\textwidth]{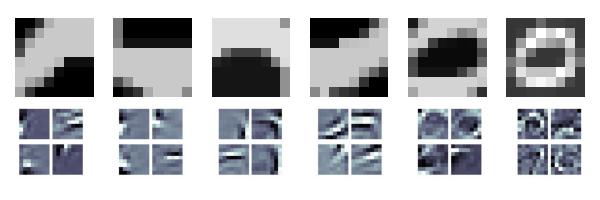} & 
    \multicolumn{1}{p{0.055cm}}{\vspace{-0.9cm}\textbf{6}} & 
    \includegraphics[width=0.3\textwidth]{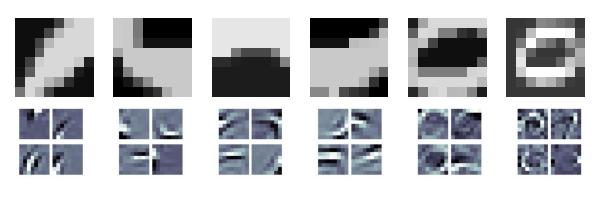} \\ 
    \multicolumn{1}{p{0.055cm}}{\vspace{-0.9cm}\textbf{7}} &  \includegraphics[width=0.3\textwidth]{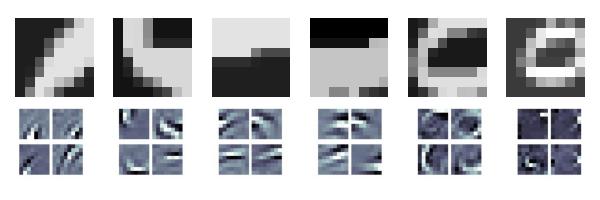} & 
    \multicolumn{1}{p{0.055cm}}{\vspace{-0.9cm}\textbf{8}} &
    \includegraphics[width=0.3\textwidth]{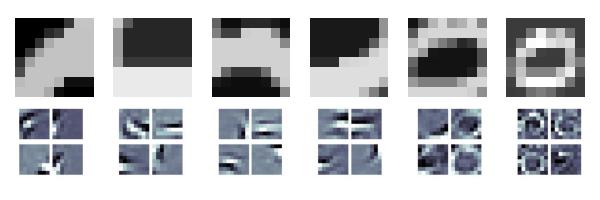} & 
    \multicolumn{1}{p{0.055cm}}{\vspace{-0.9cm}\textbf{9}} & 
    \includegraphics[width=0.3\textwidth]{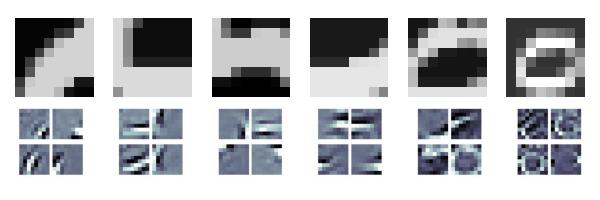} \\
    \hline
    \end{tabular}
    \end{center}
    \caption{ 
    \small{
    \textbf{A trained MPC scheme processing sensory inputs through a saccade sequence.} 
    Shown is a trained MPC scheme iteratively processing a sensory stimulus, e.g., an image of a digit seven (top group of four rows) or a zero (bottom group of four rows), through a series of (randomly selected) saccades. Within each group of rows, which pertain to a particular input digit, the first row shows a global view of the original input for reference, while the other three rows show the saccade sequence taken by the MPC scheme (the bold number indicates the saccade step $k$, out of nine total taken, corresponding to a particular saccade-sampled view). Within each saccade view, the first half of the image shows the sensory-level glimpses (at step $k$) while the bottom half contains the top four most highly activated receptive fields extracted from the MPC circuit in response to the input stimulus. 
    }
    }
    \label{fig:mpc_seq_mnist}
    \vspace{-0.5cm}
\end{figure}
%%%%%%%%%%%%%%%%%%%%%%%%%%%%%%%%%%%%%%%%%%%%%%%%%%%%%%%%%%%%%%%%%%%%%%%

Given MPC's iterative nature, when sampling sensory input over the course of several saccades, in Figure \ref{fig:mpc_seq_mnist}, we asked what an MPC scheme is doing as it processes a sensory stimulus throughout the course of nine glimpses. Specifically, we examined two different digit patterns, i.e., a seven (in the top of Figure \ref{fig:mpc_seq_mnist}) and a zero (in the bottom of Figure \ref{fig:mpc_seq_mnist}). In addition to depicting the raw sensory glances produced by our sensory glimpsing scheme, we show the top four most activated receptive fields that each neuronal stream yields in response to the observation of a glimpse (at each time step). Notice that, for the sensory glimpse produced by each saccade, the most highly-activated foveal receptive fields capture particular, essential characteristics of the examined input, e.g., the rotation/orientation of stroke/edge of the overall digit pattern, while the most highly-activated parafoveal/peripheral fields correspond to either: 
\textbf{1)} capturing broader feature shapes/profiles (lower-resolution strokes and their orientations), or, 
\textbf{2)} engaging in a form of template matching to the most relevant low-resolution object ``chunk''. In some instances, such as for the foveal streams, seemingly non-related features can appear among the more highly-activated fields, such as a stroke or edge piece that just happens to fit ``within'' the stimulus area of the general feature. 

\section{Representational Analysis of MPC's Glimpse Codes}

To study our MPC's information processing ability, we carried out several experiments to understand its representational robustness and stability. 

% effect of foraging hard boundaries 
To characterize potential representational robustness of our epistemic planning scheme for downstream tasks, we systematically evaluated a range of coordinate search boundaries. On the lower end, we study a restrictive $[-0.2, 0.2]$ prior and, on the higher end, we consider an unconstrained $[-0.95, 0.95]$ canvas across the full image test set of %MNIST ($10,000$ samples) and 
NORB ($48,600$ samples). 
For each image pattern, we collect frozen latent codes integrated across the MPC glimpse trajectories and subsequently probe categorical separability via a downstream $K$-NN non-parametric classifier (we use $K=3$ neighbors and Minkowski distance with $p=2$). 

\begin{figure}[!t]
     \centering
     \begin{subfigure}[b]{0.485\textwidth}
         \centering
         \includegraphics[width=0.95\textwidth]{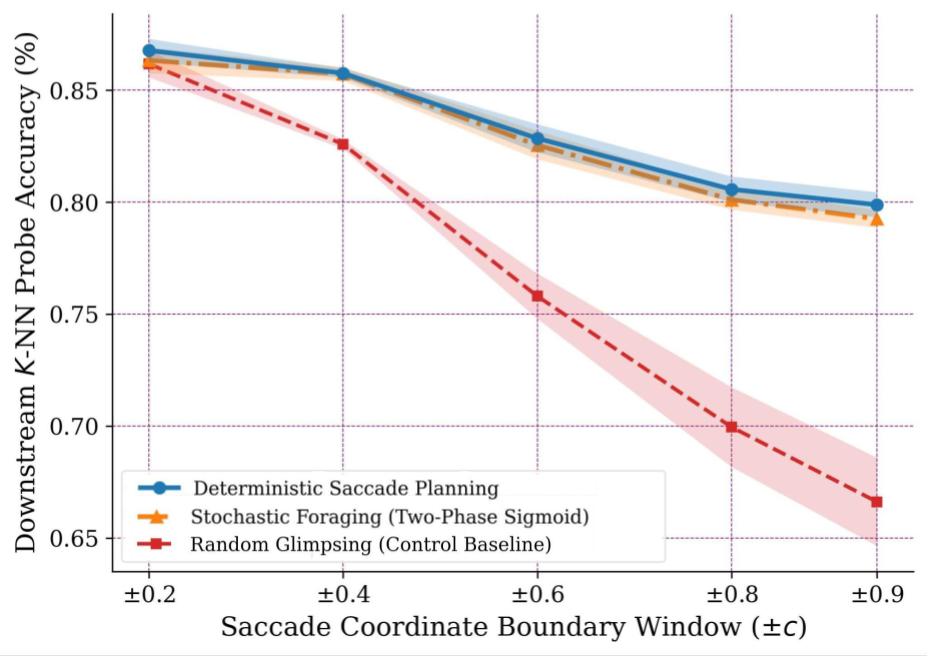}
         \caption{Representational stability under extended sensory horizons.} %NORB KNN Accuracy versus Boundary Shape.
         \label{fig:knn_vs_boundary_shape}
     \end{subfigure}\\
     \begin{subfigure}[b]{0.485\textwidth}
         \centering
         \includegraphics[width=0.95\textwidth]{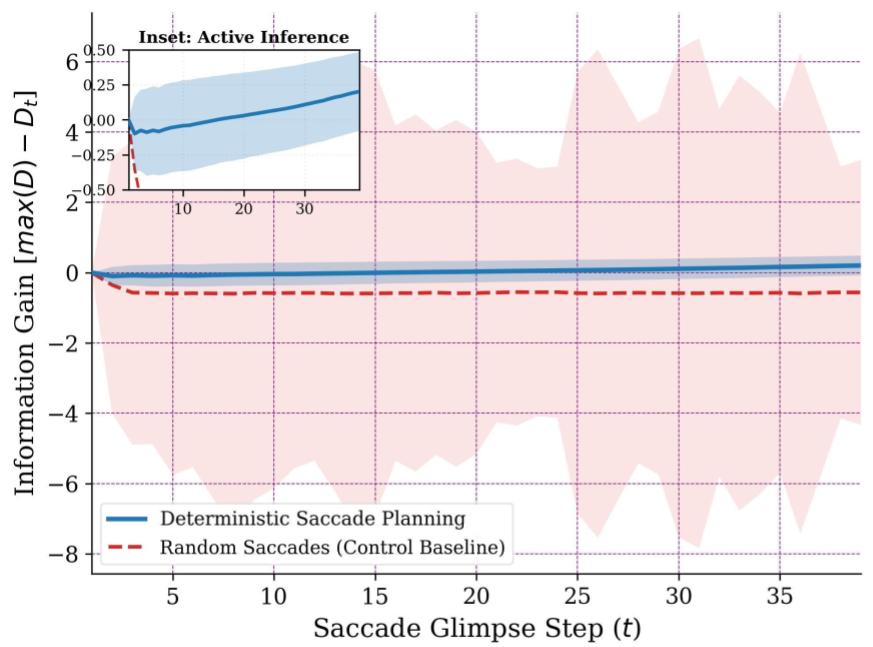}
         \caption{Epistemic saccade process efficiency.}
         \label{fig:epistemic_saccade_efficiency}
     \end{subfigure}
     \begin{subfigure}[b]{0.485\textwidth}
         \centering
         \includegraphics[width=0.95\textwidth]{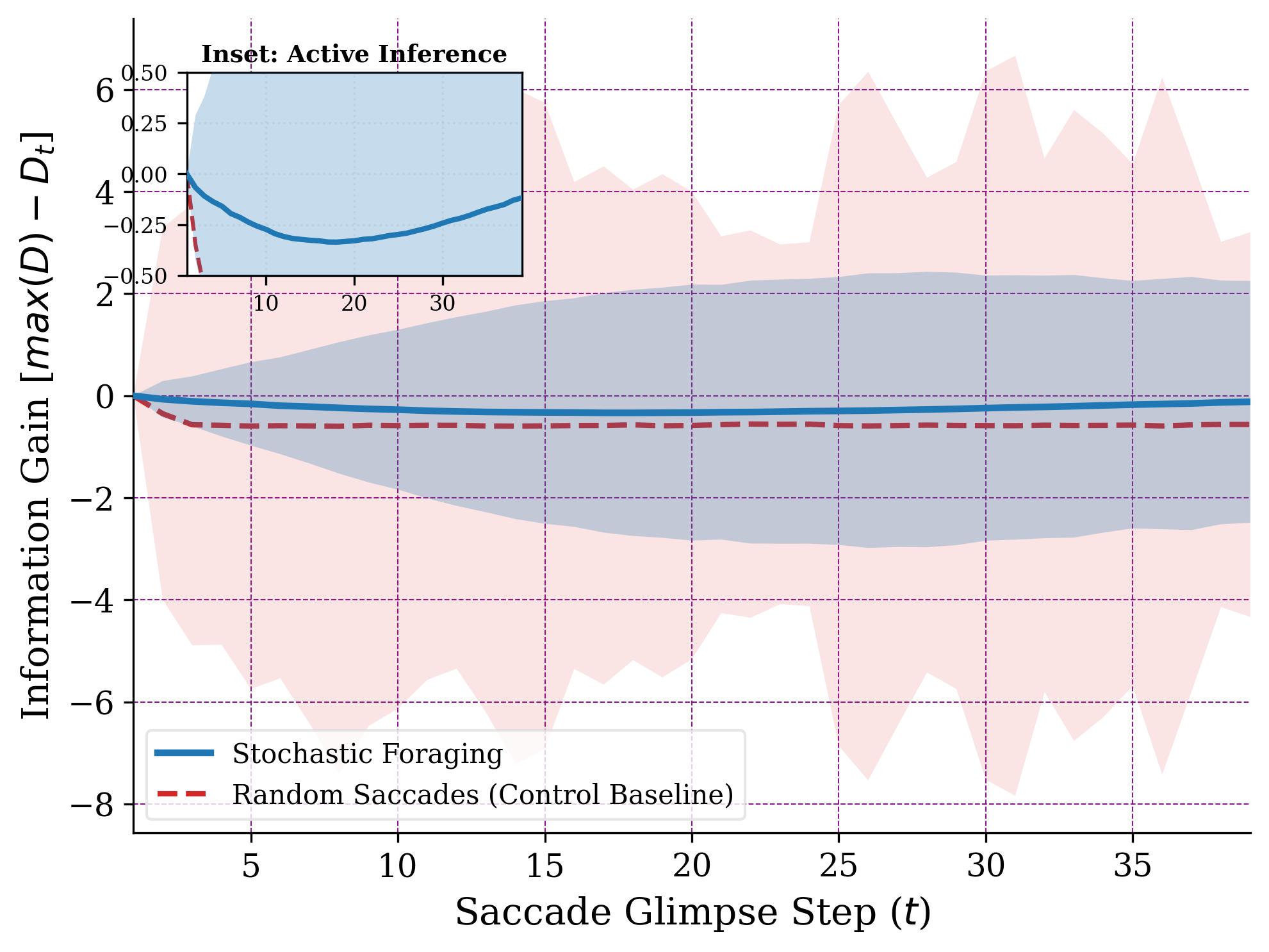}
         \caption{Stochastic foraging process efficiency.}
         \label{fig:stochastic_foraging_efficiency}
     \end{subfigure}
        \caption{ 
        \small{
        \textbf{Glimpse trajectory representation stability and efficiency (on NORB).} 
        In (panel \textbf{a}), we measure our MPC model's downstream KNN probe test accuracy as a function of its maximal allowable saccade boundary window size. 
        Notice that the KNN performance declines far less (only to about $80$\%) when using active perception (i.e., a saccade planning scheme) as opposed to merely random saccades (degradation to below $70$\%). 
        In (panel \textbf{b}), we study the epistemic saccade (deterministic) planner's step-by-step information gain (IGn) curve in comparison with the random saccade process step-by-step IGn curve. Here, we see how our MPC's target-focused %, high-predictibility 
        ocular-motor policy operates; it first prioritizes immediately hunting down local edges (local edge stabilization) before finally sweeping outwards to obtain supporting context information. This explains this scheme's higher KNN accuracy on the maximal $[-0.95, 0.95]$ glimpse boundary conditions. 
        %% NOTE: add comment on stoch foraging's wider variance...(price for exploration)
        In (panel \textbf{c}), we depict the stochastic foraging variant of our saccade planner's step-by-step IGn curve, showing how our two-phase sigmoid scheduled process works, which offers an exploratory, uncertainty-depleting oculomotor policy. Note that stochastic foraging trades off a small fraction of immediate latent code stability early on its foraging in order to aggressively acquire and map out as much information from its (global) sensory environment, dissolving structural loops and generalizing to wider boundaries. 
        \emph{Note}: for both (panels \textbf{b}) and (\textbf{c}), we visualize the mean curve and standard deviation envelop over $N=48600$ test NORB samples.
        }
        }
        \label{fig:saccade_process_efficiency}
        \vspace{-0.5cm}
\end{figure}

Observe in Figure \ref{fig:knn_vs_boundary_shape}, a restrictive spatial boundary $[-0.2, 0.2]$ operates as a strong geometric ``crop prior'', which artificially inflates the $K$-NN performance of random saccades to $~\sim 85$\%. However, as the sensory horizon/boundary expands, the random saccade (control) baseline exhibits a sharp decay, ultimately plummeting to $\sim 61$\% at the maximal boundary of $0.95$ due to ambient background noise that comes with this increased canvas/view. 
Conversely, the epistemic saccade process/planner exhibits a strong resilience. Notably, this reflex engine maintains a high, overall stable plateau, preserving categorization accuracy at $80.5$\% under the widest/maximal boundary conditions. 
In addition, observe that our stochastic foraging scheme (the two-phase sigmoidal stochastic process) well matches this trend; it only trails the deterministic epistemic saccade process by only a $\sim 1-2$\% margin, i.e., $79.8$\%. This rather minor divergence highlights a fundamental (information-theoretic) trade-off; the controlled noise (from the Gumbel perturbations) optimize the stochastic foraging planner for spatial exploration (and uncertainty resolution) at a small cost to classification quality of the manifold induced by the MPC latent codes. Collectively, these results demonstrate that our simple form of reflex-based active inference allows our self-supervised MPC encoder from handcrafted, geometric bounding box constraints and instead relies on only autonomous, active perception to track features that preserve representational stability. 
%We corroborate this result with an additional analysis conducted with respect to representational stability. 

%%%%%%%%%%%%%%%%%%%%%%%%%%%%%%%%%%%%%%%%%%%%%%%%%%%%%%%%%%%%%%%%%%%%%%%
%% NORB analysis
%% KNN probe plots as a function of glimpse boundary prior conditions; also, IG plots for saccade process schemes
%%%%%%%%%%%%%%%%%%%%%%%%%%%%%%%%%%%%%%%%%%%%%%%%%%%%%%%%%%%%%%%%%%%%%%%

In Figure \ref{fig:saccade_process_efficiency}, we further probed our representational stability by examining the instantaneous (or step-by-step) information gain across a glimpse trajectory induced by our epistemic saccade process in comparison to random saccades (over a fixed budget of $T = 40$ glimpses) on the full test-set of NORB. This was concretely measured via $\text{info}(\mathbf{o})_t = (\text{info}(\mathbf{o})_{max} - ||\mathbf{z}_t, \mathbf{z}_{t+1}||)/\text{info}(\mathbf{o})_{max}$ where $\text{info}(\mathbf{o})_{max} = ||\mathbf{z}_0, \mathbf{z}_{1}||_2$.  
When plotting out $\text{info}(\mathbf{o})_t$ over time, a curve of values that climbs above zero and remains positive means that the model has reached a (steady-)state where each glimpse is less shocking than the baseline shock $\text{info}_{max}(\mathbf{o})$ (indicating that a model has obtained instantaneous representational stabilization). 
Across the full test-set of samples, in (sub-)Figures \ref{fig:epistemic_saccade_efficiency} and \ref{fig:stochastic_foraging_efficiency}, we plot the mean information gain curve, paired with its standard deviation envelope, for each condition. Notice that the deterministic epistemic saccade process as well as the stochastic forager result in an overall upward-sloping trend, with information gain values generally climbing over the zero-mark; this behavior indicates a systematic stabilization of MPC's latent manifold. The random saccade process, in contrast, remains rather flat in its trend and generally falls below the zero mark (i.e., it becomes trapped in a ``low-performance state'', buffeted by new, jolting/chaotic spikes in information, causing representational shift).

\section{Glimpse Vector Creation: Additional Details}

Each glimpse-action pair within a $K$-length (random/involuntary) saccade trajectory or sequence $\{(\mathbf{g}(0), \mathbf{a}(0)), (\mathbf{g}(1), \mathbf{a}(1)),...,(\mathbf{g}(k), \mathbf{a}(k)),...(\mathbf{g}(K), \mathbf{a}(K))\}$ is created by first generating a 2D vector contain Cartesian x-y coordinates (in this work, a random policy is used to select each set of coordinates at each step within the trajectory), extracting the relevant glimpse vector $\mathbf{g}(k)$ containing the foveal/parafoveal/peripheral patches at the chosen coordinates, and then finally normalizing the coordinate vector to create the requisite action vector $\mathbf{a}(k)$. Specifically, the 2D x-y coordinate action vector is created by normalizing the original raw x-y Cartesian coordinates to the range of $[-1, 1]$, i.e., $\mathbf{a}(k) = 2\Big([x, y]^T\Big)/D - 1$ for a $D \times D$ ($O_W = O_W=D$, $Z=1$) pixel image where $x$ and $y$ are the original Cartesian coordinates of the glimpse center-point.

%% patch creation details
Each glimpse vector $\mathbf{g}(k)$ itself is a concatenation of several views (pixel patches) sampled from the observation $\mathbf{o}(t_g)$. Specifically, it is a combination of several types of views% ($C$ ``foveal'' views, $F$ ``parafoveal'' views, and $P$ ``peripheral'' views)
, each of which can be expressed in terms of the following piecewise function:
\begin{align}
    \mathbf{p}^v = 
    \begin{cases} 
      \mathbf{p}^v_c \in \mathbb{R}^{S_c \times S_c} & v \in \{\text{Set of foveal patch indices}\} \\
      \mathbf{p}^v_f \in \mathbb{R}^{S_f \times S_f} & v \in \{\text{Set of parafoveal patch indices}\} \\
      \mathbf{p}^v_p \in \mathbb{R}^{S_p \times S_p} & v \in \{\text{Set of peripheral patch indices}\} \\
      \emptyset & \text{otherwise.}
   \end{cases}
\end{align}
% \begin{itemize}[noitemsep,nolistsep]
%     \item $C$ ``foveal'' views, each of shape $S_C \times S_C$ pixels, i.e., the $c$-th foveal view is $\mathbf{p}_c \in \mathbb{R}^{S_C \times S_C}$; 
%     \item $F$ ``parafoveal'' views, each of shape $S_F \times S_F$ pixels, i.e., the $f$-th parafoveal view is $\mathbf{p}_f \in \mathbb{R}^{S_F \times S_F}$;  and, 
%     \item $P$ ``peripheral'' views, each of shape $S_P \times S_P$ pixels, i.e., the $p$-th peripheral view is $\mathbf{p}_p \in \mathbb{R}^{S_P \times S_P}$. 
% \end{itemize}
%We choose $C = 4$ (four overlapping foveal views, arranged in a $2 \times 2$ grid), $F = 1$ (one parafoveal view), and $P = 1$ (one peripheral view). 
To produce the final glimpse tensor, all $C$ foveal, $F$ parafoveal, and $P$ peripheral views (centered around the glimpse/gaze's center-point) are first average pooled to always be the same final shape of $S \times S$ pixels, then flattened to vectors, and finally concatenated to obtain $\mathbf{g}(k) \in \mathbb{R}^{( (C + F + P) * (S * S) ) \times 1}$. 
All of the above means that the final glimpse vector produced as a result of the $k$-th saccade is:
\begin{align}
    \mathbf{g}(k) = (<\mathbf{g}^1(k), \mathbf{g}^2(k),...,\mathbf{g}^v(k),...,\mathbf{g}^V(k)  >)^T
\end{align}
where $V = C + F + P$, brackets $<\cdot>$ denote vector concatenation, and $\mathbf{g}^v(k) = \text{Flat}\big(\text{Pool}(\mathbf{p}^v)\big)$. For the main model used in the paper, the set of foveal patch indices was $\{1,2,3,4\}$ while the set of parafoveal indices was $\{5\}$ and the peripheral was $\{6\}$. This choice was made based on preliminary experimentation (the stream configuration that resulted in the best performing MPC model was chosen), the results of which are presented in the later in this appendix.  
% specifically, indices $v = 1, 2, 3, 4$ would correspond to flattened foveal views $\mathbf{g}^v(k) = \text{Flat}(\mathbf{p}^v)$, index $v = 5$ would correspond to a flattened parafoveal view $\mathbf{g}^v(k) = \text{Flat}(\mathbf{p}^f)$, and index $v = 6$ would correspond to a flattened peripheral view $\mathbf{g}^v(k) = \text{Flat}(\mathbf{p}^p)$. 

Upon creation, the foveal, parafoveal, and peripheral views are all aligned around a glimpse center-point (the x-y coordinates of the center-point of our model's gaze) with all views arranged around the center-point in a particular topology (such as a grid). 
Foveal views are generally shaped such that $S_C = S$ whereas parafoveal views are shaped such that $S_C > S$ and peripheral views are shaped such that $S_P > S_C > S$. 
In this work, we specifically choose for each dataset the following configurations: 
\begin{itemize}[noitemsep,nolistsep]
    \item \emph{MNIST and K-MNIST}: foveal views $S = S_C = 12$ pixels, for parafoveal views $S_F = 18$ pixels, and for peripheral views $S_P = 24$ pixels; 
    \item \emph{NORB}: we chose $S = S_C = 24$, $S_F = 48$, and $S_P = 72$; and, 
    \item \emph{ETH-80}: we chose $S = S_C = 14$, $S_F = 28$, and $S_P = 36$.
\end{itemize}
As mentioned above, we use $C=4$ foveal views which are arranged in a $2 \times 2$ grid (such that the foveal views overlap with one another by $1$-$2$ pixels) centered around the whole glimpse/gaze center-point; we only use one ($F=1$) parafoveal view and one ($P=1$) peripheral views, which are both directly centered around the glimpse/gaze x-y Cartesian center-point. %See the patches within the red dot-dashed box in Figure \ref{fig:saccade_scheme_example} for an example of the four foveal, one parafoveal, and one peripheral patch view we extract at the $k$-th saccade step. 

\section{Stochastic Foraging Temperature Schedule: Dual-phase Sigmoidal Decay.} 

%govern the exploration temperature decay
For the stochastic foraging variant of the MPC saccade planner, the temperature value $\mathcal{T}$, which scales the Gumbel-Max selection noise injected into the coordinate selection process, is dynamically adjusted across a saccade trajectory. 
For the current discrete fixation step $k \in \{0, 1, \dots, K-1\}$, the (temporal) progression fraction $\rho$ is bounded to be: $\rho =\frac{k}{K-1}$. This means we design a schedule \cite{kirkpatrick1983optimization} $\mathcal{T} = \mathcal{T}(\rho)$ where the chosen temperature decays according to an inverted nonlinear logistic sigmoidal curve:
\begin{align}
    \mathcal{T}(\rho ) = \tau _{\text{min}}+\frac{\tau _{\text{max}}-\tau _{\text{min}}}{1+\exp \big(\lambda (\rho -\rho _{0})\big)} \label{eqn:dual_sigmoid_temperature}
\end{align}
where $\tau_{\text{max}}$ is the maximal temperature baseline value, $\tau_{\text{min}}$ is the minimal convergence floor, $\rho_{0}$ is the inflection midpoint $\rho_{0}$, and $\lambda$ is a logistic steepness (slope) factor. Note that we configured our schedule in Equation \ref{eqn:dual_sigmoid_temperature} above to be: $\tau_{\text{max}} = 0.012$, $\tau_{\text{min}} = 0.0012$, $\rho_0 = 0.35$, and $\lambda = 10.0$. 
% $\mathcal{T}(k)=0.0012+\frac{0.0108}{1+\exp \left(10\cdot \left[\frac{k}{K-1}-0.35\right]\right)}$

%% saccadic optimization timeline (2-phase foraging)
The above temperature schedule emulates the two general, distinct phases of visual search behavior: exploration followed by exploitation. At the start of the MPC glimpsing trajectory ($k < 0.35K$), the temperature remains high, i.e., $\mathcal{T} \approx \tau_{\text{max}}$. The higher values of $\mathcal{T}$ scale up injected Gumbel perturbations, which forces the MPC saccade planner to execute wider, highly stochastic macro-saccades in order to explore the outer boundaries of the sensory input canvas; this can further aid in preventing MPC from getting stuck in local loops induced by familiar textures. 
As progress within the trajectory crosses the midpoint ($k > 0.35K$), the exponential logistic term rapidly cools down the temperature toward the set floor $\tau_{\text{min}}$; this suppresses the Gumbel noise and causes the planner to transition to the more deterministic active coordinate selection scheme/equation in the main paper. % of Equation \ref{eqn:epistemic_foraging}. 
Ultimately, this means that, after crossing this midpoint, MPC will spend the rest of its glimpse budget focused on higher utility sensory input features.

\section{Analysis of View Extraction Schemes}

In this supplementary section, we provide early (preliminary) experimental results (on MNIST) for a small set of configurations of the input stream patch extraction scheme that we utilize for the MPC encoder models. Table \ref{table:mpc_ablations} presents the results of these particular experiments. Concretely, under each particular configuration of the input stream, we fit an MPC encoder to the training data as in the main paper, then probe the quality of the embeddings with: 
\textbf{1)} a non-parametric KNN classifier, and 
\textbf{2)} a parametric, nonlinear attentive probe/classifier (set up in the same way as in \cite{drozdov2024video}). 
The setup for the KNN probe is the same as described in the main paper. The nonlinear attention probe was also trained with frozen latent codes collected, under similar conditions to the KNN probe except that the attentive probe was fit/optimized to the codes using the Adam optimizer, used drop-out for regularization, and employed a decaying adaptive learning rate. 

%%%%%%%%%%%%%%%%%%%%%%%%%%%%%%%%%%%%%%%%%%%%%%%%%%%%%%%%%%%%%%%%%%%%%%%
\begin{table}[!t]
\begin{center}
\begin{tabular}{c | c | c | c | c} 
 \hline
  & \multicolumn{2}{c|}{\textbf{F-PF-P}} & \multicolumn{2}{c}{\textbf{F-P}} \\
  \textbf{\# Fovea Streams} & \textbf{KNN} (\%) & \textbf{Attn-ACC} (\%)  & \textbf{KNN} (\%) & \textbf{Attn-ACC} \\
 \hline\hline
 $C = 1$ & $96.47 \pm 0.13$ & $97.30 \pm 0.09$ & $95.79 \pm 0.11$ & $97.66 \pm 0.06$  \\
 $C = 2$ & $97.21 \pm 0.06$ & $98.02 \pm 0.12$  & $96.74 \pm 0.09$ & $98.17 \pm 0.04$  \\
 $C = 4$ & $98.10 \pm 0.10$ & $98.80 \pm 0.05$  & $97.73 \pm 0.14$ & $98.44 \pm 0.06$  \\
 \hline 
\end{tabular}
\vspace{0.2cm}
\caption{ 
\small{
\textbf{Generalization on MNIST of meta-representational predictive coding under different multi-stream structures.} 
We examine MPC generalization ability under different formulations of its stream structure -- `F-PF-P` denotes foveal-parafoveal-peripheral whereas `F-P` denotes foveal-peripheral; $C = \{1, 2, 4\}$ controls the amount of foveal streams arranged for the structure (while one parafoveal or peripheral view is used). We specifically measure generalization in terms of downstream classification ability (in terms of \%), as measured by a linear probe (Lin-ACC) and a nonlinear attentive probe (Attn-ACC) \cite{drozdov2024video}.
}
}
\label{table:mpc_ablations}
\vspace{-0.5cm}
\end{center}
\end{table}
%%%%%%%%%%%%%%%%%%%%%%%%%%%%%%%%%%%%%%%%%%%%%%%%%%%%%%%%%%%%%%%%%%%%%%%

Specifically, we analyze the performance of MPC on MNIST under a small set of several possible stream configurations; concretely, we investigate the value of having only one coarser grained view (i.e., just the peripheral) or two (i.e., the parafoveal patch and the peripheral patch) as well as having only a single foveal or multiple (two or four foveal patches). As a result, the input stream configurations that we preliminarily investigated included the following:
\begin{itemize}[noitemsep,nolistsep]
    \item \textbf{Foveal, parafoveal, peripheral (F-PF-P)}: 
    \begin{itemize}
        \item One foveal ($C=1$) + parafoveal + peripheral
        \item Two foveal ($C=2$) + parafoveal + peripheral
        \item Four foveal ($C=4$) + parafoveal + periperhal
    \end{itemize}
    \item \textbf{Foveal and only peripheral (F-P)}: 
    \begin{itemize}
        \item One foveal ($C=1$) + peripheral
        \item Two foveal ($C=2$) + peripheral
        \item Four foveal ($C=4$) + periperhal
    \end{itemize}
\end{itemize}

In Table \ref{table:mpc_ablations}, we observe that the best classification performance (in terms of both the KNN and nonlinear attentive probes) is obtained with $C=4$, $F=1$, and $P=1$. 
It is important to note that more foveal streams resulted in improved performance in either of these overall settings (with $C=4$ giving the best in the set of configurations explored). It would prove beneficial if future work were to explore other configurations of the input streams, including both the quantity as well as the spatial/topological arrangement.

\section{Limitations and Directions for Neuroscience-informed Self-Supervised Learning} 
\label{sec:limitations_future_work}

As promising as the proposed framework of MPC is for (Neuro-)SSL, there are several limitations which afford avenues for future research and development. From a computational neuroscience perspective, while our MPC model designs are biomimetic, leading us to a computational architecture with foveal, parafoveal, and peripheral streams, MPC is still a loose inspiration and does not faithfully model the bio-circuitry that underlie visual hierarchies and the oculomotor system. Future effort that modifies the MPC architecture to better emulate biological details (e.g., crafting constrained layered structures that directly adhere to known neuroanatomy, operating with spikes as opposed to rate-codes, etc.) may further benefit the generalization ability of models constructed within our framing. Furthermore, it might prove fruitful to carry out neurobiological studies that ask if a inter/intra-stream prediction/message passing scheme, like that of MPC, affords a useful explanation of empirical neuronal responses. 

From a machine intelligence point-of-view, while this work demonstrates that MPC can extract representations that facilitate promising downstream performance -- demonstrating the viability of our neuro-mimetic scheme -- further experimental studies will be needed, e.g., those carried out on more complex data; including video databases. These efforts may help determine what extensions/mechanisms might be required to ensure generalization across a greater variety of (visual) sensoria/niches. It is noteworthy that our MPC framework does not rely on large batches\footnote{We used a batch size greater than one solely to speed up simulation; MPC is inherently an online learning framework.} or batch statistics/normalization \cite{grill2020bootstrap} %\cite{grill2020bootstrap,chen2021exploring} 
or negative samples as many of the modern deep learning \cite{chen2020simple} %\cite{oord2018representation,jaiswal2020survey,chen2020simple}  
and neuro-mimetic schemes \cite{hinton2022forward,ororbia2024contrastive} %\cite{hinton2002training,hinton2022forward,ororbia2023predictive,ororbia2024contrastive} 
do. This means MPC may offer a regularization-based \cite{garrido2024learning} %\cite{garrido2024learning,bardes2024revisiting}
SSL method that performs biologically-plausible inference and credit assignment. However, the failure cases of MPC -- as well as the missing heuristics needed to scale it to be competitive with modern-day generative AI -- will need to be developed. It will further be crucial to adapt and apply the analytical apparatus of self-supervised and representational learning \cite{geiping2023cookbook}, such as characterizing the problems of representational and dimensionality collapse. %\cite{jing2021understanding}. 
It is our hope that our MPC framework encourages exploration of neuroscience-motivated intelligence and NeuroAI \cite{ororbia2023brain,ororbia2024review,zador2026neuroai} in the context of SSL, a space that we label as biomimetic or \emph{neuroscience-informed self-supervised learning} (Neuro-SSL), of which MPC represents one possible computational instantiation. 

Another key aspect that would benefit our framework the most is 
%that is missing from our framework is
a more sophisticated mechanism for context-driven motor control\footnote{Such mechanisms would prove useful for incorporating voluntary oculomotor control as well as facilitate other types of eye movement, such as smooth pursuit or vergence.} further inspired by machine intelligence efforts such as \cite{mnih2014recurrent,sharafeldin2024active}, as it well-known that contextual task knowledge plays an important role in determining where a biological vision system looks \cite{rothkopf2007task,ballard2009modelling}. This next generalization is more broadly related to an implementation of the free energy principle known as active inference \cite{friston2016active} and, since our MPC model is already action-conditional (i.e., its representations are ``aware'' of actions taken) and it takes actions in accordance to a reflex-based (unlearned) form of epistemic foraging, our framework could yield to stronger, more flexible generalizations under active inference. 
In other words, it could deploy saccadic eye movements based on a learned motor-control policy that maximizes expected information gain -- or epistemic affordance -- of visual samples. See \cite{friston2012perceptions,parr2017working,parr2017active,parr2017uncertainty} for worked examples. 

\end{document}